\documentclass{article}

\usepackage{microtype}
\usepackage{graphicx}
\usepackage{subcaption}
\usepackage{url}            
\usepackage{doi}
\usepackage{booktabs}       
\usepackage{amsfonts}       
\usepackage{nicefrac}       
\usepackage{booktabs} 

\usepackage{mathtools}
\usepackage{bbm}
\usepackage{wrapfig}
\usepackage{subcaption}
\usepackage{caption}
\usepackage{multirow}
\usepackage{makecell}
\usepackage{pifont}
\usepackage[dvipsnames]{xcolor}
\usepackage{longtable}
\usepackage[T1]{fontenc}

\usepackage{hyperref}

\usepackage[accepted]{icml2026}

\usepackage{amsmath}
\usepackage{amssymb}
\usepackage{mathtools}
\usepackage{amsthm}

\usepackage[capitalize,noabbrev]{cleveref}

\theoremstyle{plain}

\theoremstyle{definition}

\theoremstyle{remark}

\usepackage[textsize=tiny]{todonotes}

\icmltitlerunning{Query Circuits: Explaining How Language Models Answer User Prompts}

\definecolor{citeblue}{HTML}{1565C0}
\definecolor{citeorange}{HTML}{D35400}
\hypersetup{
  colorlinks=true,
  linkcolor=citeblue,
  citecolor=citeblue,
  urlcolor=citeblue
}

\begin{document}

\twocolumn[
  \icmltitle{Query Circuits: Explaining How Language Models Answer User Prompts}



  \icmlsetsymbol{equal}{*}

\begin{icmlauthorlist}
    \icmlauthor{Tung-Yu Wu}{oxford}
    \icmlauthor{Fazl Barez}{oxford,martian}
\end{icmlauthorlist}

\icmlaffiliation{oxford}{University of Oxford}
\icmlaffiliation{martian}{Martian}

\icmlcorrespondingauthor{Tung-Yu Wu}{b08901133@ntu.edu.tw}

  \icmlkeywords{Circuit Discovery, AI Interpretability, Machine Learning, ICML}

  \vskip 0.3in
]



\printAffiliationsAndNotice{}  

\addtocontents{toc}{\protect\setcounter{tocdepth}{0}}
\begin{abstract}
Explaining why a language model produces a particular output requires local, input-level explanations. Existing methods uncover global capability circuits (e.g., indirect object identification), but not why the model answers a specific input query in a particular way. We introduce \textit{query circuits}, which directly trace the information flow inside a model that maps a specific input to the output. Unlike surrogate-based approaches (e.g., sparse autoencoders), query circuits are identified within the model itself, resulting in more faithful and computationally accessible explanations. To make query circuits practical, we address two challenges. First, we introduce Normalized Deviation Faithfulness (NDF), a robust metric to evaluate how well a discovered circuit recovers the model's decision for a specific input, and is broadly applicable to circuit discovery beyond our setting. Second, we develop sampling-based methods to efficiently identify circuits that are sparse yet faithfully describe the model’s behavior. Across benchmarks (IOI, arithmetic, MMLU, and ARC), we find that there exist sparse query circuits within the model that recover much of its performance on single queries. For example, on average, a circuit covering only 1.3\% of model connections can recover about 60\% of performance on an MMLU question. Overall, query circuits provide a step towards faithful, scalable explanations of how language models process individual inputs. The project page is at \url{https://tony10101105.github.io/query-circuit/}.
\end{abstract}
\section{Introduction}
\label{intro}


Explaining the decisions of large language models (LLMs) is essential for their deployment in high-stakes domains such as medicine~\citep{amann2020explainability} and autonomous systems~\citep{9616449}. For example, when a medical AI agent receives a query from clinicians and decides whether a patient should undergo surgery, its reasoning must be interpretable to ensure the decision does not rely on spurious/shortcut features~\citep{yuan-etal-2024-llms}; when an autonomous vehicle selects an incorrect control action, its failure mode must be explainable to allow accurate attribution of responsibility.

Circuit discovery~\citep{conmy2023towards, hanna2024have, ameisen2025circuit} has emerged as a popular approach for explaining model mechanisms~\citep{kharlapenko2025scaling}. However, most studies only investigate circuits of simple inference patterns, such as indirect object identification (IOI)~\citep{wang2023interpretability} and greater-than (GT) comparison~\citep{hanna2023how}. Though valuable, these circuits do not explain how a model produces a particular output for a user input query given in the wild.

Approaches for identifying instance-level circuits have recently emerged~\citep{marks2025sparse, ameisen2025circuit}. They largely rely on surrogate models such as sparse autoencoders (SAEs)~\citep{sae} and cross-layer transcoders (CLTs)~\citep{clt}. A prominent example is Circuit Tracing~\citep{ameisen2025circuit}, a CLT-based framework for discovering instance-level circuits that explain input-specific model behavior. While circuit discovery is often easier in surrogate models due to their sparse activations, these surrogates frequently fail to faithfully reconstruct model activations~\citep{ameisen2025circuit} and may not capture the true computational mechanisms of the LLM~\citep{marks2024enhancing, cltunfaithfulness}, limiting their reliability. In addition, training surrogate models is computationally expensive~\citep{templeton2024scaling}, reducing their accessibility. Other prior works, including circuit discovery in vision models~\citep{kwon2025granular} and input-dependent feature analyses in LLMs~\citep{selfie, ghandeharioun2024patchscopes}, likewise do not provide prompt-level explanations without reliance on surrogate models.

\begin{figure*}[tb]
  \centering
  \includegraphics[width=0.9\textwidth]{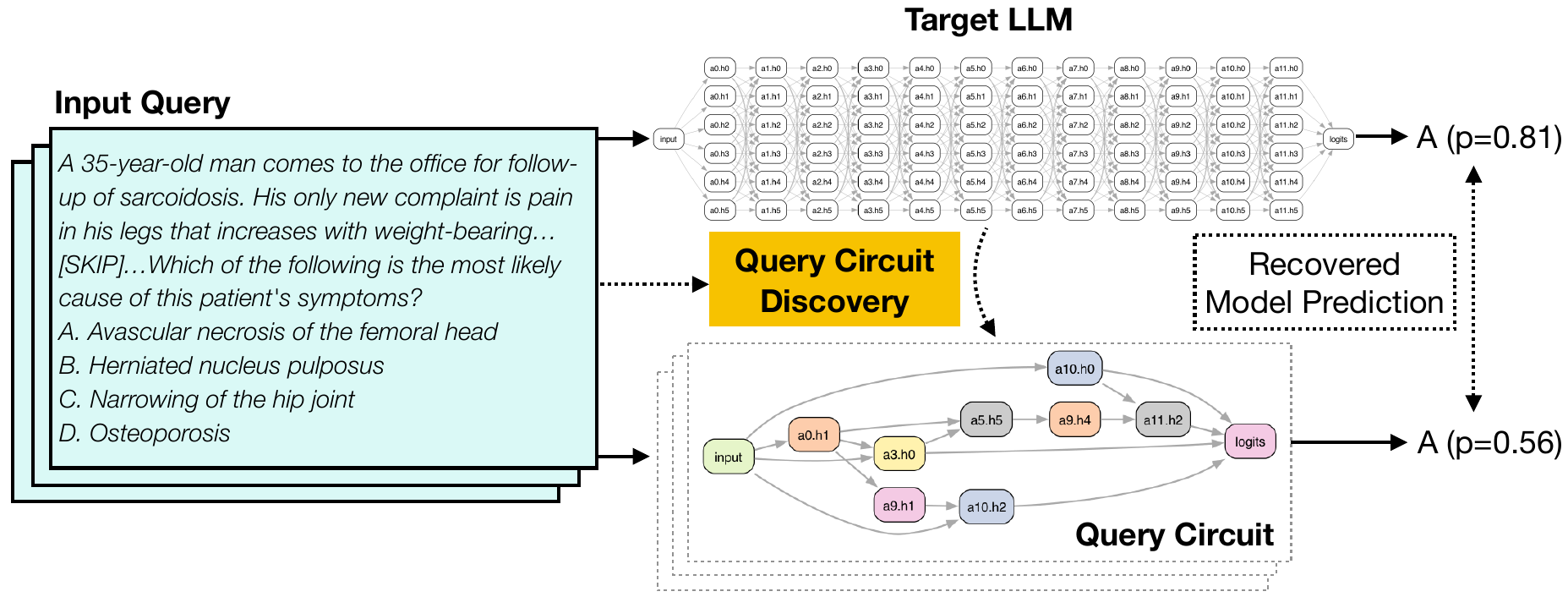}
  \caption{\textbf{Query circuit discovery aims to identify a sparse sub-network within the LLM that underlies the model response to a user input query.} The LLM and circuit in this illustration are simplified for visualization.}
  \label{fig: teaser}
\vspace{-1em}
\end{figure*}

This paper introduces the task of \textbf{query circuit discovery}: uncovering the specific circuit \textit{directly within} (i.e., in-place) an LLM that drives its response to a single input query. Unlike prior work, which either discovers in-place \textit{capability circuits} (e.g., IOI) or instance-level circuits within surrogates (e.g., Circuit Tracing), query circuits provide instance-level explanations by tracing information flow inside the original model (Figure~\ref{fig: teaser}). More specifically, circuits discovered by Circuit Tracing consist of nodes and edges defined entirely on CLTs, whereas our circuits are defined directly on the original model, with SAEs optionally applied post hoc for representation decomposition. Overall, our proposed framework is more accessible for instance-level mechanistic analysis, eliminating dependence on CLTs while making SAEs optional rather than foundational.

The two approaches are complementary: CLT-based circuits offer finer-grained representations but require highly faithful CLTs, whereas query circuits with optional SAEs are more portable, and the circuit construction process itself does not depend on SAE faithfulness.

We highlight key challenges of query circuit discovery and propose methods to address them. First, the widely adopted Normalized Faithfulness Score (NFS)~\citep{hanna2024have, zhang2025eapgp, marks2025sparse} used to assess how well a circuit recovers the model performance becomes unstable on general datasets (e.g., MMLU), and thus fails to reliably indicate when circuits of increasing size begin to capture model behavior. We therefore introduce \textit{Normalized Deviation Faithfulness (NDF)}, a more robust metric for query circuit evaluation. Second, existing methods from capability circuit discovery often fail to yield compact and faithful query circuits. To overcome this, we propose \textit{Best-of-N (BoN)} sampling and two variants—interpolated BoN (iBoN) and BoN with constraint-adaptive score matrix (BoN-CSM)—which reliably recover faithful query circuits.

With BoN, we conduct experiments across multiple benchmarks (IOI, arithmetic, MMLU~\citep{hendrycks2021measuring}, and ARC Challenge~\citep{allenai:arc}) to show that \textit{even for complex natural queries, compact query circuits can still be found within the LLM that account for a considerable portion of its responses}. For example, using BoN, we find that for a multiple-choice question (MCQ) in MMLU, a query circuit with only 1.3\% of the target LLM’s edges can, on average, recover roughly 60\% of the model’s behavior on that query. In summary, our contributions are threefold:
\begin{itemize}
    \item We formulate the task of \textbf{query circuit discovery}, contrasting it with both capability circuit discovery and surrogate-model-based approaches.
    \item We identify and address two key technical challenges: (i) unreliable evaluation of query circuits by the previous metric (NFS), for which we propose \textit{Normalized Deviation Faithfulness}; and (ii) failure of existing methods to find compact and faithful query circuits, for which we propose \textit{Best-of-N} sampling and its variants.
    \item Across diverse datasets, we demonstrate that even a small circuit within the model can explain much of the model behavior on the individual query, showing query circuit discovery as a practical path toward faithful and scalable prompt-level LLM decision explanations.
\end{itemize}

\section{Background: Capability Circuit Discovery}
\label{sec: background}
This section provides the technical background of circuit discovery. Section~\ref{subsec: transformer circuit} reviews transformer circuits, while Section~\ref{subsec: capability circuit discovery} introduces capability circuit discovery, including how to discover a circuit (Section~\ref{subsubsec: score edge and construct circuit}) and how to evaluate both the circuit and the discovery method (Section~\ref{subsubsec: evaluate capability circuit}).

\subsection{Transformer Circuits}
\label{subsec: transformer circuit}
Transformer circuits~\citep{elhage2021mathematical} represent an LLM $M$ as a directed acyclic graph with node and edge sets $\{V,E\}$, where, following prior work~\citep{syed-etal-2024-attribution, conmy2023towards}, each node in $V$ is an MLP or attention head, and edges in $E$ are where the outputs of earlier nodes feed into later ones, defined via \textit{residual rewrite}~\citep{elhage2021mathematical, nanda2022transformerlens}. WLOG, a circuit can be specified by its edge set $E$, omitting the explicit node set $V$. For a given capability, a compact, faithful circuit that captures the critical information flow among components enables more precise and efficient interpretability research~\citep{quirke2024understanding, lan-etal-2024-towards}.



\subsection{Capability Circuit Discovery}
\label{subsec: capability circuit discovery}
Given a target LLM $M$ with edge set $E$ and a capability of interest (e.g., IOI), capability circuit discovery aims to identify a capability circuit $C_c$ with edge set $E_c \subset E$ that captures $M$’s underlying mechanisms for this capability. To study the capability, it is instantiated as a dataset $D$ of queries, where each query is designed so that answering it correctly requires the model to use that capability.
\subsubsection{Edge Scoring and Circuit Construction}
\label{subsubsec: score edge and construct circuit}
Prior methods from capability circuit discovery typically construct the circuit by selecting edges based on their influence on the model’s outputs. An edge $e$'s importance score $a_e$ is defined as its averaged indirect effect (IE)~\citep{NEURIPS202092650b2e} on the model's performance over the dataset $D$:
\begin{equation}
    a_e \coloneqq \frac{1}{|D|} \sum_{q \in D} 
    \Bigl( 
        L\!\left(M\!\left(q \,\middle|\, \operatorname{do}(e \leftarrow e')\right)\right) 
        - L\!\left(M(q)\right) 
    \Bigr),
    \label{eq: ie}
\end{equation}
where $L(\cdot)$ is a performance metric for each query $q$, such as the logit difference between the correct and incorrect tokens~\citep{heimersheim2024use}. The operator $\operatorname{do}(e \leftarrow e')$ denotes corrupting edge $e$ by replacing its propagated feature with a corrupted feature $e'$. The $e'$ is obtained by feeding the LLM a corrupted query $q'$, constructed by removing the key factual or linguistic cue in the original query $q$ that guides the model’s solution. Details of corrupted queries for different question types we studied are provided in Appendix~\ref{sup: experimental details}. The scores of all edges can arrange as an edge score matrix $S\in\mathbb{R}^{n \times n}$, where $n$ denotes the number of nodes. Notably, Equation~\ref{eq: ie} is not additive, i.e., $a_{e_i \cup e_j} \neq a_{e_i} + a_{e_j}$, where $a_{e_i \cup e_j}$ denotes the effect of corrupting $e_i$ and $e_j$ in the same forward pass.


Approaches that compute $a_e$ directly via Equation~\ref{eq: ie}, such as ACDC~\citep{conmy2023towards}, are referred to as edge activation patching methods~\citep{zhang2024towards}. They require two forward passes of $M$ to score each edge. To improve efficiency, some recent studies~\citep{hanna2024have, marks2025sparse} reformulate IE computation as integrated gradients (IG)~\citep{sundararajan2017axiomatic}:
\begin{equation}
\label{eq: ig}
\begin{aligned}
a_e
&= (e-e')^{\top}\int_{0}^{1}
\nabla_{e} M\!\left(z' + \alpha (z - z')\right)\, d\alpha \\
&\approx (e-e')^{\top}\frac{1}{m}\sum_{k=1}^{m}
\nabla_{e} M\!\left(z' + \frac{k}{m}(z - z')\right).
\end{aligned}
\end{equation}

where $z$ and $z'$ are the token embeddings of $q$ and $q'$. $m$ is the discretization step. Averaging over $D$ is omitted for simplicity. Equation~\ref{eq: ig} approximates all edges’ IEs in parallel, requiring a fixed number of forward passes regardless of the edge count. Approaches applying Equation~\ref{eq: ig}, such as EAP~\citep{syed-etal-2024-attribution}\footnote{Although the original EAP paper~\citep{syed-etal-2024-attribution} frames it as a linear approximation of Equation~\ref{eq: ie} via the Taylor series, it can also be interpreted as applying integrated gradients with a discretization step of $m=1$.} and EAP-IG~\citep{hanna2024have}, are referred to as edge attribution patching methods.

Using the computed edge scores, capability circuit discovery methods construct the capability circuit $C_c$ given a budget of $N$ edges. Two straightforward approaches are: (i) greedily selecting $N$ edges with the highest scores~\citep{hanna2024have}, and (ii) selecting nodes or edges whose scores exceed a predefined threshold~\citep{conmy2023towards, marks2025sparse}. A more sophisticated method is Dijkstra-like iterative construction~\citep{conmy2023towards, hanna2024have}: Start from the logit node and iteratively add back influential edges whose child node is already included in the circuit.

\subsubsection{Evaluation of Capability Circuit and discovery method}
\label{subsubsec: evaluate capability circuit}
Normalized Faithfulness Score (NFS)~\citep{marks2025sparse, zhang2025eapgp, mueller2025mib} has been widely adopted to quantify how well the discovered capability circuit $C_c$ recovers the original LLM $M$'s performance on $D$. It is defined as:
\begin{equation}
    NFS(C_c)\coloneqq 
    \frac{L(C_c(D))-L(M(D'))}{L(M(D))-L(M(D'))},
    \label{eq: nfs}
\end{equation}
where $L(C_c(D))$ denotes the overall performance of $C_c$ on $D$. $D'\coloneqq\{q_i^{'} \mid q_i \in D \}$. $NFS(C_c)$ measures the fraction of $M$'s performance on $D$ recovered by $C_c$. $NFS(C_c)=1$ indicates $C_c$ perfectly recovers $M$'s performance. $NFS(C_c)=0$ means $C_c$ performs the same as $M$ on corrupted queries. In toy tasks (e.g., IOI), where capability circuits have mainly been studied, NFS typically falls within [0, 1]~\citep{hanna2024have}, although its definition (Equation~\ref{eq: nfs}) does not guarantee boundedness. A discovery method is more effective if, across varying numbers of edges $N$, it consistently identifies circuits with higher (close to 1) NFS than the counterparts (i.e., a better Pareto frontier).

\section{Proposal: Query Circuit Discovery}
\label{sec: problem formulation}
This section formulates the objective and evaluation of query circuit discovery (Sections~\ref{subsec: objective of query circuit} and \ref{subsec: evaluation of query circuit}), and discusses technical challenges (Section~\ref{subsec: technical challenges of query circuit}), including the instability of existing evaluation metrics (Section~\ref{subsubsec: nfs bad}) and the ineffectiveness of vanilla capability circuit discovery methods in the single-query setting (Sections~\ref{subsubsec: capability circuit discovery degrades} and \ref{subsubsec: high edge budge for complex query}).

\subsection{Objective}
\label{subsec: objective of query circuit}
Our proposed query circuit discovery seeks methods that, for any natural query $q$ and an edge budget $N$, consistently identify a faithful circuit $C_q$ defined by the edge set $E_q \subset E $ that captures the mechanisms by which the target LLM $M$ answers that query. Its goal differs from capability circuit discovery: the former aims to trace and analyze the internal states of an LLM as it processes and responds to a user input (\textit{local interpretations}), whereas the latter examines how an LLM implements particular algorithmic skills (\textit{global interpretations})~\citep{bereska2024mechanistic}.



\subsection{Evaluation of Query Circuit and Discovery Method}
\label{subsec: evaluation of query circuit}
Similar to capability circuits, we aim to develop a faithfulness measure to quantify how well the discovered query circuit $C_q$ recovers the original LLM $M$'s performance on the query $q$, denoted as $F(\cdot): C_q \to \mathbb{R}$. To evaluate a discovery method, we average $F(C_q)$ of different queries across a dataset $D$ (e.g., MMLU). Under an edge budget $N$, the performance of a discovery method is
\begin{equation}
    \frac{1}{|D|} \sum_{q \in D} F(C_q),
    \label{eq: 6}
\end{equation}
where each $C_q$ has $N$ edges. A query circuit discovery method is more effective if, under varying $N$, it consistently produces query circuits with higher faithfulness scores than the counterpart. A straightforward choice of $F(\cdot)$ is inheriting the NFS metric, but we argue that it is unreliable and a suboptimal choice for evaluating query circuits and discovery methods, detailed in Section~\ref{subsubsec: nfs bad}.


\subsection{Technical Challenges}
\label{subsec: technical challenges of query circuit}

\subsubsection{Instability of Normalized Faithfulness Score on general datasets}
\label{subsubsec: nfs bad}
NFS has primarily been used to evaluate capability circuits on toy tasks with researcher-curated data (e.g., IOI). However, we find that it is not a reliable faithfulness measure on more general datasets of greater interest (e.g., MMLU). Figure~\ref{subfig: metric-unfaithful} reports the NFS of three query circuits discovered by EAP-IG under varying edge budgets $N$. Unless otherwise stated, we adopt EAP-IG~\citep{hanna2024have} throughout this paper as the MIB benchmark~\citep{mueller2025mib} finds it to be the most effective method. We randomly sample three queries from the MMLU Marketing. Llama-3.2-1B-Instruct (386713 edges)~\citep{dubey2024llama} is the target model. Results show large fluctuations, with NFS values often exceeding 1 or dropping below 0 at different $N$. This instability undermines both the evaluation of circuit quality and the monitoring of discovery progress as $N$ increases~\citep{miller2024transformer}. We therefore propose NDF as an alternative metric to evaluate the faithfulness of query circuits, detailed in Section~\ref{sec: ndf}.

\subsubsection{Degradation of Capability Circuit Discovery Methods in Query Settings}
\label{subsubsec: capability circuit discovery degrades}
\begin{figure*}[tb]
  \centering
  \begin{subfigure}{0.3\linewidth}
    \includegraphics[width=\textwidth]{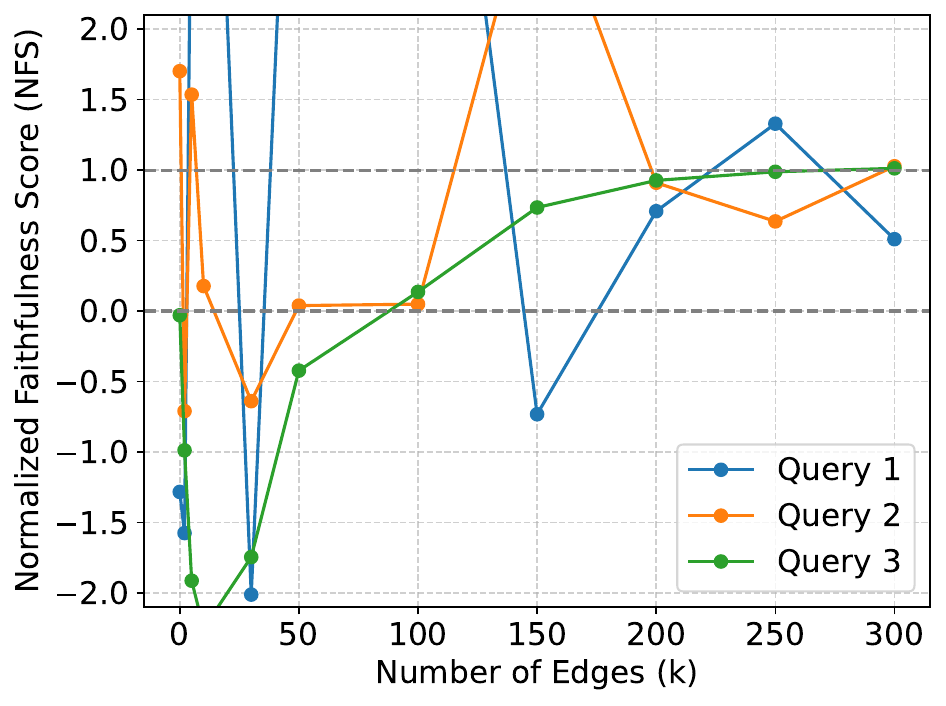}
    \caption{Case study of three queries from MMLU Marketing showing \textbf{NFS' instability when applied to assess query circuits of different sizes.}}
    \label{subfig: metric-unfaithful}
  \end{subfigure}
  \hfill
  \begin{subfigure}{0.3\linewidth}
    \includegraphics[width=\textwidth]{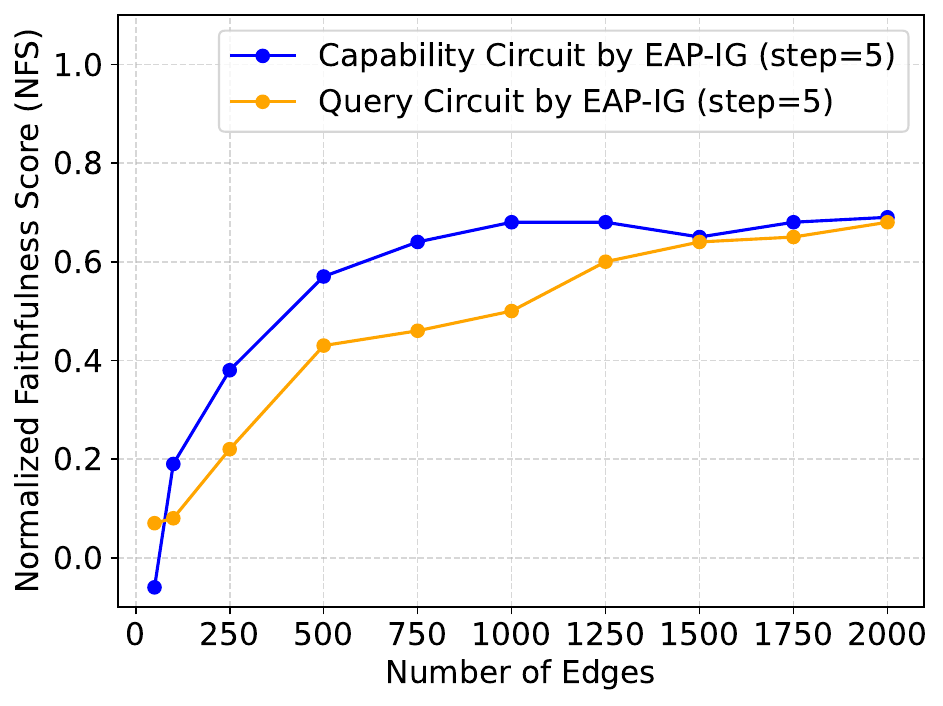}
    \caption{Case study on IOI dataset showing \textbf{directly applying methods for capability circuits may yield less-faithful query circuits.}}
    \label{subfig: single-query-low-performance}
  \end{subfigure}
  \hfill
  \begin{subfigure}{0.3\linewidth}
    \includegraphics[width=\textwidth]{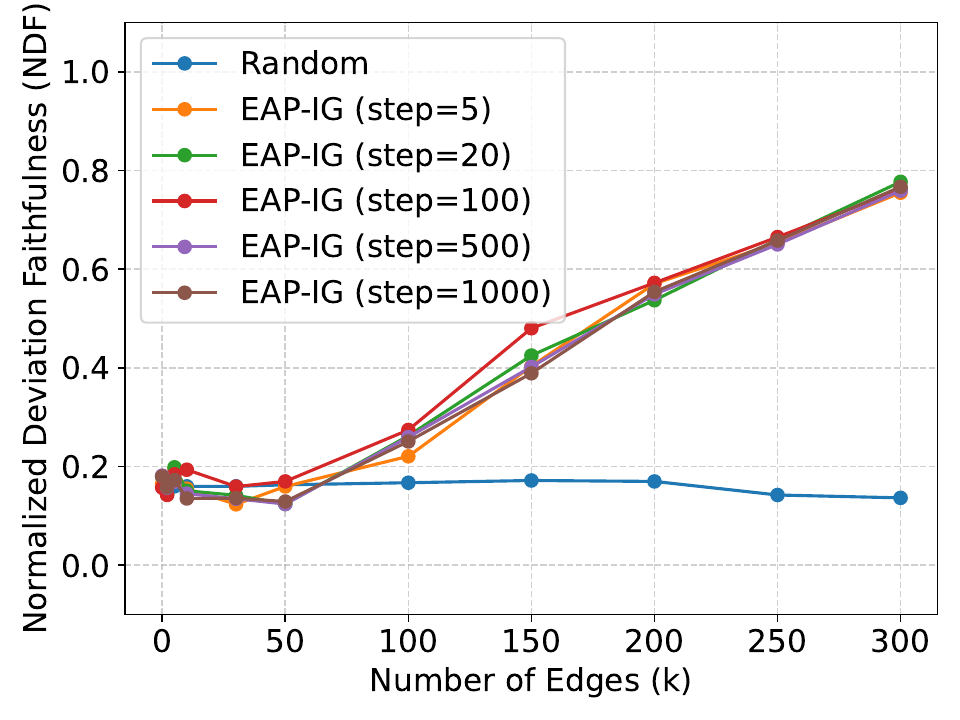}
    \caption{Case study on MMLU Astronomy showing that \textbf{on complex queries, many edges may be needed to recover non-trivial circuits.}}
    \label{subfig: astronomy-not-work}
  \end{subfigure}
  \caption{Technical challenges of query circuit discovery.}
  \label{fig: challenges}
\end{figure*}

We find that directly applying methods from capability circuit discovery to identify query circuits generally yields suboptimal results. Figure~\ref{subfig: single-query-low-performance} presents a case study on the IOI dataset. In IOI, all queries require the same capability to generate correct tokens, allowing the construction of both a capability circuit for all queries and individual query circuits for comparison. With GPT-2 Small (32491 edges)~\citep{radford2019language} as the target LLM and an edge budget of $N=1000$, the query circuit recovers on average less than 50\% of GPT-2 Small’s performance per query, while the capability circuit recovers roughly 65\% of the model’s overall performance on the dataset.

We attribute this degradation to two factors: (1) feature attribution suffers from \textit{gradient noise}~\citep{smilkov2017smoothgrad, kapishnikov2021guided, kim2019saliency}; and (2) the IE calculation (Equation~\ref{eq: ie}) ignores combinatorial effects among edges~\citep{shapley:book1952, shap}. For a given input, an edge transmitting irrelevant features may still exhibit non-zero gradients (and thus a non-zero attribution score) while contributing little when combined with others. This issue is less pronounced in capability circuit discovery, where $a_e$ is averaged over dataset $D$, diluting edges with only sporadically high scores.

\subsubsection{High Edge Budget Requirements for complex queries}
\label{subsubsec: high edge budge for complex query}
We find that directly applying capability-circuit methods to a single complex query requires far more edges to form a non-trivial circuit. Figure~\ref{subfig: astronomy-not-work} presents a case study on MMLU Astronomy with Llama-3.2-1B-Instruct as the target model, using random edge selection as a baseline. On average, EAP-IG needs about $100k$ edges ($25.9\%$) to surpass the random baseline. This ineffectiveness reflects either (i) the inherent need for many edges to capture natural-form MCQs or (ii) EAP-IG’s inability to identify faithful circuits. Thus, we propose BoN sampling (Section~\ref{sec: bon}) to test this hypothesis and attribute the issue to (ii) (Section~\ref{sec: exp}). Moreover, Figure~\ref{subfig: astronomy-not-work} shows that increasing the IG step does not improve discovery, consistent with our discussion (Section~\ref{subsubsec: capability circuit discovery degrades}) on gradient noise and combinatorial effects, which cannot be resolved by refining single-edge IEs.
\section{Normalized Deviation Faithfulness}
\label{sec: ndf}

\subsection{Definition and Properties}
\label{subsec: ndf definition}
The Normalized Deviation Faithfulness (NDF) of a query circuit $C_q$ is defined as
\begin{equation}
NDF(C_q) 
= 1 - \min\!\left( 
    \left| 
        \frac{L(M(q)) - L(C_q(q))}{L(M(q)) - L(M(q'))} 
    \right| , 1 
\right),
\label{eq: ndf}
\end{equation}
which measures the performance deviation of a query circuit $C_q$ from the target LLM $M$, normalized by $M$'s performance gain from the corrupted query to the original query. NDF is derived from the integrated circuit-model distance (CMD) introduced by the MIB benchmark, which quantifies the overall performance of circuit discovery methods. NDF differs from NFS in two key aspects. First, it is symmetric around $L(M(q))$, equally penalizing deviations above and below $M$'s performance on $q$. Second, NDF is bounded within the interval $[0, 1]$. $NDF(c_q)=0$ if the performance deviation exceeds $M$'s performance gap between the original and corrupted query; $NDF(C_q)=1$ when $C_q$ has the same performance as $M$. More discussions on the relations between NFS, NDF, and CMD are in Appendix~\ref{sup: joint discussion of nfs ndf cmd}.
\subsection{Qualitative Comparison with Normalized Faithfulness Score}
\label{subsec: ndf qualitative comparison}
\begin{figure*}[tb]
  \centering
  \includegraphics[width=0.9\textwidth]{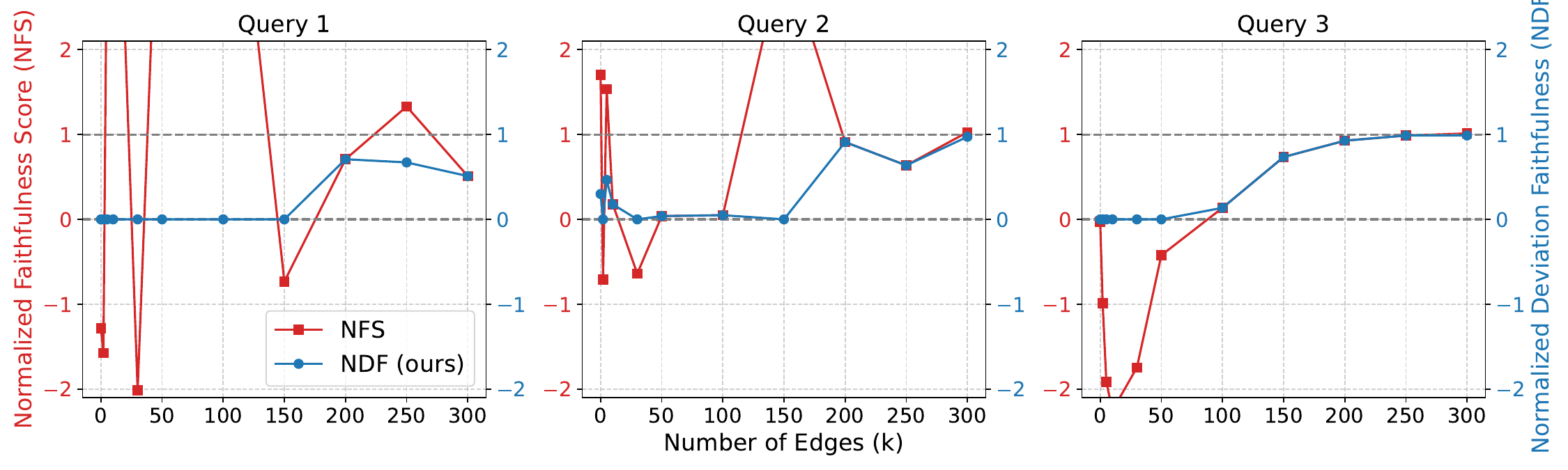}
  \caption{Complete evaluation results for the three queries adopted in Figure~\ref{subfig: metric-unfaithful} and Table~\ref{tab: nfs failure example}.}
  \label{fig: nfs_ndf_comparison}
\vspace{-1em}
\end{figure*}

Table~\ref{tab: nfs failure example} presents three examples of query circuit faithfulness evaluated using NFS and NDF. These queries are MCQs from the MMLU Marketing dataset. Target LLM $M$ is Llama-3.2-1B-Instruct. Performance metric $L$ is the probability difference between the correct option and the average of the three incorrect options. NFS exhibits numerical instability in several scenarios—for example, when $M$'s performance gap between $q$ and $q'$ is small (as in Query 1), or when $M$ achieves non-zero performance on $q'$ (e.g., due to position bias~\citep{zheng2024large}, as in Query 3). In contrast, our proposed NDF, which measures the faithfulness of $C_q$ as its normalized performance deviation from $M$, provides a more stable and reliable evaluation. Accordingly, we adopt NDF as the primary metric for all subsequent experiments. Figure~\ref{fig: nfs_ndf_comparison} presents complete evaluation results for three queries, further supporting this choice, with additional results provided in Appendix~\ref{supsub: nfs ndf complete comparison}.


\begin{table}[t]
  \caption{\textbf{Examples of evaluating three query circuits from Figure~\ref{subfig: metric-unfaithful} using NFS and NDF.} The corresponding queries are multiple-choice questions from the MMLU Marketing category. 
  }
  \label{tab: nfs failure example}
  \centering
  \addtolength{\tabcolsep}{4pt}
  \resizebox{1\linewidth}{!}{
  \begin{tabular}{@{}p{0.12\textwidth}p{0.08\textwidth}p{0.08\textwidth}p{0.08\textwidth}p{0.08\textwidth}p{0.08\textwidth}@{}}
    \toprule
    Query and Circuit Info & $L(M(q))$ & $L(M(q'))$ & $L(C(q))$ & \makecell[c]{NFS} & \makecell[c]{NDF}\\
    \specialrule{1.2pt}{2pt}{2pt} 
    \makecell[l]{Query 1\\$|C_q|=5k$} & \makecell[c]{-0.04} & \makecell[c]{-0.16} & \makecell[c]{0.10} & \makecell[c]{\textcolor{red}{2.15}} & \makecell[c]{\textcolor{ForestGreen}{0.00}}\\
    \midrule
    \makecell[l]{Query 2\\$|C_q|=250k$} & \makecell[c]{0.17} & \makecell[c]{0.39} & \makecell[c]{0.09} & \makecell[c]{\textcolor{red}{1.32}} & \makecell[c]{\textcolor{ForestGreen}{0.68}}\\
    \midrule
    \makecell[l]{Query 3\\$|C_q|=5k$} & \makecell[c]{0.96} & \makecell[c]{0.53} & \makecell[c]{-0.13} & \makecell[c]{\textcolor{red}{-1.57}} & \makecell[c]{\textcolor{ForestGreen}{0.00}}\\
    \specialrule{1.2pt}{0pt}{0pt} 
  \end{tabular}}
\vspace{-1.5em}
\end{table}

\section{Best-of-N Sampling for Query Circuit Discovery}
\label{sec: bon}
In this section, we introduce Best-of-N (BoN) sampling for query circuit discovery. We first present our motivation—a preliminary observation of circuit discovery on a query and its paraphrases in Section~\ref{subsec: po}, introduce BoN in Section~\ref{subsec: bon}, and then detail two extensions: (1) interpolated BoN (iBoN) in Section~\ref{subsec: ibon} and (2) BoN with Constraint-adaptive Score Matrix (BoN-CSM) in Section~\ref{subsec: bon-csm}.

\subsection{Observation: Failure in a Query, Success in Its Paraphrases}
\label{subsec: po}
\begin{figure}[t]
    \centering
    \small
    \includegraphics[width=0.75\linewidth]{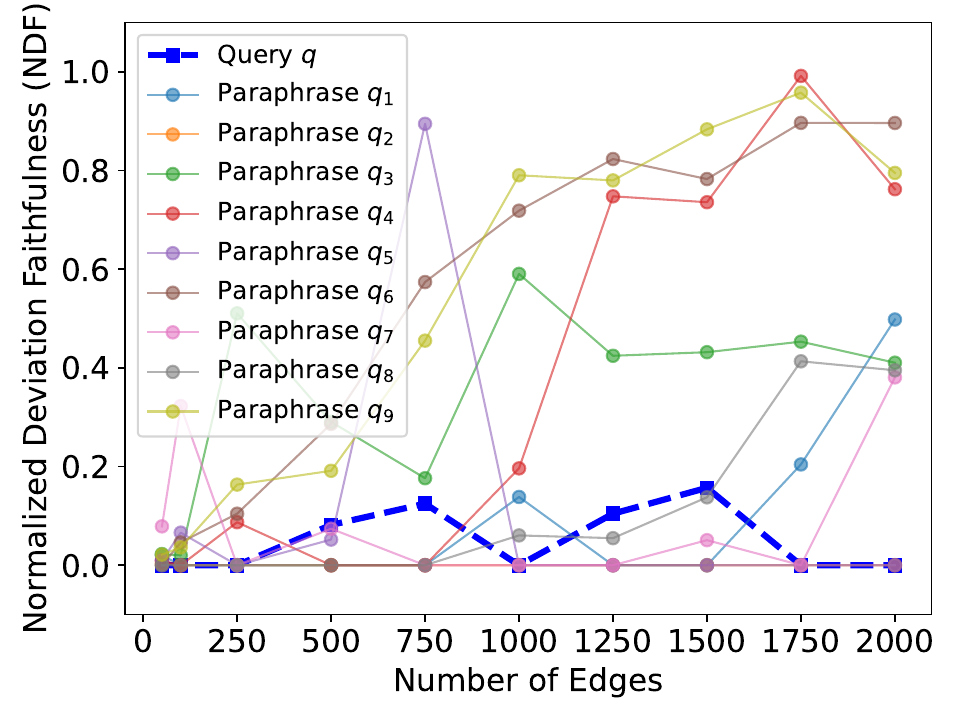}
    \caption{A case study on IOI dataset. \textbf{Circuits discovered by the original input query's paraphrases may recover model performance on the query.}}
    \vspace{-1.5em}
    \label{fig: observation-para-work}
\end{figure}
We find that \textit{while the circuit discovered on the original query may fail to faithfully recover model performance, circuits discovered on its paraphrases can succeed.} Figure~\ref{fig: observation-para-work} illustrates this with a query $q$ from the IOI dataset using GPT-2 Small as the target model. Although EAP-IG fails to directly identify a faithful circuit for $q$, it finds small and faithful ones in randomly selected paraphrases of $q$.

We argue that, due to gradient noise and the neglect of combinatorial effects (Section~\ref{subsubsec: capability circuit discovery degrades}), edge scoring based on Equations~\ref{eq: ie} and~\ref{eq: ig} for a query $q$ can only capture coarse score patterns—represented as a score matrix $S$ (examples are in Appendix~\ref{supsub: more examples of score matrix})—that roughly separate crucial from trivial edges, but are not precise enough to consistently select a set of edges that forms a faithful circuit. Score matrices from paraphrases can be viewed as perturbations of $S$: while they share similar patterns, small differences in edge scores can considerably alter which edges are selected. In this case, finding a faithful circuit within the model is akin to a lottery~\citep{frankle2018the}: circuits discovered by the original query and its paraphrases are ``tickets,'' and the one that successfully recovers the model performance on the query is the ``winning ticket.''

\subsection{Best-of-N Sampling}
\label{subsec: bon}
\begin{figure*}[tb]
  \centering
  \includegraphics[width=0.8\textwidth]{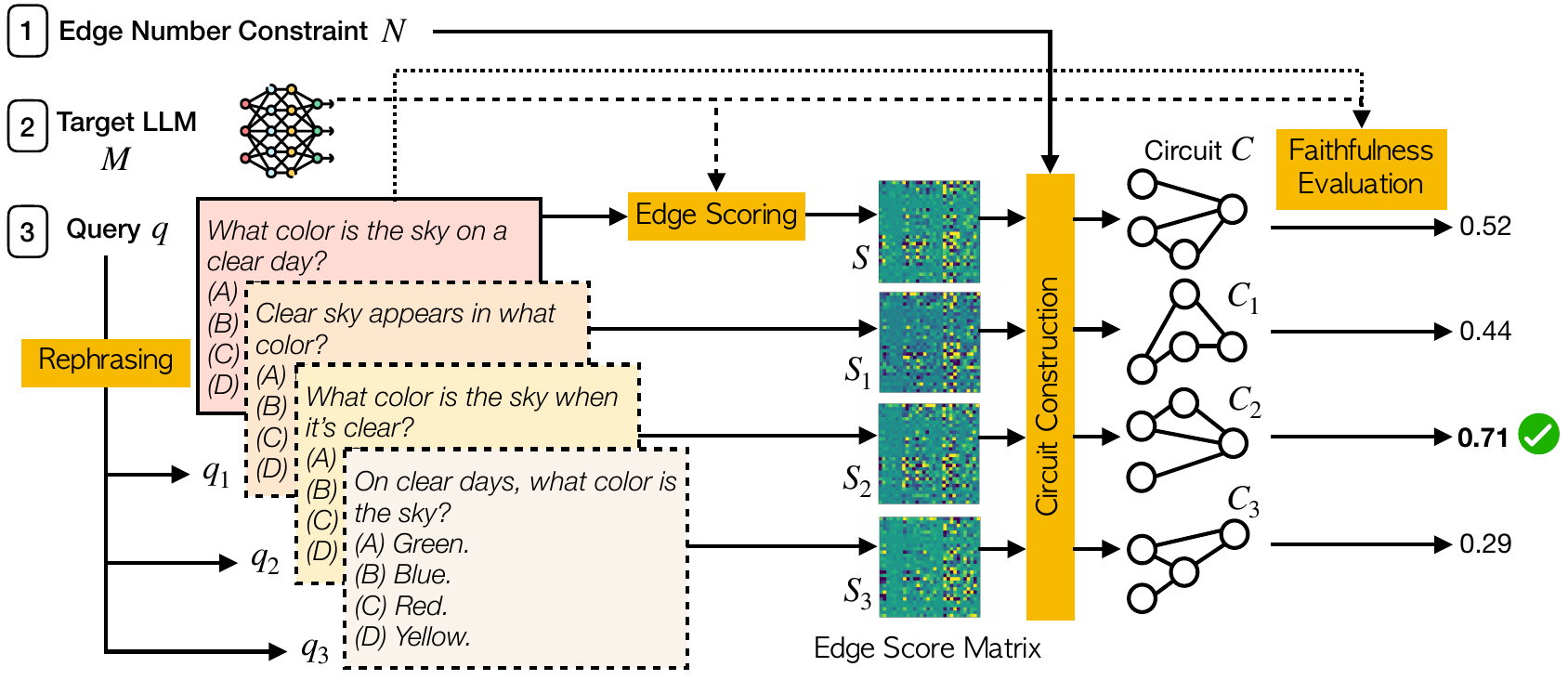}
  \caption{\textbf{The pipeline of Best-of-N sampling for discovering a faithful query circuit of $N$ edges for an input query $q$,} for which it generates $p$ paraphrases. $p=3$ in this illustration.}
  \label{fig: bon-main-framework}
\vspace{-0.5em}
\end{figure*}

Based on the observation in Section~\ref{subsec: po}, we introduce Best-of-N (BoN) sampling for query circuit discovery. As shown in Figure~\ref{fig: bon-main-framework}, to find a ``winning ticket'', BoN first generates $p$ paraphrases of the original query $q$, denoted as $\{q_1, ..., q_p\}$ (e.g., p = 3 in Figure~\ref{fig: bon-main-framework}). Then, it calculates edge importance scores $a_e$ by each of $\{q, q_1, ..., q_p\}$, represented as edge score matrices $\{S, S_1, ..., S_p\}$ (see Appendix~\ref{supsub: more examples of score matrix} for examples of these matrices). Finally, it leverages $\{S, S_1, ..., S_p\}$ to form $p+1$ circuits, measure their faithfulness score, and select the one with the highest score.


Steps 1 and 2 are required only once when constructing circuits with different edge budgets $N$. However, step 3 needs $p+1$ forward passes of the target LLM $M$ to identify the best circuit for a given $N$, which becomes a time bottleneck if one aims to construct circuits of many sizes. To address this issue, we introduce two simple extensions of BoN: iBoN and BoN-CSM. Both build on BoN-discovered faithful circuits to accelerate the discovery of circuits of varying sizes.

\subsection{Interpolated Best-of-N}
\label{subsec: ibon}
Algorithm~\ref{alg:ibon} shows the procedure of interpolated Best-of-N (iBoN), with circuits denoted as their edge sets for simplicity. iBoN \textit{interpolates} between two previously discovered faithful circuits to efficiently form a new one without an LLM. Assume one has applied BoN to discover $k$ circuits $\{E_1, ..., E_k\}$ with different edge counts $N$ (WLOG assume $|E_i|<|E_j|$ if $i<j$). Then, for a new $N$ of interest where $N \notin \{|E_1|, ..., |E_k|\}$ and $|E_1|<N<|E_k|$, iBoN constructs an intermediate circuit by augmenting the best available smaller circuit with additional high-scoring edges from a larger one that are not already included.


\subsection{BoN with Constraint-Adaptive Score Matrix}
\label{subsec: bon-csm}
BoN with Constraint-adaptive Score Matrix (BoN-CSM, Algorithm~\ref{alg:bon-csm}) leverages all $k$ previously discovered circuits $\{E_1, ..., E_k\}$ of different edge budgets (i.e., constraints) to establish a score matrix $S$ and a tier matrix $T$, which are then used to efficiently form new circuits. It first initializes $S$, $T$, and an auxiliary index matrix $B$. Starting from the smallest circuit $E_1$, it iteratively records each edge $e$’s score $a_e$ to $S$ and the current circuit index (e.g., $i$ for $E_i$) to $T$, while using $B$ to avoid duplicate entries. In this way, $S$ and $T$ determine the importance scores and priorities of all edges that have been identified in $\{E_1, \dots, E_k\}$. When constructing a new circuit of size $N$ where $|E_1|<N<|E_k|$, it first sorts all edges in $S$ by their tiers in $T$ to prioritize those from smaller (i.e., high-tier) circuits. It further sorts the edges within each tier by their importance scores in $S$. Then, it selects top-$N$ edges from this tier-then-score order to form the circuit, requiring no additional LLM forward pass.
\section{Experiments}
\label{sec: exp}
\begin{figure*}[tb]
  \centering
  \begin{subfigure}{0.3\linewidth}
    \includegraphics[width=\textwidth]{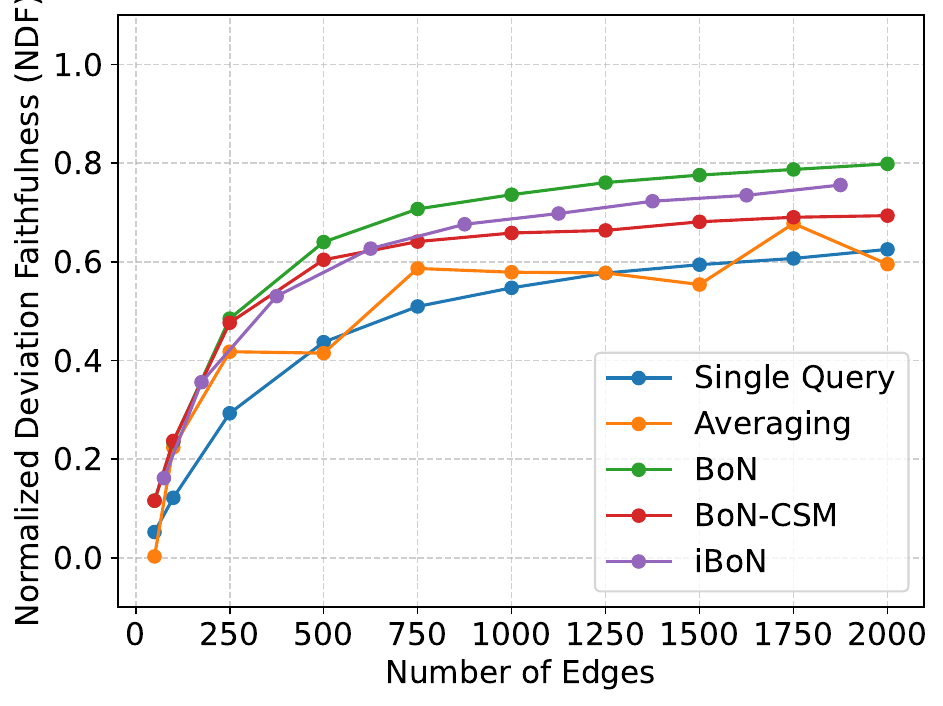}
    \caption{IOI dataset.}
    \label{subfig: main-results-ioi}
  \end{subfigure}
  \hfill
  \begin{subfigure}{0.3\linewidth}
    \includegraphics[width=\textwidth]{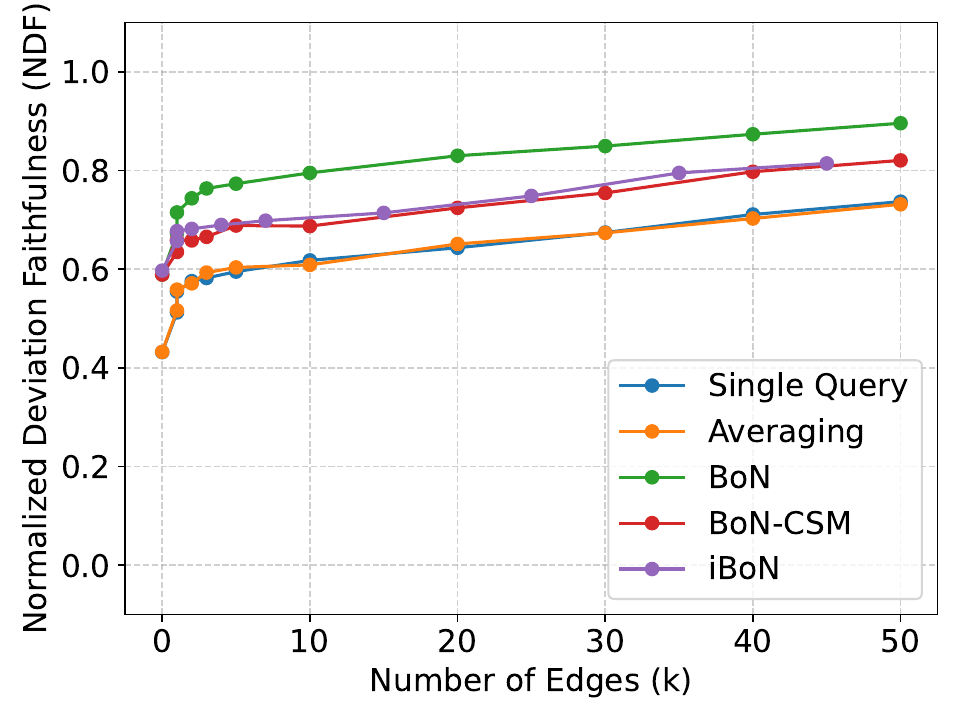}
    \caption{Arithmetic addition.}
    \label{subfig: main-results-arithmetic-add}
  \end{subfigure}
  \hfill
  \begin{subfigure}{0.3\linewidth}
    \includegraphics[width=\textwidth]{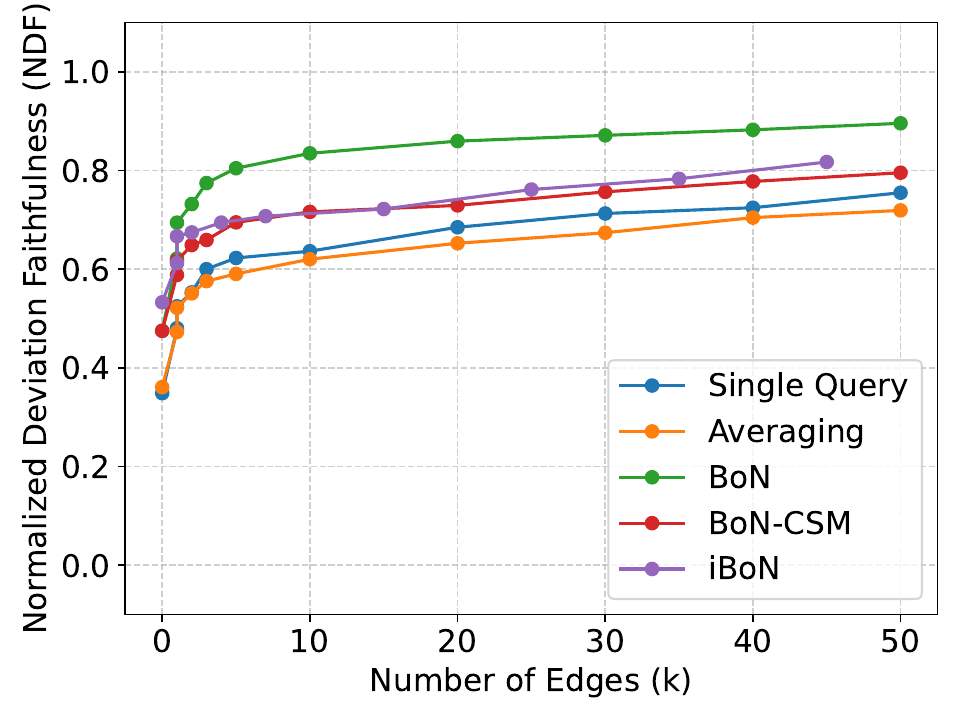}
    \caption{Arithmetic multiplication.}
    \label{subfig: main-results-arithmetic-mul}
  \end{subfigure}
  \hfill
  \begin{subfigure}{0.3\linewidth}
    \includegraphics[width=\textwidth]{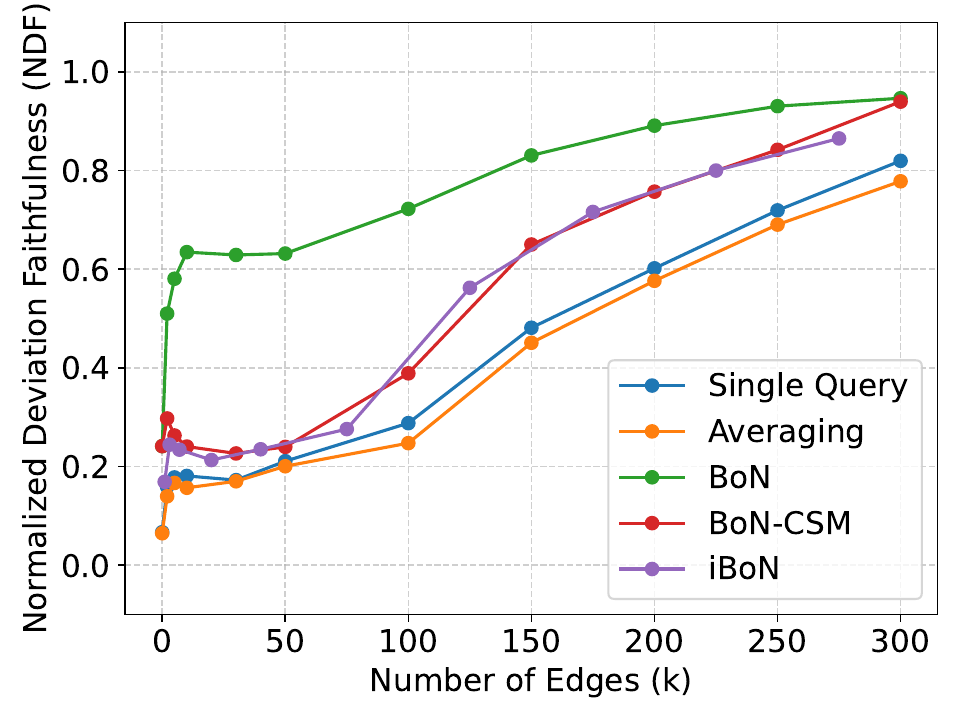}
    \caption{MMLU Marketing.}
    \label{subfig: main-results-mmlu-marketing}
  \end{subfigure}
  \hfill
  \begin{subfigure}{0.3\linewidth}
    \includegraphics[width=\textwidth]{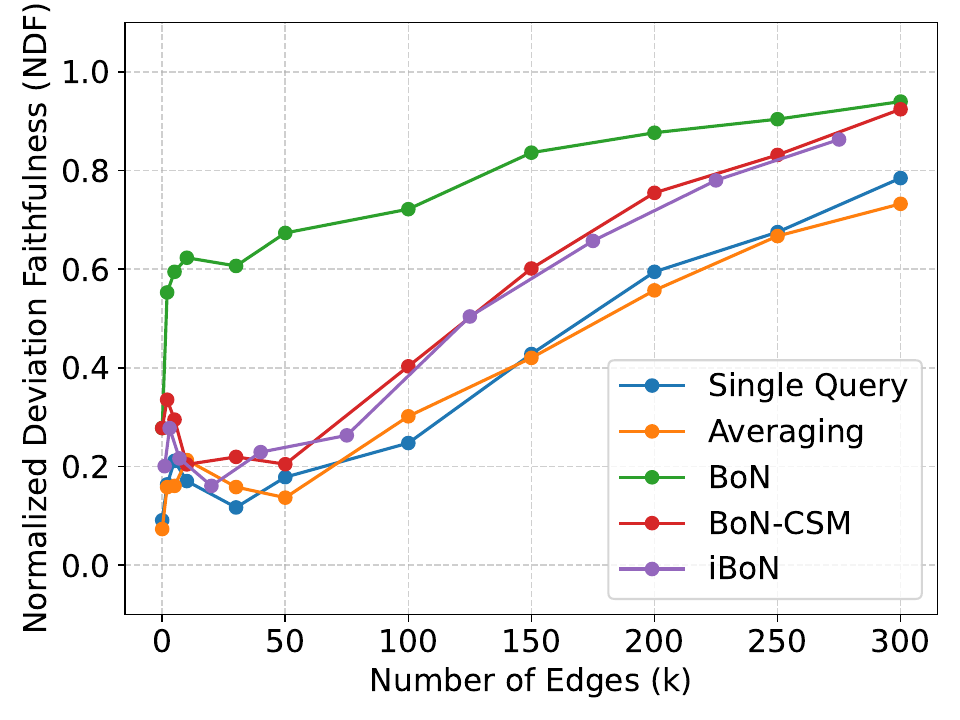}
    \caption{MMLU Astronomy.}
    \label{subfig: main-results-mmlu-astronomy}
  \end{subfigure}
  \hfill
  \begin{subfigure}{0.3\linewidth}
    \includegraphics[width=\textwidth]{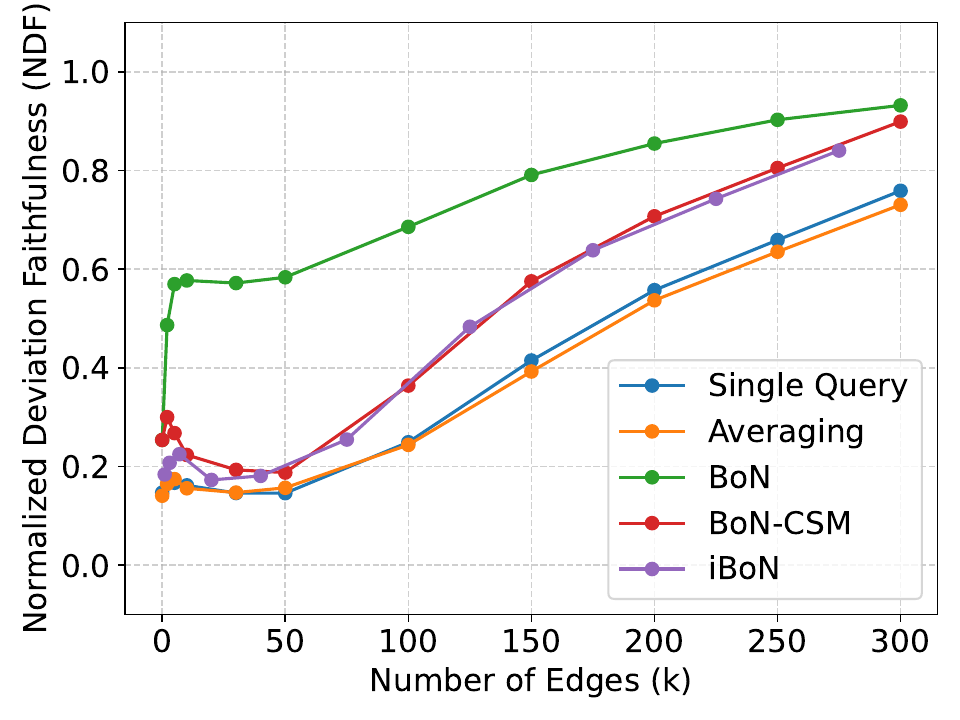}
    \caption{ARC Challenge.}
    \label{subfig: main-results-arc-challenge}
  \end{subfigure}
  \caption{\textbf{Main results of BoN sampling for query circuit discovery.} BoN substantially outperforms all other methods. Although iBoN and BoN-CSM, two fast approximations to BoN, perform worse than it, they still clearly exceed both baselines.}
  \label{fig: main results}
\vspace{-0.5em}
\end{figure*}
\subsection{Experimental Setup}
\label{subsec: experimental setup}
We conduct query circuit discovery with BoN sampling on IOI~\citep{wang2023interpretability}, arithmetic addition, arithmetic multiplication, ARC Challenge~\citep{allenai:arc}, and nine categories of MMLU~\citep{hendrycks2021measuring}. Performances of circuits are averaged over all queries in the datasets. For each IOI query, we randomly select nine other queries from the dataset as its paraphrases (i.e., $p=9$). For arithmetic addition and multiplication, paraphrases are produced by permuting the operands. For ARC Challenge and MMLU, we use GPT-4o~\citep{hurst2024gpt} to generate nine paraphrases of the question stem. We adopt EAP-IG (step $m=20$) as the backbone method to estimate edge scores and use greedy selection to construct edges. We consider two baselines: (i) estimate each edge's importance score $a_e$ simply on that query; and (ii) estimate $a_e$ as the average over the query and its paraphrases. Unless otherwise specified, we adopt GPT-2 Small as the target LLM for IOI and Llama-3.2-1B-Instruct for all other tasks. Refer to Appendix~\ref{sup: experimental details} for detailed experimental setup and design choices and Appendix~\ref{sup: more experimental results} for additional experiments.

\subsection{Main Results}
\begin{figure*}[tb]
  \centering
  \begin{subfigure}{0.3\linewidth}
    \includegraphics[width=\textwidth]{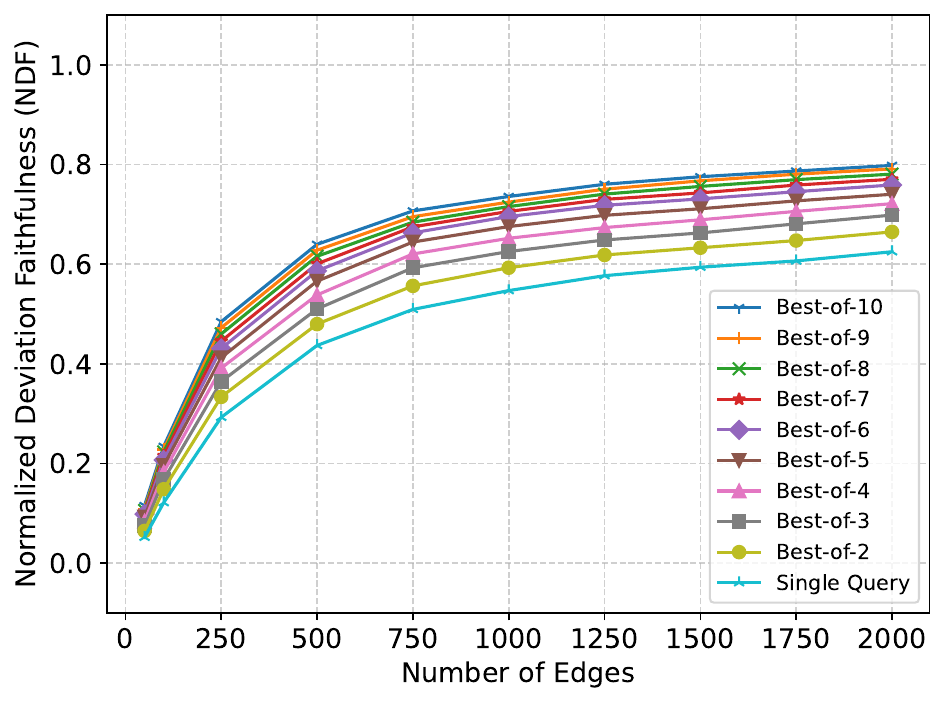}
    \caption{IOI dataset.}
    \label{subfig: ioi-para-num}
  \end{subfigure}
  \hfill
  \begin{subfigure}{0.3\linewidth}
    \includegraphics[width=\textwidth]{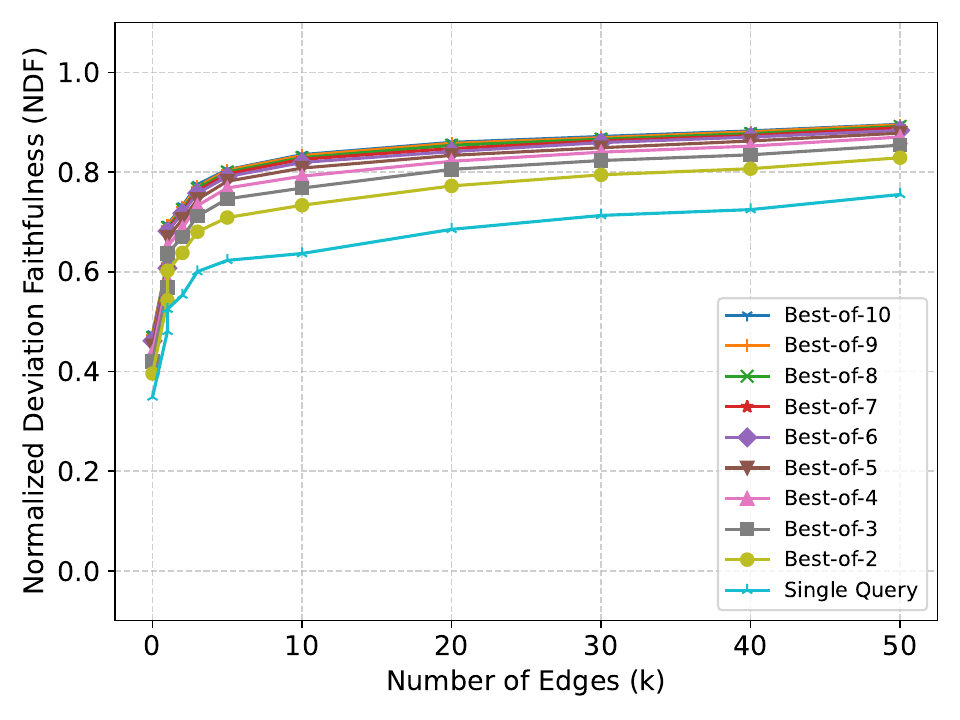}
    \caption{Arithmetic multiplication.}
    \label{subfig: arithmetic-mul-para-num}
  \end{subfigure}
  \hfill
  \begin{subfigure}{0.3\linewidth}
    \includegraphics[width=\textwidth]{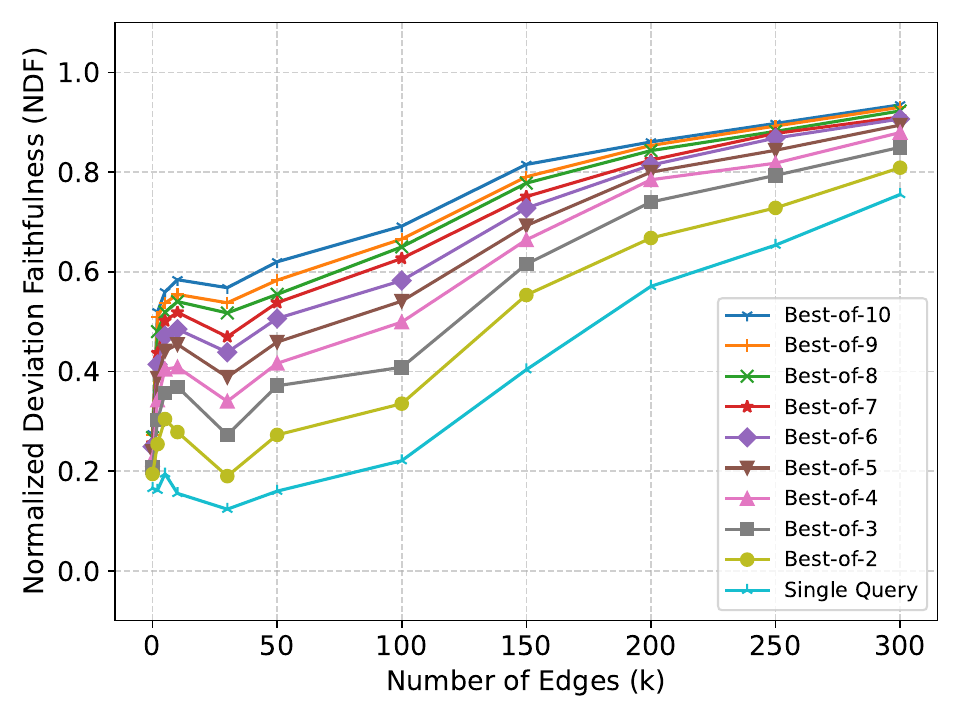}
    \caption{MMLU Astronomy.}
    \label{subfig: mmlu-astronomy-para-num}
  \end{subfigure}
  \caption{\textbf{Performance of BoN sampling with different numbers of paraphrases}. As BoN selects the most faithful circuit, its performance exhibits monotonically increasing yet diminishing returns.}
  \label{fig: ablation_study_num_of_para}
\vspace{-1em}
\end{figure*}
\begin{figure*}[tb]
  \centering
  \includegraphics[width=0.7\textwidth]{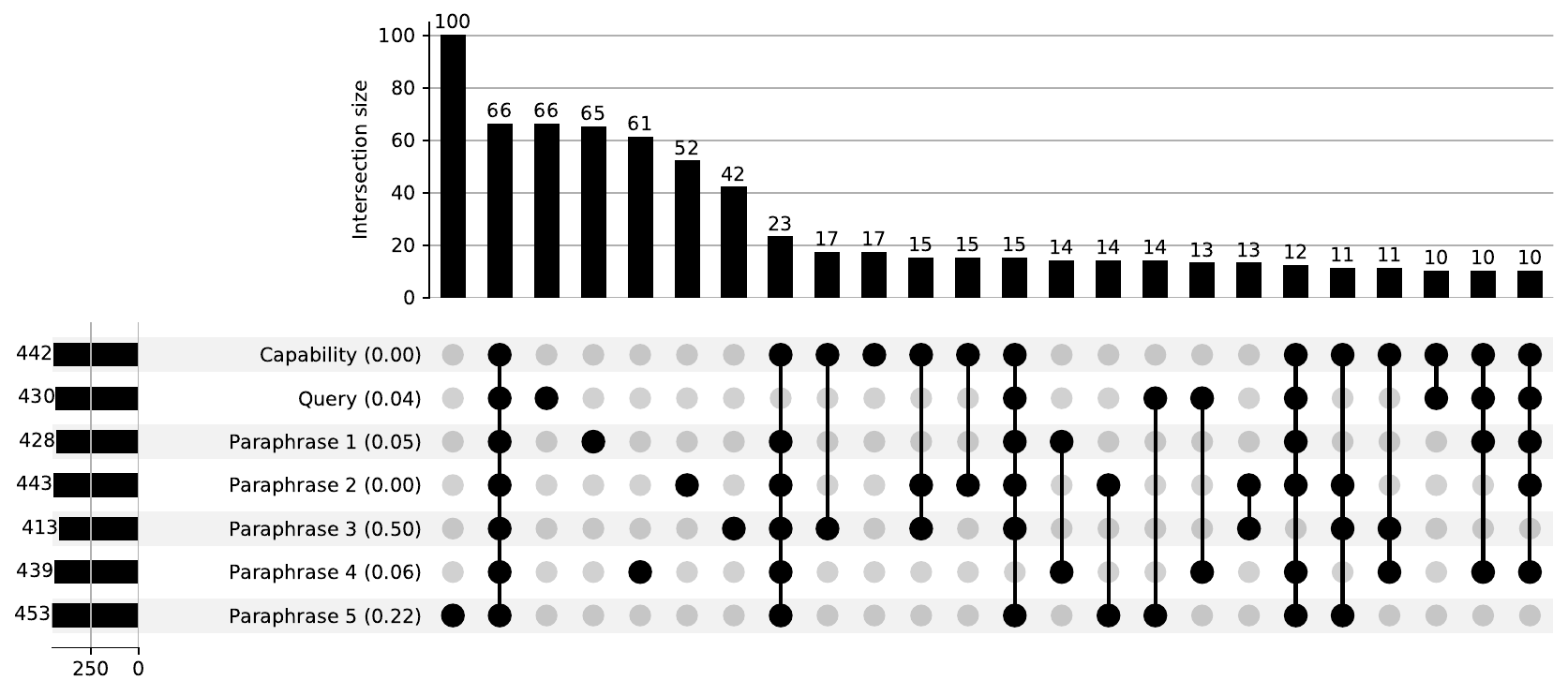}
  \caption{\textbf{UpSet plot of the capability circuit and query circuits of a randomly selected query discovered by it or its paraphrases}. A total of 66 edges are shared across all circuits.}
  \label{fig: upset plot}
\end{figure*}
Figure~\ref{fig: main results} presents the results of query circuit discovery across different tasks (complete MMLU results are in Appendix~\ref{supsub: complete mmlu results}). BoN, iBoN, and BoN-CSM consistently outperform both baseline (i) (\textit{Single Query}) and baseline (ii) (\textit{Averaging}). In particular, BoN surpasses other methods by a large margin, requiring orders of magnitude fewer edges to construct non-trivial query circuits. In MMLU, a circuit with only $5k$ edges (1.3\% of Llama-3.2-1B-Instruct’s all edges) achieves an average NDF of 0.6, whereas vanilla EAP-IG (single-query baseline) suggests that $200k$ (51.7\%) edges are needed to reach that level. The results advance recent findings of input-dependent activation sparsity~\citep{li2023the, szatkowski2025universal} to circuitry sparsity, demonstrating the promise of finding compact, critical information flow within an LLM responsible for answering an input query. Notably, the averaging method does not perform better than the single-query one. This is potentially because, while prioritizing edges scored high on both the input query and its paraphrases, it may simultaneously downweight edges that are crucial only to the original query.

\subsection{Ablation Study on Numbers of Paraphrases}
Figure~\ref{fig: ablation_study_num_of_para} shows the performance of BoN with different numbers of paraphrases. Since BoN selects the most faithful circuit, its performance increases monotonically as the number of paraphrases grows. However, the gains diminish because additional paraphrases often provide overlapping or redundant information, making it less likely for them to contribute new high-quality circuits.

\subsection{Circuit Variances and Shared Sub-circuit}
Using IOI, we further examine the relationship between query and capability circuit (IEs averaged over 1000 queries). Specifically, we study whether BoN sampling (1) just randomly picks circuits which coincidentally output the correct token, or (2) discovers variants of a common mechanism that preserve a shared set of critical edges (i.e., shared sub-circuit), regardless of how the query is phrased.

Figure~\ref{fig: upset plot} provides preliminary support for (2). It shows the UpSet plot of the capability circuit and the query circuits of a randomly selected query (edge budget $N=500$). Each row corresponds to a circuit, with its number of edges and NDF score; each column indicates the number of edges shared among the circuits in black dots. Notably, 66 edges appear in all circuits, supporting the existence of a shared sub-circuit. We also observe 23 edges (8th column) missing only from the circuit derived from the original query—meaning that relying solely on the original query would fail to recover these edges. Additional evidence for (2) is in Appendix~\ref{supsub: shared circuit}.

\subsection{Query Circuits with Human-Readable Features}
\label{subsec: qcwithsae-main-paired}
\begin{table*}[tb]
\caption{\textbf{Model bias reduction by gender-feature ablation on the best- and worst-performing circuits (out of 10 per query).} We report one-sided Wilcoxon signed-rank $p$-values and Rosenthal's $r$ as effect size. Each sample pair consists of the best and worst circuit for the same query, evaluated by NDF. Baseline probability bias before steering: $|P(\text{he}) - P(\text{she})| = 0.542 \pm 0.031$.}
\label{tab: bias-reduction-paired}
\centering
\setlength{\tabcolsep}{6pt}
\renewcommand{\arraystretch}{1.2}
\resizebox{0.7\linewidth}{!}{%
\begin{tabular}{@{}p{3.2cm} l l c c c@{}}
\toprule
\textbf{Metric} & \textbf{Scale} & \textbf{Circuit} & \textbf{Mean $\pm$ Std} & \textbf{$p$-value} & \textbf{Rosenthal's $r$} \\
\midrule
\multirow{6}{3.2cm}{\textbf{Absolute Bias Reduction}}
& \multirow{2}{*}{Logit}
  & Best  & $0.810 \pm 0.581$ & \multirow{2}{*}{$<\!0.0001$} & \multirow{2}{*}{0.787} \\
\cmidrule{3-4}
&  & Worst & $0.234 \pm 0.278$ & & \\
& \multicolumn{5}{l}{\quad $\Delta\text{Mean} = +0.576$ \quad \textbf{(****)}} \\
\cmidrule{2-6}
& \multirow{2}{*}{Probability}
  & Best  & $0.063 \pm 0.057$ & \multirow{2}{*}{$<\!0.0001$} & \multirow{2}{*}{0.737} \\
\cmidrule{3-4}
&  & Worst & $0.011 \pm 0.020$ & & \\
& \multicolumn{5}{l}{\quad $\Delta\text{Mean} = +0.052$ \quad \textbf{(****)}} \\
\midrule
\multirow{6}{3.2cm}{\textbf{Avg.\ Bias Reduction per Gender Feature}}
& \multirow{2}{*}{Logit}
  & Best  & $0.073 \pm 0.055$ & \multirow{2}{*}{$<\!0.0001$} & \multirow{2}{*}{0.836} \\
\cmidrule{3-4}
&  & Worst & $0.014 \pm 0.017$ & & \\
& \multicolumn{5}{l}{\quad $\Delta\text{Mean} = +0.059$ \quad \textbf{(****)}} \\
\cmidrule{2-6}
& \multirow{2}{*}{Probability}
  & Best  & $0.006 \pm 0.005$ & \multirow{2}{*}{$<\!0.0001$} & \multirow{2}{*}{0.803} \\
\cmidrule{3-4}
&  & Worst & $0.001 \pm 0.001$ & & \\
& \multicolumn{5}{l}{\quad $\Delta\text{Mean} = +0.005$ \quad \textbf{(****)}} \\
\bottomrule
\end{tabular}}
\end{table*}
Finally, we examine equipping query circuits with SAEs to acquire natural language explanations and test whether our proposed NDF, as a proxy for circuit quality, can reliably estimate how actionable a circuit is for feature-level steering.

With GPT-2 Small as the target model and the Gender Bias dataset~\citep{NEURIPS2020_92650b2e}, we first select queries on which the model exhibits strong bias, defined as a probability difference greater than 0.5 between the stereotypical and anti-stereotypical genders:
\begin{equation}
P(\text{stereotypical}) - P(\text{anti-stereotypical}) > 0.5.
\end{equation}
This yields 50 out of 986 samples. For each of these samples, we generate 9 additional paraphrases and apply EAP-IG to each, producing 10 query circuits per sample. We retain only samples whose best circuit (i.e., the circuit with the highest NDF) achieves NDF $> 0.8$, resulting in 32 samples.

For these 32 samples, we measure bias reduction after zeroing out gender-related SAE features identified in the best and worst circuits (Table~\ref{tab: bias-reduction-paired}), using the OpenAI-released SAE suite for GPT-2 Small~\citep{gao2025scaling}. Across all four metric--scale combinations, the best circuits consistently achieve statistically larger bias reduction than the worst circuits (all values averaged over 32 samples). Under the logit 
scale, the best circuits yield an absolute bias reduction of $0.810 \pm 0.581$, compared 
to $0.234 \pm 0.278$ for the worst circuits ($\Delta\text{Mean} = +0.576$); analogous trends 
hold for both the probability scale and the average bias reduction per feature ablated. One-sided Wilcoxon 
signed-rank tests confirm that all differences are highly significant ($p < 0.0001$), 
with large effect sizes (Rosenthal's $r \in [0.737, 0.836]$). Together, these results 
suggest that a higher NDF, i.e., circuit faithfulness, indicates greater actionability—circuits more faithfully capture the model's 
underlying computations to process the queries. Additional quantitative experiments and a qualitative case study are provided in Appendix~\ref{supsubsec: qcwithsae}.
\section{Conclusion}
\label{sec: conclusion}
We introduce query circuit discovery, the task of identifying the information flow within an LLM responsible for answering an input query. We formalize the task, distinguish it from capability circuit discovery, identify its technical challenges, and introduce methods to tackle them. In particular, we propose NDF as a more reliable metric for evaluating circuit faithfulness and BoN sampling as a simple technique for discovering faithful query circuits. Experiments reveal compact sub-networks within the model that recover much of its performance even for complex queries, establishing BoN as a useful method and query circuit discovery as a promising direction for explaining LLM decisions.

\section*{Impact Statement}
\label{sec: ethics statement}
This paper presents work whose goal is to advance the field of Machine Learning. This paper adheres to the ICML Code of Conduct and contains no confidential data, sensitive content, or experiments involving human subjects. We note that mechanistic interpretability methods, including ours, should be used with caution to avoid incorrect interpretations that could lead to adverse consequences.


\bibliography{reference}
\bibliographystyle{icml2026}
\clearpage
\appendix
\onecolumn
\renewcommand{\thetable}{A\arabic{table}}
\renewcommand{\thefigure}{A\arabic{figure}}
\renewcommand{\thealgorithm}{A\arabic{algorithm}}
\renewcommand*\contentsname{Supplementary Material}
\clearpage
\section{Detailed Experimental Setup}
\label{sup: experimental details}
This section serves as an extension of Section~\ref{sec: exp} to provide a more detailed experimental setup, design choices, and their implications.
\paragraph{Datasets.}We conduct query circuit discovery on the IOI dataset, arithmetic addition, arithmetic multiplication, ARC Challenge, and nine categories from MMLU. We randomly select the nine categories from all 52 categories in which Claude 3.5 Sonnet (2024/10/22 version)~\citep{claude35sonnet} achieves at least 95\% accuracy on the Stanford HELM MMLU leaderboard~\citep{liang2023holistic}. The IOI dataset follows \citet{hanna2024have}'s implementation and has 1000 queries. The arithmetic addition and multiplication datasets each consist of 500 queries, covering operands of length 2–5 (125 queries per length). Each query’s answer is an integer less than 1000. These datasets are more challenging than the two-operand arithmetic addition and subtraction tasks used in the MIB benchmark. For ARC Challenge, we adopt the test split (1172 MCQs). The nine selected MMLU categories are: marketing (234 MCQs), professional medicine (272 MCQs), astronomy (152 MCQs), college biology (144 MCQs), high school computer science (100 MCQs), logical fallacies (163 MCQs), nutrition (306 MCQs), international law (121 MCQs), and management (103 MCQs).

\paragraph{Paraphrases.}For each IOI query, we randomly select nine other queries from the dataset as paraphrases since every query in the IOI dataset is itself a word-swapped variant of another. In arithmetic addition and multiplication, paraphrases are generated by permuting the operands. The number of available paraphrases varies with the number of operands, but we limit each query to at most nine paraphrases. For ARC Challenge and MMLU, we use GPT-4o to generate nine paraphrases of the question stem.

\begin{table}[tb]
  \caption{Examples of original and corrupted queries from the datasets used in this paper. For arithmetic questions, the corrupted query has the same number of operands but a different answer. For MCQs, the corrupted query preserves the options, but the question stem is replaced with a prompt that simply asks the model to choose one.}
  \label{tab: clean corrupted query comparison}
  \centering
  \addtolength{\tabcolsep}{4pt}
  \begin{tabular}{@{}p{0.1\textwidth}|p{0.42\textwidth}|p{0.42\textwidth}@{}}
    \toprule
    Dataset & Clean Query & Corrupted Query\\
    \midrule
    \makecell[l]{\parbox{5.5cm}{IOI}} & \makecell[l]{When Amy and Laura got a snack at\\ the house, Laura decided to give it to} & \makecell[l]{When Amy and Laura got a snack at\\ the house, Nicholas decided to give it to}\\
    \midrule
    \makecell[l]{\parbox{5.5cm}{Arithmetic\\Add.}}  & \makecell[l]{41+260+303+48+87=} & \makecell[l]{11+52+23+18+6=}\\
    \midrule
    \makecell[l]{\parbox{5.5cm}{Arithmetic\\Mul.}}  & \makecell[l]{7*2*2*3*10=} & \makecell[l]{2*2*14*4*4=}\\
    \midrule
    \makecell[l]{\parbox{5.5cm}{MMLU}}  & \makecell[l]{What is true for a type-Ia (""type one-a"") \\supernova?\\(A) This type occurs in binary systems.\\
(B) This type occurs in young galaxies.\\
(C) This type produces gamma-ray bursts.\\
(D) This type produces high amounts of \\X-rays.\\
Answer: (} & \makecell[l]{Which is the most possible answer?\\(A) This type occurs in binary systems.\\
(B) This type occurs in young galaxies.\\
(C) This type produces gamma-ray bursts.\\
(D) This type produces high amounts of \\X-rays.\\
Answer: (}\\
    \midrule
    \makecell[l]{\parbox{5.5cm}{ARC\\Challenge}} & \makecell[l]{Two girls are pulling on opposite ends of a \\thick rope. Both girls pull on the rope with \\the same force but in opposite directions. \\If both girls continue to pull with the same \\force, what will most likely happen?\\
(A) One girl will pull the other toward her.\\
(B) Both girls will stay in the same place.\\
(C) Gravity will cause the rope to sag.\\
(D) The rope will break.\\
Answer: (} & \makecell[l]{Which is the most possible answer?\\
(A) One girl will pull the other toward her.\\
(B) Both girls will stay in the same place.\\
(C) Gravity will cause the rope to sag.\\
(D) The rope will break.\\
Answer: (}\\
    \bottomrule
  \end{tabular}
\end{table}
\paragraph{Corrupted queries.} For IOI, a corrupted query is constructed by replacing the repeated name in the original query with a third name, as described in Section~\ref{subsubsec: score edge and construct circuit}. For arithmetic addition and multiplication, the corrupted query is another sample with the same number of operands but a different answer. For ARC Challenge and MMLU, the corrupted query is created by replacing the question stem with \textit{``Which is the most possible answer?''}. Table~\ref{tab: clean corrupted query comparison} shows the examples of original and corrupted queries. 

Note that the form of corrupted queries directly affects both the functionality and interpretation of the discovered circuits. For MCQs, under our proposed corruption strategy, the discovered edges capture critical interactions between the stem and the choices. This arises because such interactions are present in the original query but absent in the corrupted one. By contrast, the MIB benchmark, as an early attempt at circuit discovery for MCQs, constructs corrupted queries through semantics-irrelevant rephrasing—for example, changing option IDs from (A), (B), (C), (D) to (1), (2), (3), (4). Under this formulation, the discovered circuits primarily contain edges associated with ID matching rather than meaningful stem–choice reasoning and factual retrieval.

\paragraph{Baselines.} We adopt EAP-IG as the backbone method to score edges since it is one of the most effective current approaches. The original EAP-IG implementation employs a Dijkstra-like iterative construction introduced in Section~\ref{subsubsec: score edge and construct circuit}. Our replications show that it achieves comparable performance to greedy selection but with higher runtime (see Appendix~\ref{supsub: comp-greedy-dijkstra}). As a result, we adopt greedy selection to construct the circuit after edge scoring. The IG step is set to 20 throughout the experiments. Two baselines are: (i) applying EAP-IG directly to each original query, i.e., estimating each edge’s IE on that query; and (ii) applying EAP-IG to estimate each edge’s IE averaged over the original query and its paraphrases. The latter is exactly the way methods in capability circuit discovery score edges for capability circuit construction.

\paragraph{Target Model and Performance Metric.} For IOI, we use GPT-2 Small (32491 edges) as the target LLM, following prior work; for all other tasks, we use Llama-3.2-1B-Instruct (386713 edges). For performance metric $L$, we adopt logit difference as it provides a more natural unit for transformers than probability difference~\citep{heimersheim2024use}. Specifically, for IOI, the logit difference between the correct and incorrect name is adopted. For arithmetic addition and multiplication, the logit difference between the correct and corrupted answers is used. For ARC Challenge and MMLU, we consider the logit difference between the correct option and the average of the incorrect ones. Performances of circuits are averaged over all queries in the datasets.

\paragraph{Edge budgets.} When testing all methods except iBoN: For IOI, we set $N\in \{50, 100, 250, 500, 750, 1k, 1.25k, 1.5k, 1.75k, 2k\}$; For arithmetic addition and arithmetic multiplication, we use $N\in \{500, 1k, 1.5k, 2k, 3k, 5k, 10k, 20k, 30k, 40k, 50k\}$; For ARC Challenge and MMLU, we consider $N\in \{500, 2k, 5k, 10k, 30k, 50k, 100k, 150k, 200k, 250k, 300k\}$. When testing iBoN, we adopt interpolated budgets because iBoN will produce the same performance as BoN if it has the same edge budget. Specifically, for iBoN on IOI, we set $N\in\{75,175,375,625,875,1.125k,1.375k,1.625k,1.875k\}$; For arithmetic addition and arithmetic multiplication, we use $N\in\{750,1.25k,1.75k,2.5k,4k,7.5k,15k,25k,35k,45k\}$; For ARC Challenge and MMLU, we consider $N\in\{1.25k,3.5k,7.5k,20k,40k,75k,125k,175k,225k,275k\}$.

\clearpage
\section{Detailed Algorithm}
\label{sup: algorithm}

\begin{algorithm}[tb]
\caption{iBoN}
\label{alg:ibon}
\begin{algorithmic}[1]
\INPUT Circuits (edge sets) $\{E_1, \dots, E_k\}$ with size in ascending order and edge number constraint $N$.\\
\OUTPUT An edge set (circuit) $E$.

\STATE Initialize $E$ as an empty edge set.
\STATE Find the largest $i$ such that $|E_i| < N$.
\STATE $E \gets E_i$.
\STATE $K \coloneqq N - |E|$.
\STATE $E^{1:K}_{i+1} \coloneqq$ top-$K$ edges of $E_{i+1}$ not in $E_i$.
\STATE $E \gets E \cup E^{1:K}_{i+1}$.
\STATE \textbf{return} $E$.
\end{algorithmic}
\end{algorithm}

\begin{algorithm}[tb]
\caption{BoN-CSM}
\label{alg:bon-csm}

\begin{algorithmic}[1]
\INPUT Circuits (edge sets) $\{E_1, \dots, E_k\}$ with size in ascending order.\\
\OUTPUT Score matrix $S$ and tier matrix $T$.

\STATE Initialize $S$, $T$, and boolean matrix $B$.
\FOR{$i, E_i$ in enumerate($\{E_1, \dots, E_k\}$)}
  \FOR{$e$ in $E_i$}
    \STATE $a_e \coloneqq$ attribution score of $e$.
    \STATE $(j, k) \coloneqq$ score matrix index of $e$.
    \IF{$B(j, k)$ is not \algorithmictrue}
      \STATE $S(j, k) \gets a_e$; $T(j, k) \gets i$.
      \STATE $B(j, k) \gets \algorithmictrue$.
    \ENDIF
  \ENDFOR
\ENDFOR
\STATE \textbf{return} $S$ and $T$.
\end{algorithmic}
\end{algorithm}

This section shows the detailed pseudocode of our iBoN and BoN-CSM methods. The former interpolates between two discovered circuits to form a new one of a given size. The latter forms the score matrix $S$ based on all discovered circuits to sample new circuits.
\clearpage
\section{More Experimental Results}
\label{sup: more experimental results}

\subsection{More Comparisons of Query Circuit Evaluation Results by NFS and NDF}
\label{supsub: nfs ndf complete comparison}
\begin{figure*}[tb]
  \centering
  \includegraphics[width=0.79\textwidth]{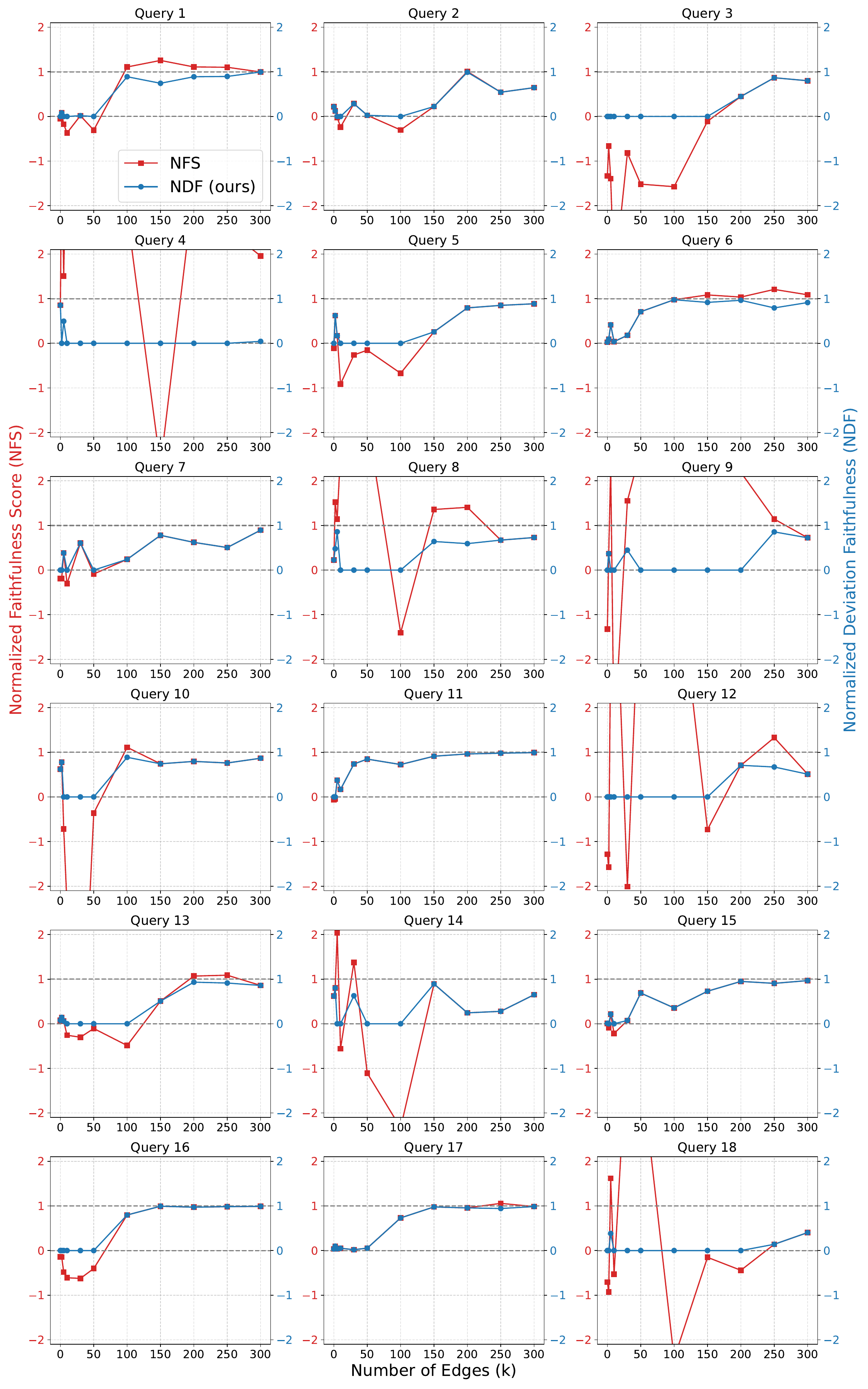}
  \caption{More query circuit evaluation results using NFS and our proposed NDF, which provides more stable evaluation and can better track the discovery progress as the circuit size grows.}
  \label{fig: nfs-ndf-comparison-full-resuls}
\end{figure*}
Figure~\ref{fig: nfs-ndf-comparison-full-resuls} presents 18 examples of query circuit evaluation using NDF and NFS. The queries are the first 18 samples from MMLU Marketing. These results show that our proposed NDF provides a more stable assessment and can track discovery progress as the circuits become larger.

\subsection{Examples of Score Matrix}
\label{supsub: more examples of score matrix}
\begin{figure*}[tb]
  \centering
  \includegraphics[width=0.98\textwidth]{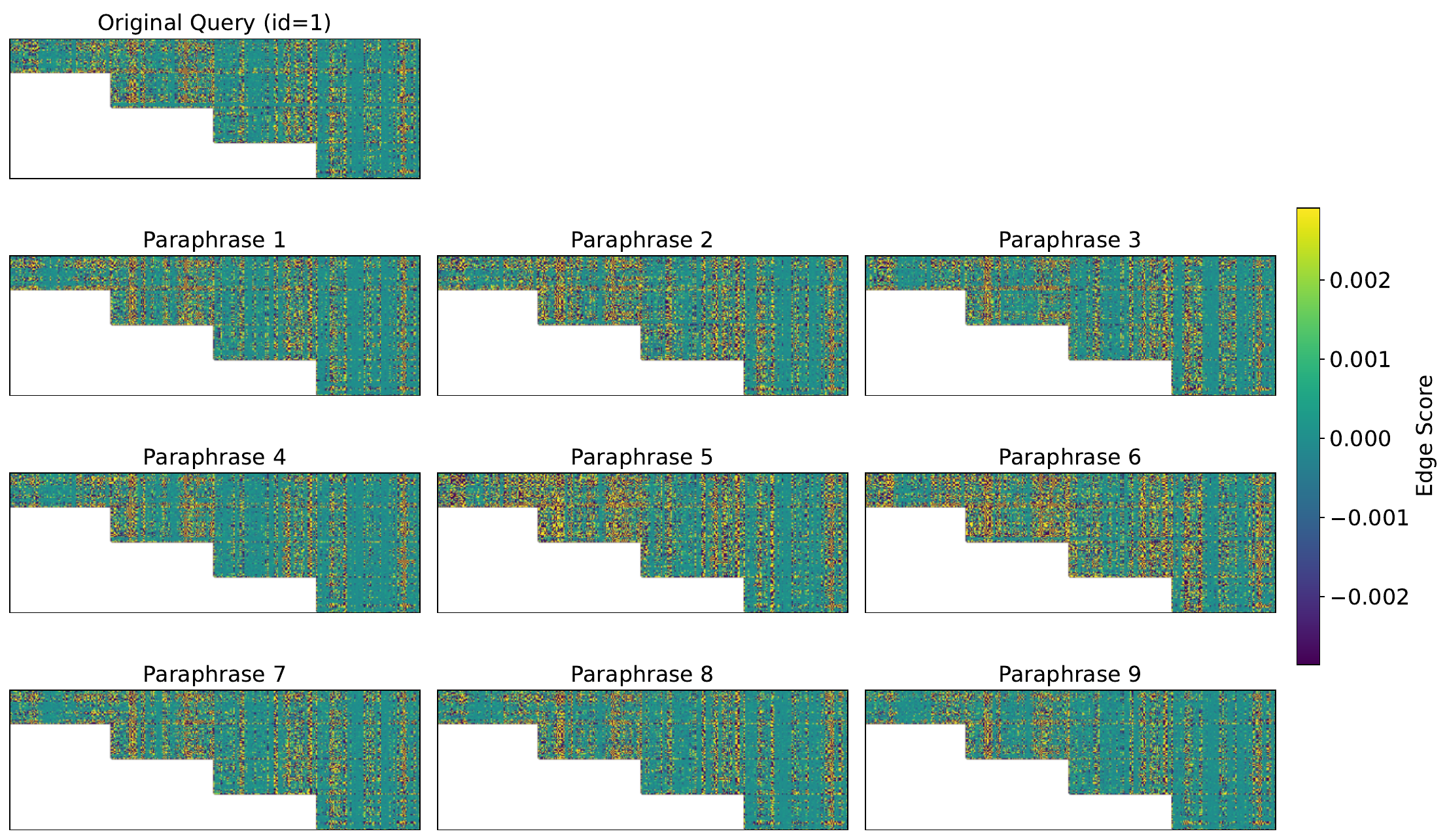}
  \caption{Edge score matrices of a query and its nine randomly selected paraphrases in IOI. EAP-IG is used to calculate the score of each edge (i.e., each entry in the matrix). These matrices share similar patterns.}
  \label{fig: score matrix ioi}
\end{figure*}
\begin{figure*}[tb]
  \centering
  \includegraphics[width=0.98\textwidth]{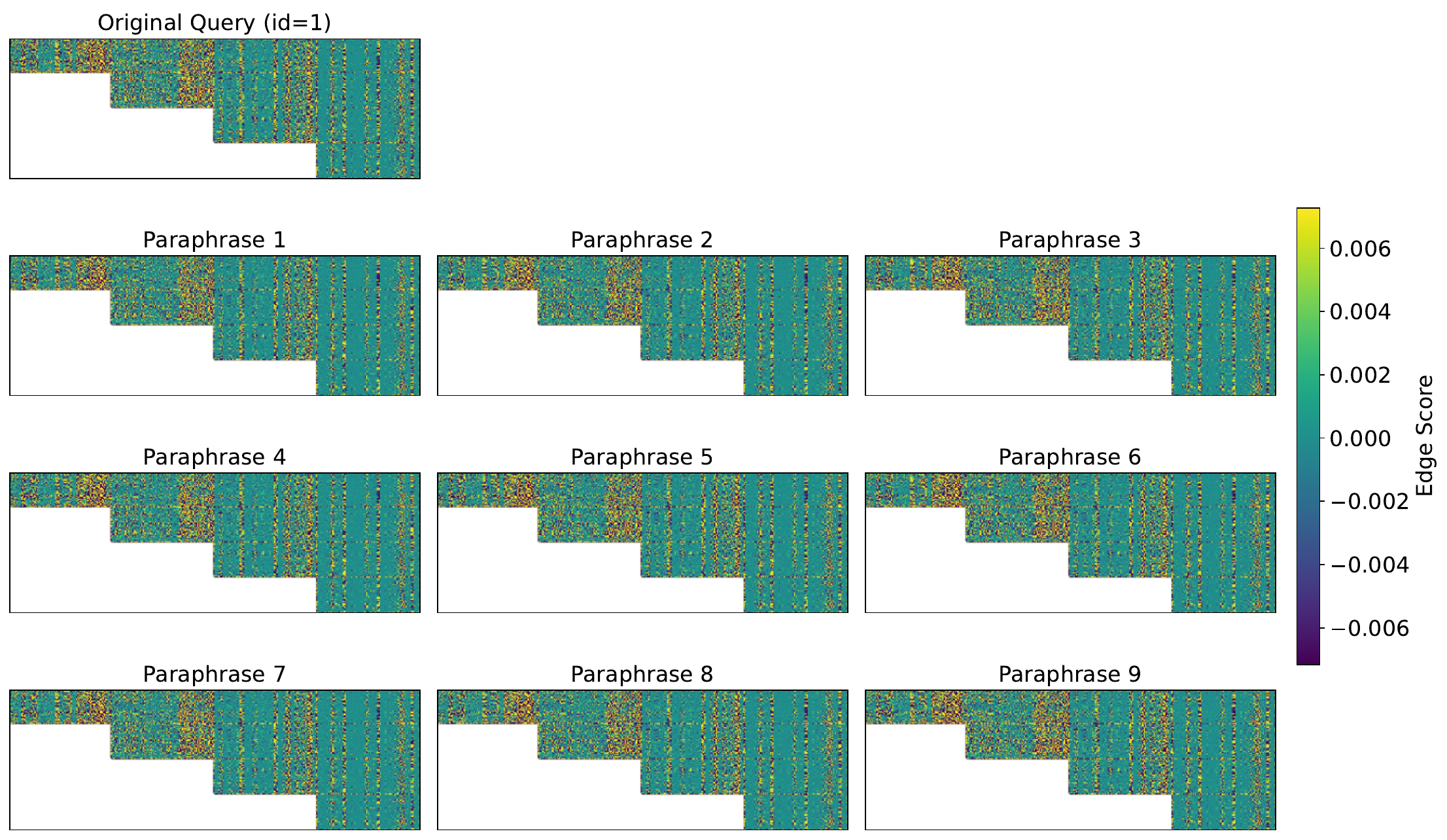}
  \caption{Edge score matrices of a query and its nine paraphrases in MMLU Astronomy. EAP-IG is used to calculate the score of each edge (i.e., each entry in the matrix). These matrices share similar patterns, and are dissimilar to those in Figure~\ref{fig: score matrix ioi}.}
  \label{fig: score matrix mmlu astronomy}
\end{figure*}
Figure~\ref{fig: score matrix ioi} and Figure~\ref{fig: score matrix mmlu astronomy} show the score matrices for the first query in the IOI dataset and MMLU Astronomy, along with their nine paraphrases. The target LLM is Llama-3.2-1B-Instruct. For clear visualization, we visualize only a subset of the score matrix—specifically, edges within layers 7–10 (out of the 16 layers). The discovery method is EAP-IG with step size $m=20$. The matrices are not square because, in practice, when parent nodes are attention heads, they will be split into query, key, and value heads by computing the gradients flowing through each.

The score matrices of the original query and its paraphrases exhibit similar patterns: the scores of certain edges remain high, while others are consistently low. On the other hand, score patterns between two different query types (IOI vs. MMLU MCQs) are more distinct. As argued and experimented in the main paper, the score pattern of a single query, though meaningful, is often not sufficiently precise for constructing faithful query circuits—motivating our exploration of paraphrase-based discovery to generate slightly different yet pattern-aligned score matrices.

\subsection{Complete Results of Nine MMLU Categories}
\label{supsub: complete mmlu results}
\begin{figure*}[tb]
  \centering
  \begin{subfigure}{0.32\linewidth}
    \includegraphics[width=\textwidth]{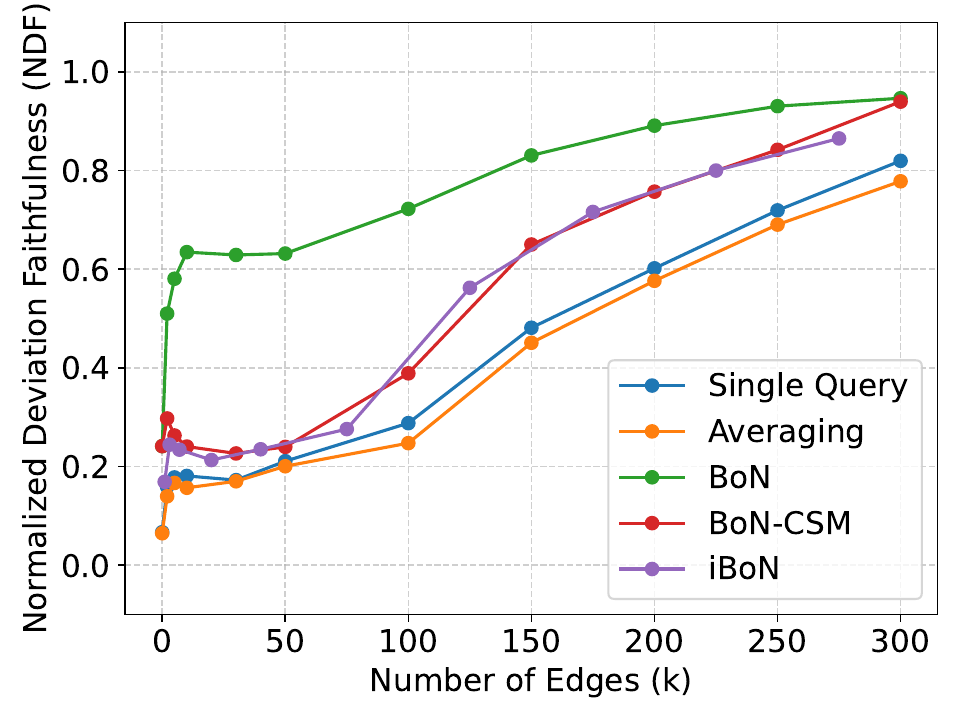}
    \caption{Marketing.}
  \end{subfigure}
  \hfill
  \begin{subfigure}{0.32\linewidth}
    \includegraphics[width=\textwidth]{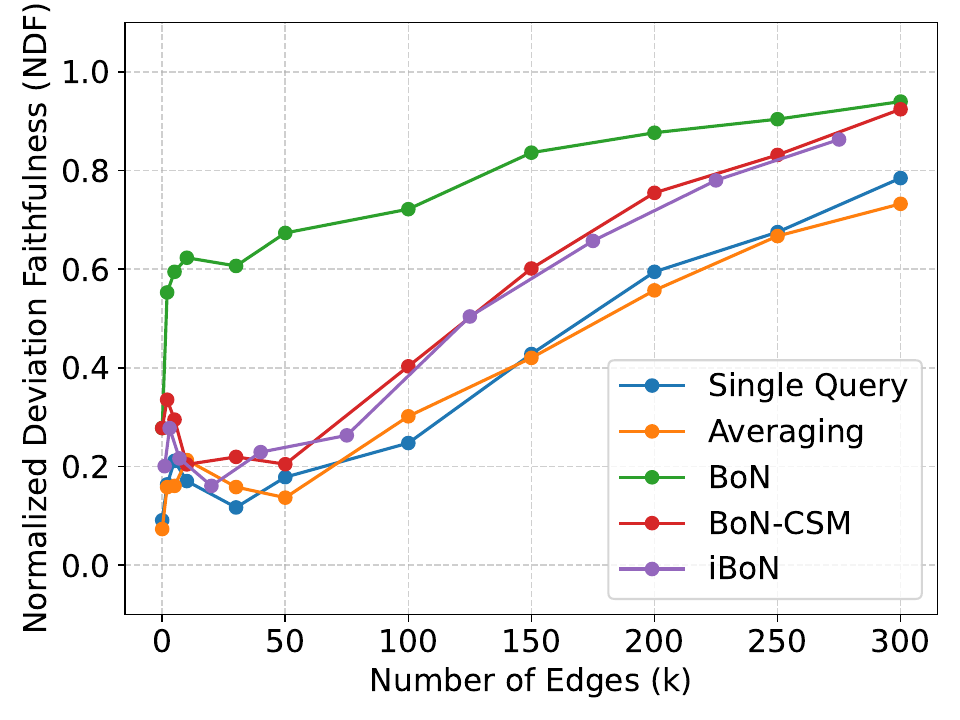}
    \caption{Astronomy.}
  \end{subfigure}
  \hfill
  \begin{subfigure}{0.32\linewidth}
    \includegraphics[width=\textwidth]{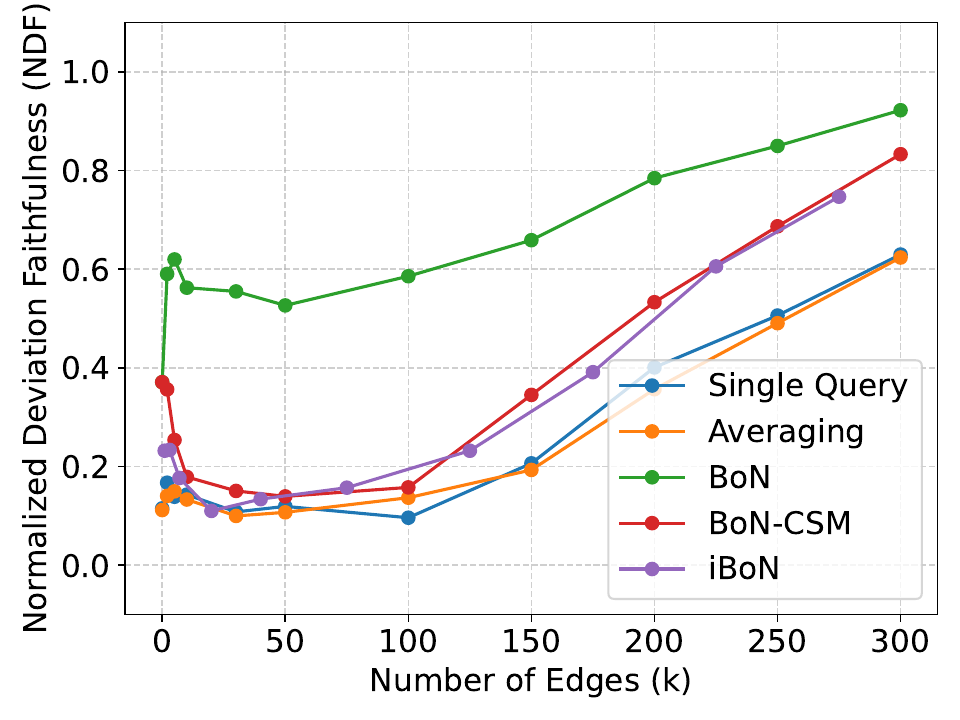}
    \caption{Professional Medicine.}
  \end{subfigure}
  \hfill
  \begin{subfigure}{0.32\linewidth}
    \includegraphics[width=\textwidth]{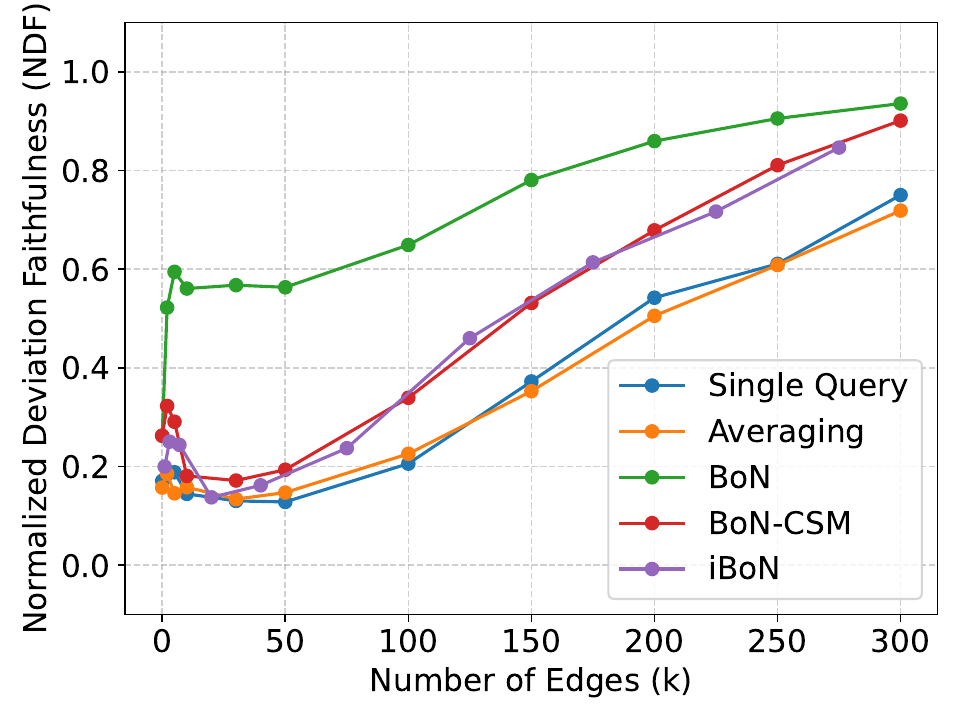}
    \caption{College Biology.}
  \end{subfigure}
  \hfill
  \begin{subfigure}{0.32\linewidth}
    \includegraphics[width=\textwidth]{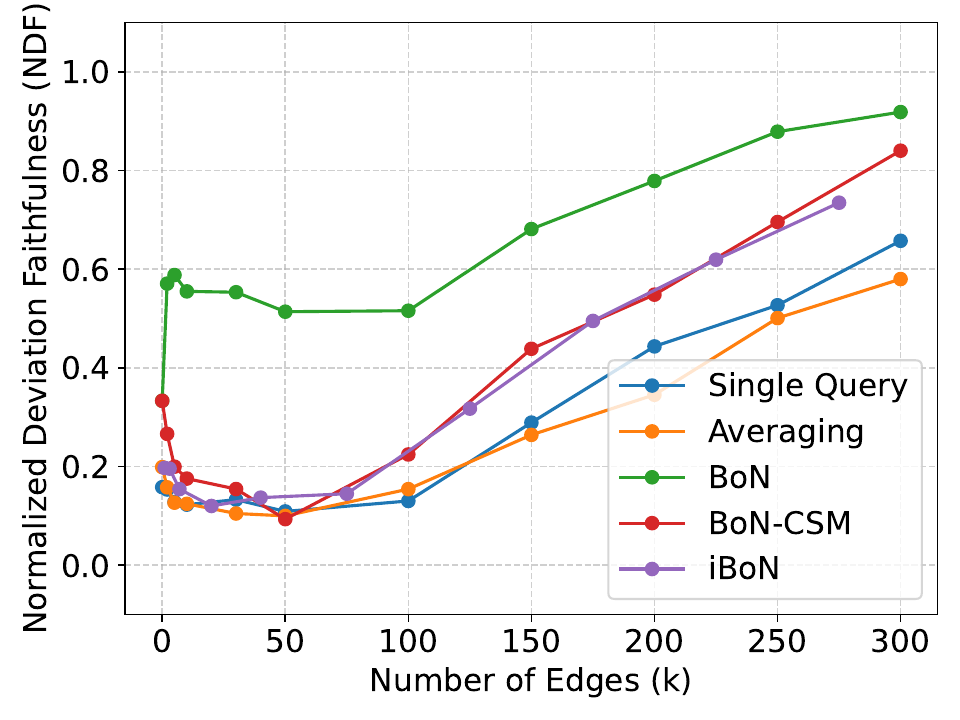}
    \caption{High School Computer Science.}
  \end{subfigure}
  \hfill
  \begin{subfigure}{0.32\linewidth}
    \includegraphics[width=\textwidth]{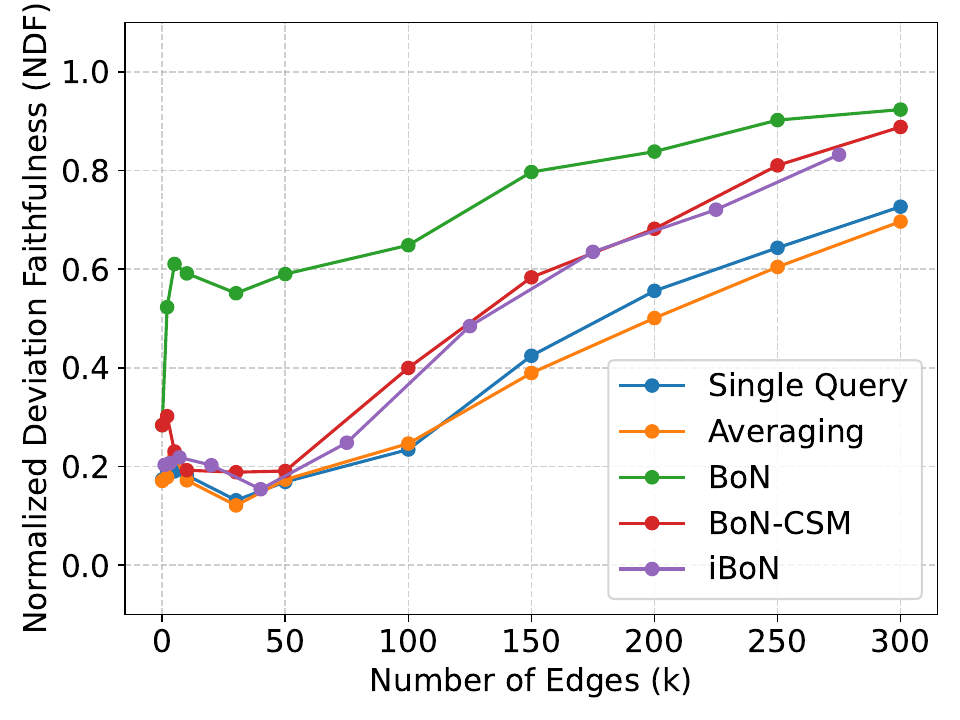}
    \caption{Logical Fallacies.}
  \end{subfigure}
  \hfill
  \begin{subfigure}{0.32\linewidth}
    \includegraphics[width=\textwidth]{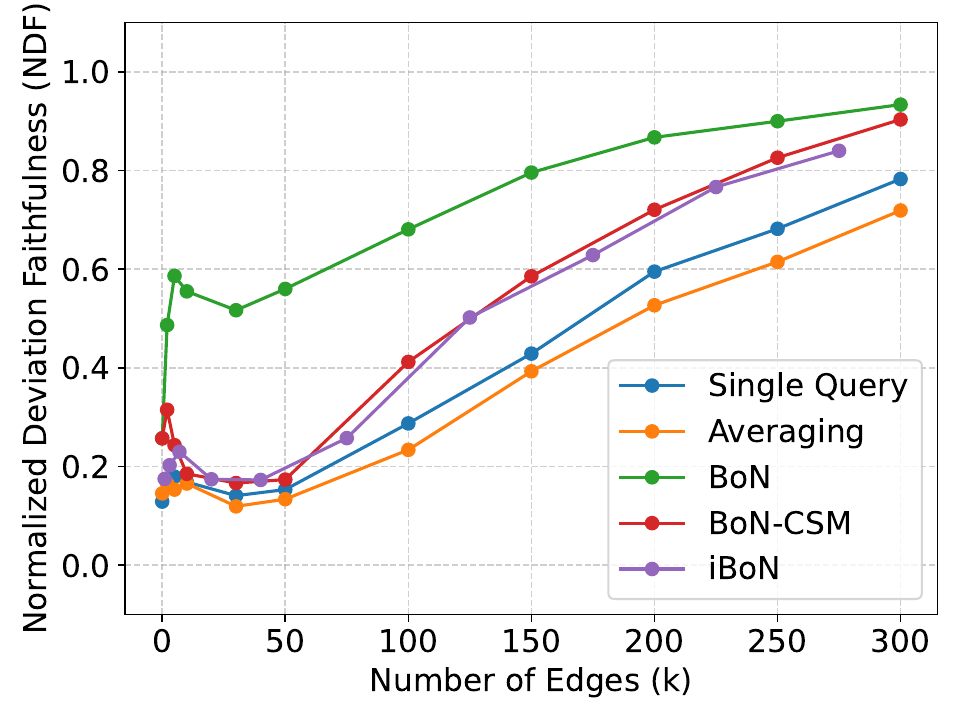}
    \caption{nutrition.}
  \end{subfigure}
  \hfill
  \begin{subfigure}{0.32\linewidth}
    \includegraphics[width=\textwidth]{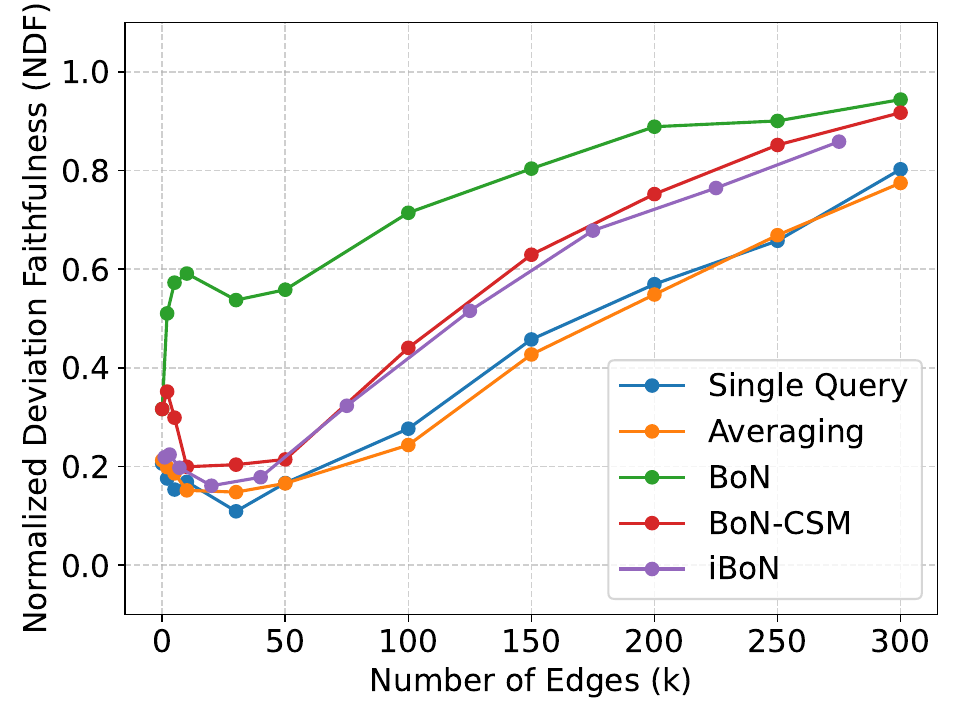}
    \caption{International Law.}
  \end{subfigure}
  \hfill
  \begin{subfigure}{0.32\linewidth}
    \includegraphics[width=\textwidth]{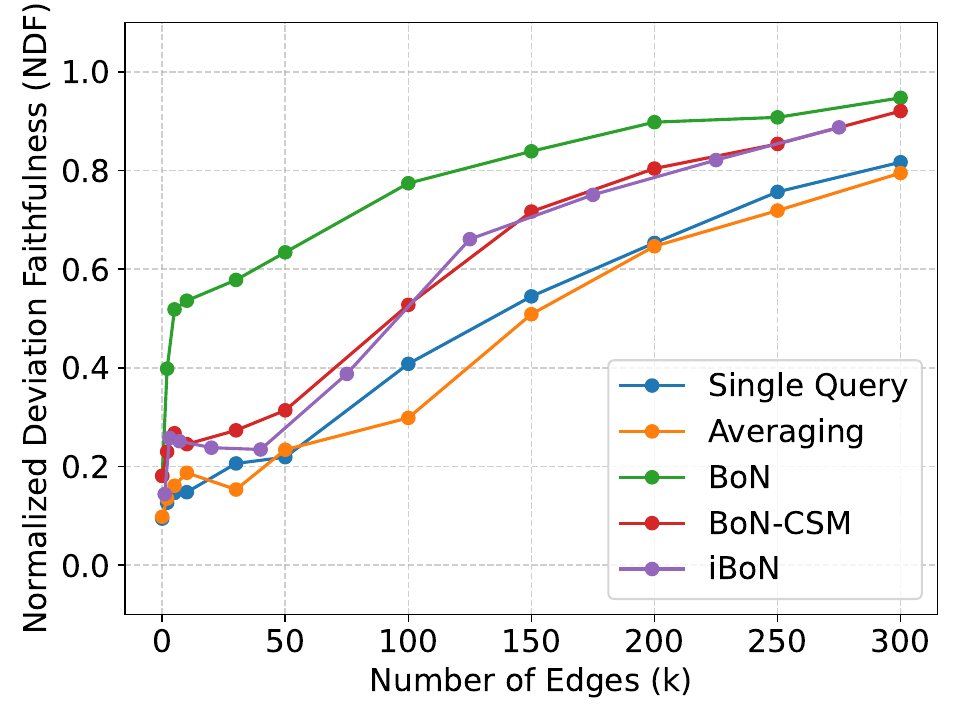}
    \caption{Management.}
  \end{subfigure}
  \caption{Complete results of BoN sampling for query circuit discovery on nine MMLU categories.}
  \label{fig: bon-mmlu-full-results}
\end{figure*}
Figure~\ref{fig: bon-mmlu-full-results} presents the full results for nine randomly selected MMLU categories. BoN consistently achieves around 0.6 NFS using only 5000 of the 386713 edges (1.3\%) in Llama-3.2-1B-Instruct. While iBoN and BoN-CSM do not yield circuits that are as faithful yet sparse as BoN, they still outperform the baseline methods clearly and consistently.

\subsection{Faithfulness Scores of Complement Circuits}
\label{supsub: complement circuit}
\begin{figure*}[tb]
  \centering
  \begin{subfigure}{0.32\linewidth}
    \includegraphics[width=\textwidth]{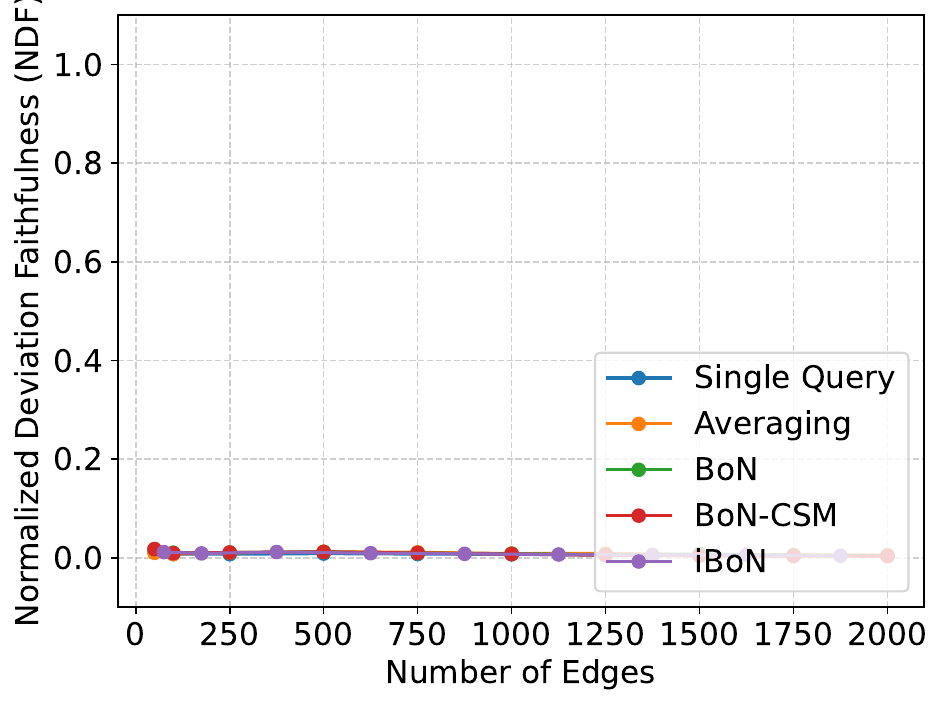}
    \caption{IOI.}
  \end{subfigure}
  \hfill
  \begin{subfigure}{0.32\linewidth}
    \includegraphics[width=\textwidth]{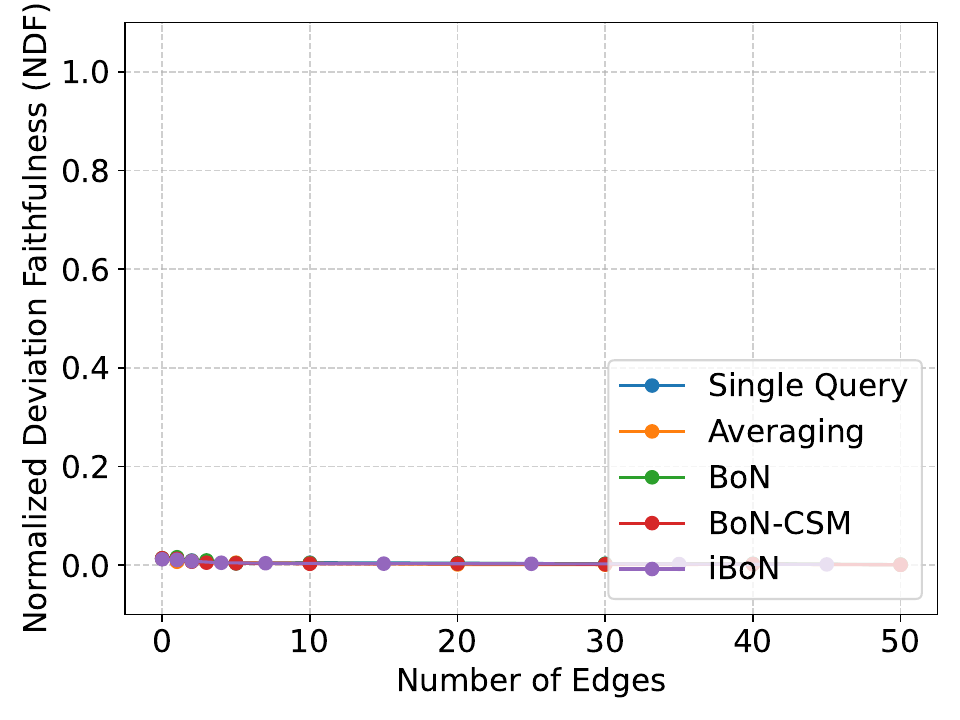}
    \caption{Arithmetic Addition.}
  \end{subfigure}
  \hfill
  \begin{subfigure}{0.32\linewidth}
    \includegraphics[width=\textwidth]{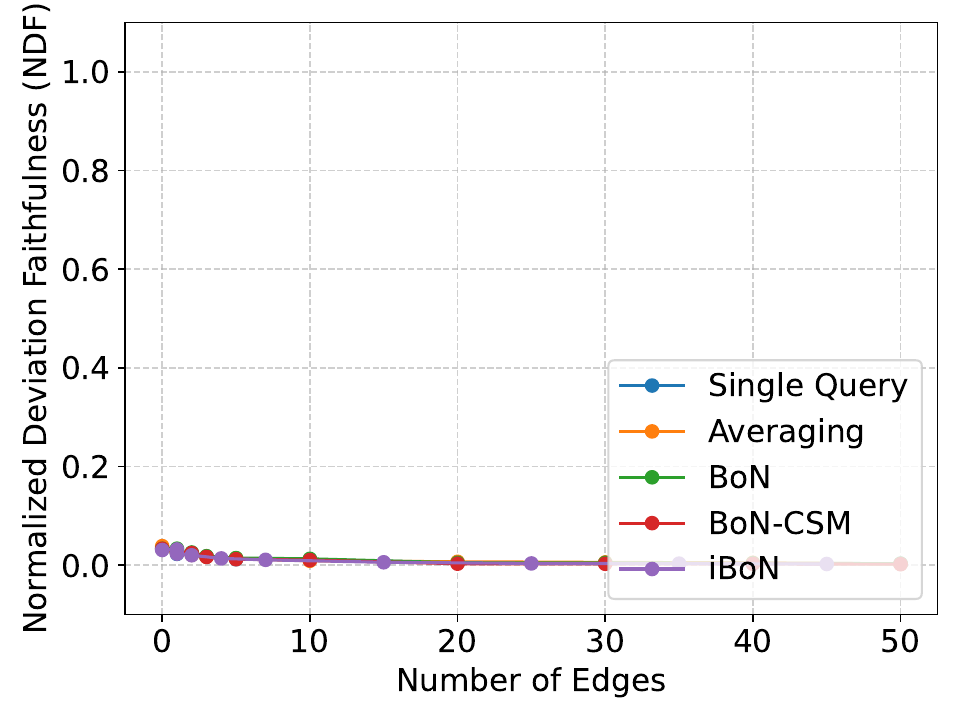}
    \caption{Arithmetic Multiplication.}
  \end{subfigure}
  \hfill
  \begin{subfigure}{0.32\linewidth}
    \includegraphics[width=\textwidth]{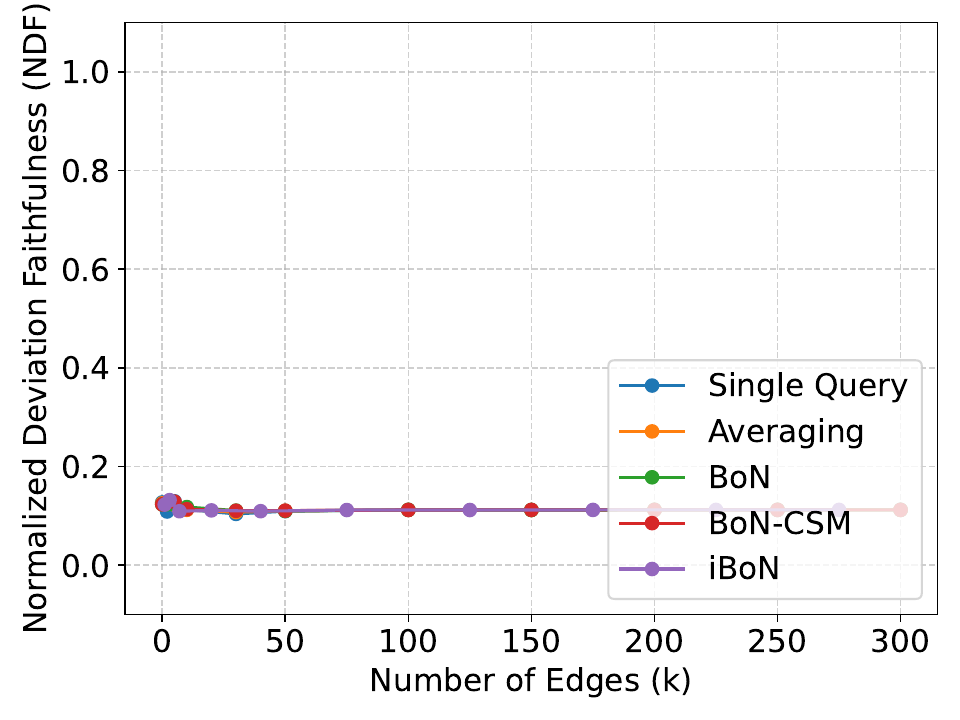}
    \caption{MMLU Marketing.}
  \end{subfigure}
  \hfill
  \begin{subfigure}{0.32\linewidth}
    \includegraphics[width=\textwidth]{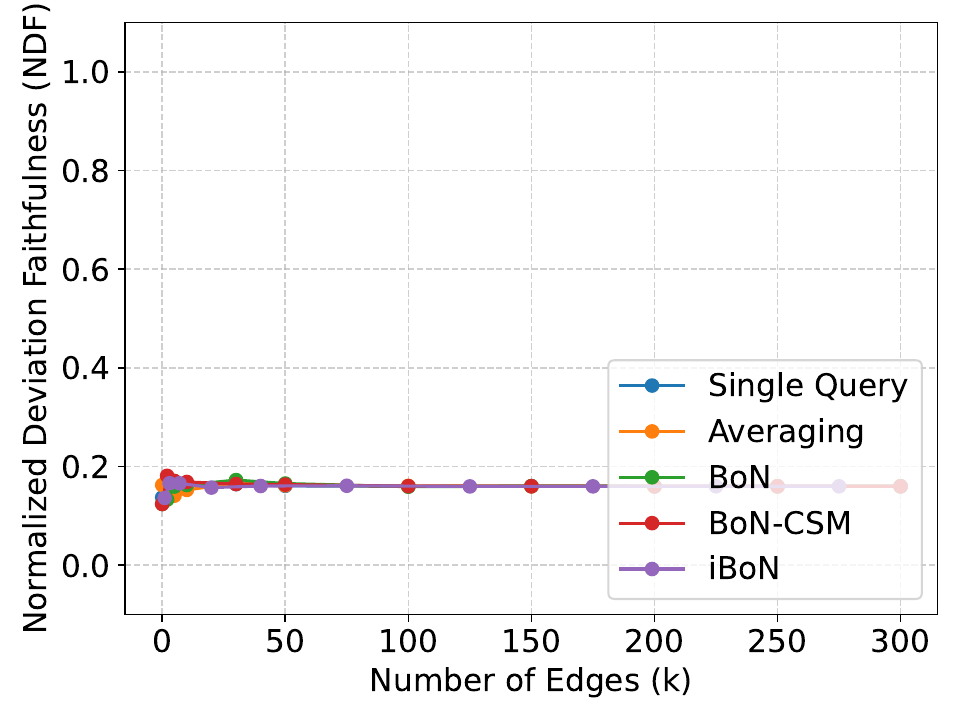}
    \caption{MMLU Astronomy.}
  \end{subfigure}
  \hfill
  \begin{subfigure}{0.32\linewidth}
    \includegraphics[width=\textwidth]{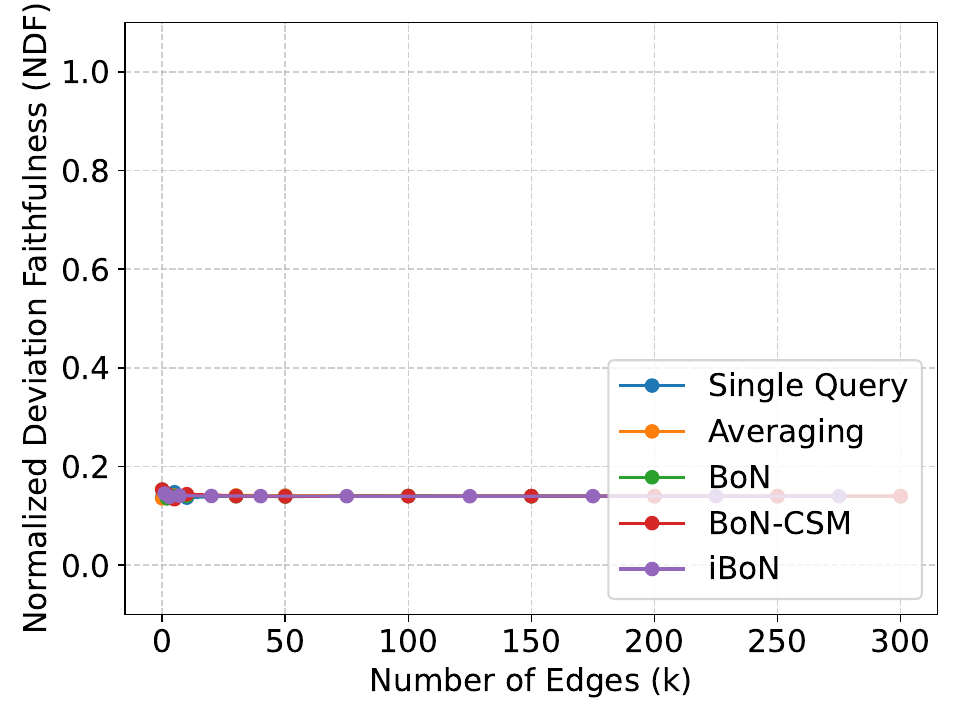}
    \caption{ARC Challenge.}
  \end{subfigure}
  \caption{NDF scores of complement circuits of the discovered query circuits.}
  \label{fig: bon-complement-main-results}
\end{figure*}
Figure~\ref{fig: bon-complement-main-results} shows the NDF scores of complement circuits $C^c \coloneqq E \setminus E_q$, where $E$ is the LLM’s edge set and $E_q$ the query circuit’s. Low, near-random faithfulness scores of complement circuits indicate that critical information flow indeed resides within the query circuits.

This experiment reflects a standard practice in circuit-discovery studies: following prior work (e.g., Figure 3 in feature circuits~\citep{marks2025sparse}), we adopt counterfactual evaluations by measuring the faithfulness of both a circuit $C$ (Figure~\ref{fig: main results}) and its complement $C^c$ (Figure~\ref{fig: bon-complement-main-results}). Faithfulness of $C$ corresponds to a sufficiency test—whether $C$ alone can reconstruct model behavior; whereas faithfulness of the complement $C^c$ acts as a necessity test: if $C$ truly contains the information required for the model to parse the input and generate the correct response, then ablating $C$ should break model performance, yielding low faithfulness for $C^c$. Consistent with \citet{marks2025sparse}, we observe uniformly low NDFs across methods for complement circuits, supporting the necessity of the discovered query circuits.

This analysis also clarifies why Figure~\ref{fig: main results} shows performance differences across discovery methods, while complement circuits in Figure~\ref{fig: bon-complement-main-results} remain uniformly unfaithful: low NDFs for both a query circuit and its complement simply indicate that neither alone forms a precise, self-sufficient combination of edges capable of reconstructing the full model behavior—only the true underlying circuit does.

Finally, for MMLU and ARC Challenge, complement circuits (Figure~\ref{fig: bon-complement-main-results}) and randomly constructed circuits (Figure~\ref{subfig: astronomy-not-work}) both yield NDF scores around 0.1–0.2, rather than 0. This is because we compute logit differences only among the options, and both original and corrupted queries still contain signals that lead the model to distribute logits across option IDs. As a result, even if a circuit fails to capture model performance well, its performance deviations from the original LLM may still be smaller than the gap between the original and corrupted queries in a few samples.

\subsection{Scaling BoN to Larger Models}
\label{supsub: comp-scalebon}
\begin{figure*}[tb]
  \centering
  \begin{subfigure}{0.3\linewidth}
    \includegraphics[width=\textwidth]{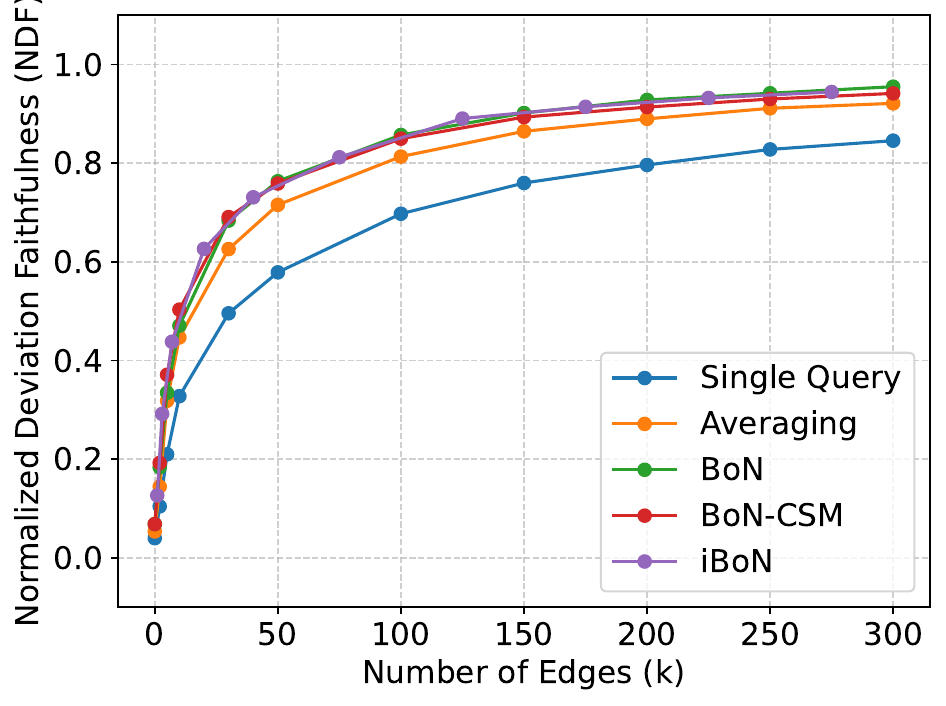}
    \caption{IOI dataset.}
    \label{subfig: ioi-gpt2-xl}
  \end{subfigure}
  \hfill
  \begin{subfigure}{0.3\linewidth}
    \includegraphics[width=\textwidth]{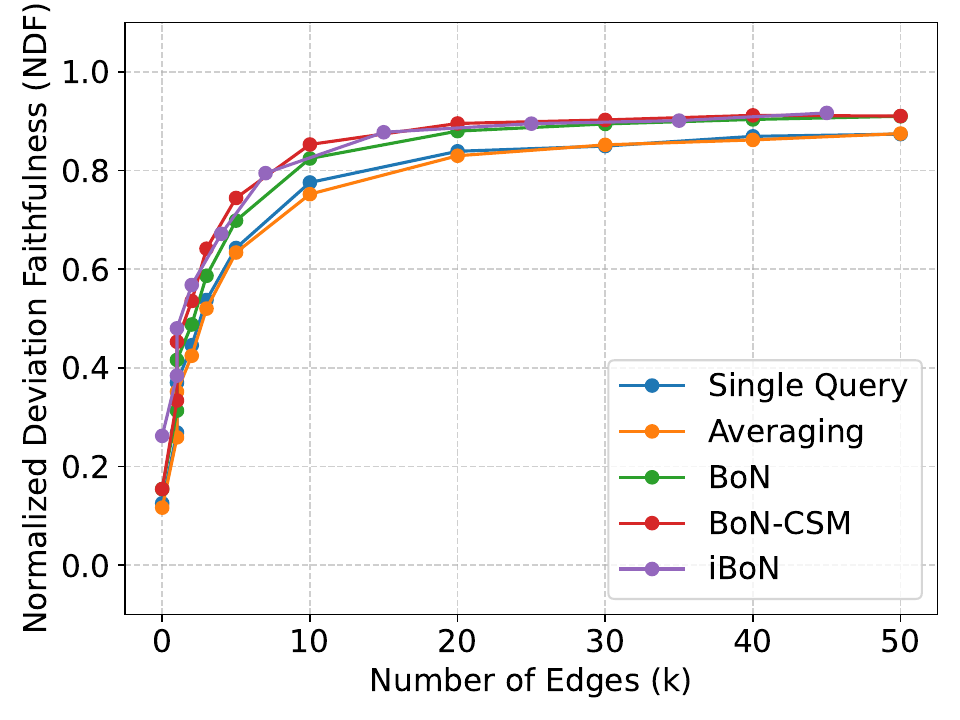}
    \caption{Arithmetic multiplication.}
    \label{subfig: arithmetic-mul-llama3-8b}
  \end{subfigure}
  \hfill
  \begin{subfigure}{0.3\linewidth}
    \includegraphics[width=\textwidth]{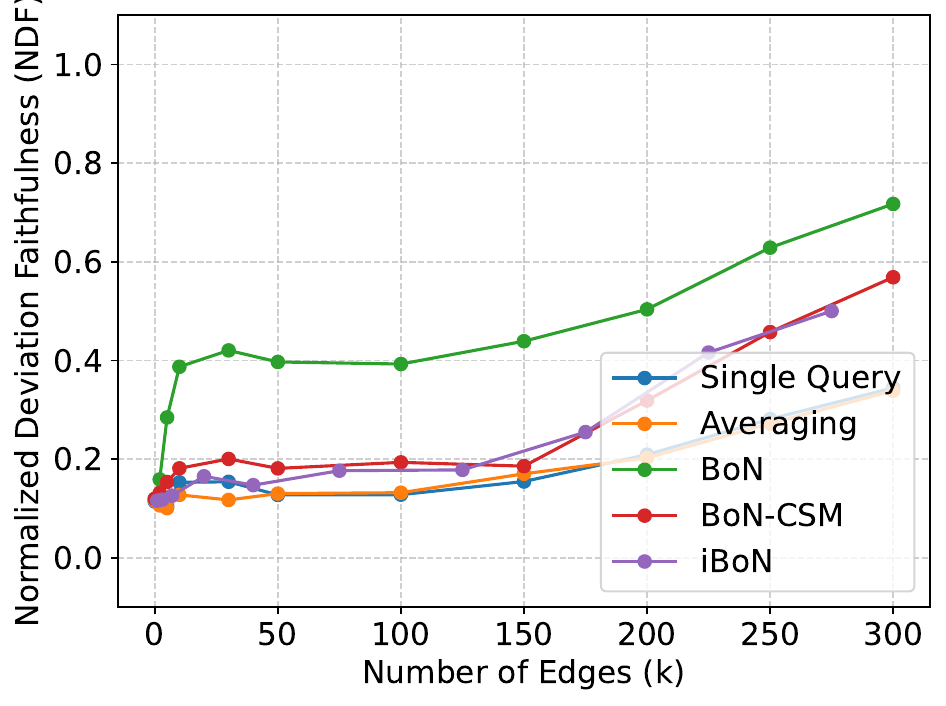}
    \caption{MMLU Astronomy.}
    \label{subfig: mmlu-astronomy-llama3-8b}
  \end{subfigure}
  \caption{Scaling BoN sampling for query circuit discovery to larger models (GPT-2 XL for IOI; Llama-3-8B-Instruct for arithmetic multiplication and MMLU astronomy). BoN, iBoN, and BoN-CSM still consistently outperform both baselines.}
  \label{fig: bon results on larger models}
\end{figure*}
We further scale the target models to GPT-2 XL (1.5B; 2235025 edges) for IOI and Llama-3-8B-Instruct (1592881 edges) for arithmetic multiplication and MMLU astronomy, as shown in Figure~\ref{fig: bon results on larger models}. Our methods continue to consistently outperform both baselines. Noe provide many tably, on average, BoN discovers a 5,000-edge query circuit (0.3\% of all edges) in Llama-3-8B-Instruct that reconstructs 0.4 NDF for the input query. In contrast, vanilla EAP-IG misleadingly suggests that even a 300k-edge circuit (18\% of the model) cannot achieve this level of faithfulness, giving a false impression that the underlying computational mechanism is much denser than it actually is. 

\subsection{Runtime Comparison of EAP-IG and BoN}
\label{supsub: comp-eapig-bon}

\begin{table}[tb]
  \caption{Average runtime of EAP-IG and BoN for discovering and evaluating a query circuit.}
  \label{tab: runtime analysis}
  \centering
  \addtolength{\tabcolsep}{4pt}
  \begin{tabular}{@{}p{0.12\textwidth}cccccccc@{}}
    \toprule
    Method & \multicolumn{5}{c}{EAP-IG} & \multicolumn{3}{c}{BoN} \\
    \cmidrule(lr){2-6} \cmidrule(lr){7-9}
    Parameter & 5 & 20 & 100 & 500 & 1000 & 1 & 4 & 9 \\
    \specialrule{1.2pt}{2pt}{2pt} 
    \makecell[l]{Per-query \\Runtime ($s$)} 
      & \makecell[c]{4.3}
      & \makecell[c]{9.5}
      & \makecell[c]{27.9} 
      & \makecell[c]{120.2} 
      & \makecell[c]{237.5} 
      & \makecell[c]{25.4} 
      & \makecell[c]{66.0}
      & \makecell[c]{132.0} \\
    \specialrule{1.2pt}{0pt}{0pt} 
  \end{tabular}
\end{table}


Table~\ref{tab: runtime analysis} reports the average runtime for discovering and evaluating query circuits on MMLU Astronomy. For each query, we identify and evaluate 11 circuits of varying sizes as in Figure~\ref{fig: main results}. The reported runtime is averaged over the 11 circuits and then over all 152 samples. We compare EAP-IG with varying IG steps and BoN with different numbers of additional paraphrases. Runtime of BoN with nine paraphrases is slightly longer than that of EAP-IG with 500 steps, while EAP-IG consistently yields suboptimal performance in query circuit discovery even with 1000 steps (Figure~\ref{subfig: astronomy-not-work}).

\subsection{Runtime Comparison of Greedy Selection and Dijkstra-like Construction}
\label{supsub: comp-greedy-dijkstra}
\begin{figure*}[t]
  \centering
  \begin{subfigure}{0.32\linewidth}
    \includegraphics[width=\textwidth]{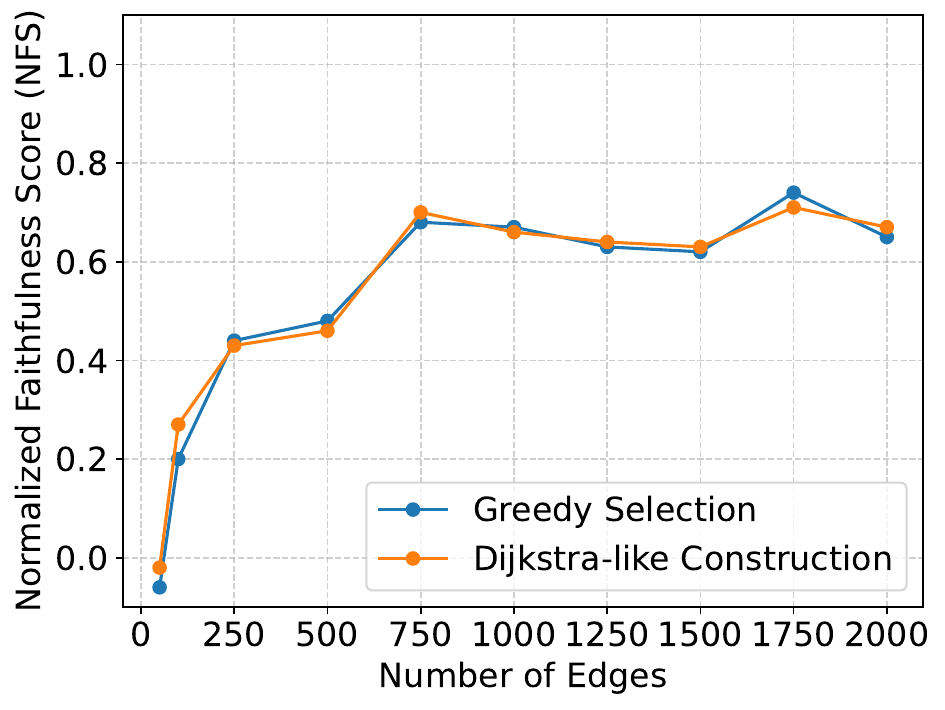}
    \caption{IOI.}
  \end{subfigure}
  \hfill
  \begin{subfigure}{0.32\linewidth}
    \includegraphics[width=\textwidth]{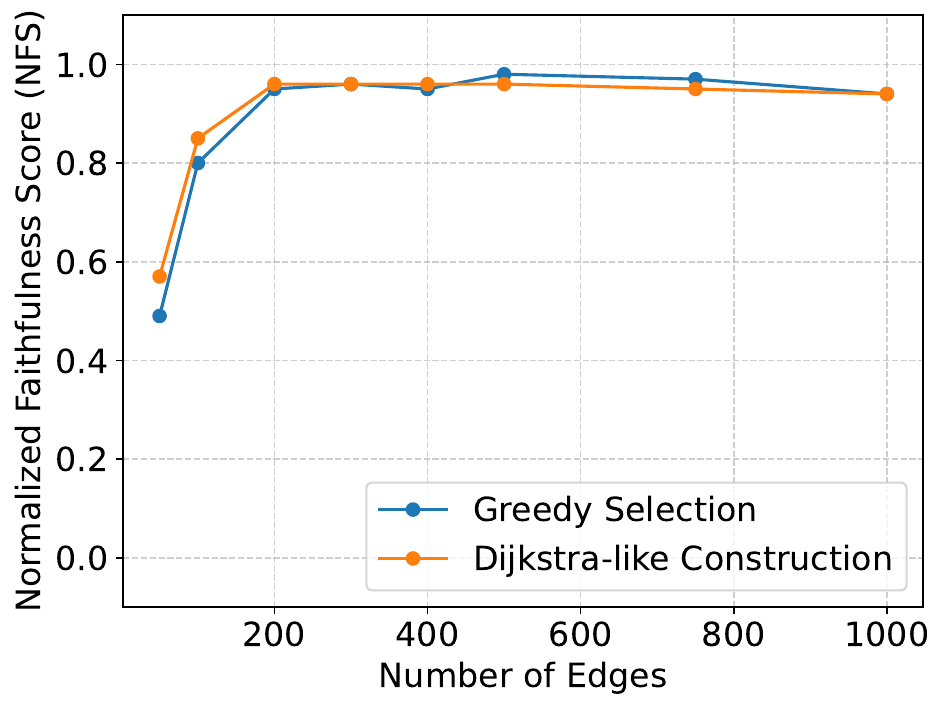}
    \caption{Greater-than comparison.}
  \end{subfigure}
  \hfill
  \begin{subfigure}{0.32\linewidth}
    \includegraphics[width=\textwidth]{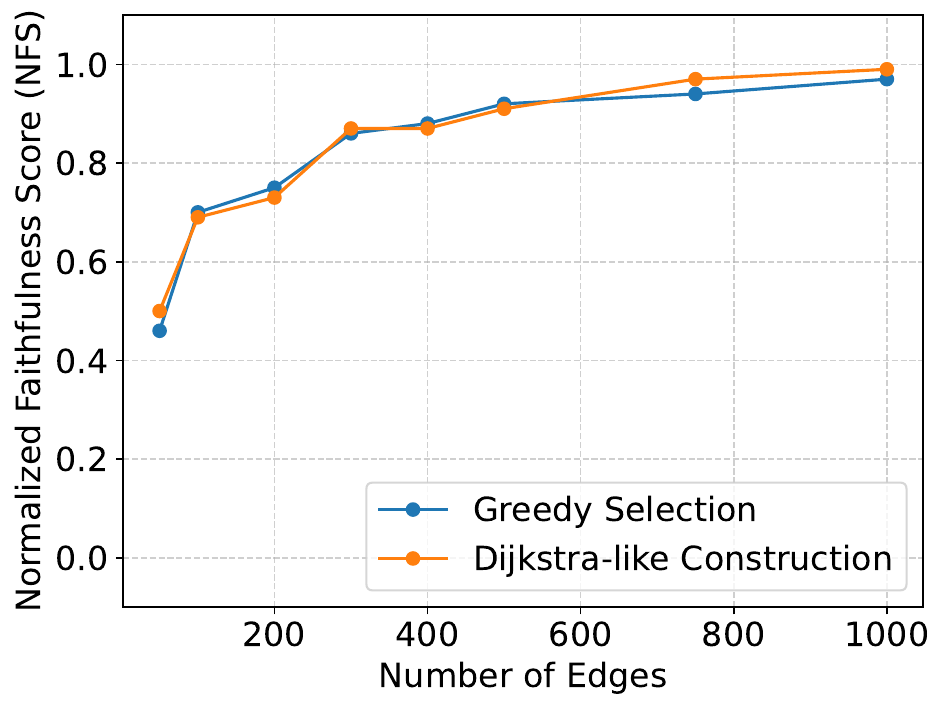}
    \caption{Gender Bias.}
  \end{subfigure}
  \caption{Performance comparisons between greedy selection and Dijkstra-like iterative construction for forming a circuit. The two methods achieve similar results on all three tested datasets.}
  \label{fig: greedy dijkstra comparison}
\end{figure*}
\begin{table}[tb]
  \caption{Runtime of greedy selection and Dijkstra-like iterative construction for forming circuits based on edge scores. The latter's runtime increases with respect to the number of edges $N$.}
  \label{tab: construction runtime}
  \centering
  \addtolength{\tabcolsep}{4pt}
  \begin{tabular}{@{}p{0.16\textwidth}|ccc|ccc@{}}
    \toprule
    Construction Method & \multicolumn{3}{c|}{Greedy Selection} & \multicolumn{3}{c}{Dijkstra-like Iteration} \\
    \cmidrule(lr){1-4} \cmidrule(lr){5-7}
    edge number $N$ & 10$k$ & 100$k$ & 300$k$ & 10$k$ & 100$k$ & 300$k$ \\
    \midrule
    \makecell[l]{Per-circuit \\Runtime ($s$)} 
      & \makecell[c]{$<$ 0.1}
      & \makecell[c]{$<$ 0.1} 
      & \makecell[c]{$<$ 0.1} 
      & \makecell[c]{23.9} 
      & \makecell[c]{274.8}
      & \makecell[c]{729.9} \\
    \bottomrule
  \end{tabular}
\end{table}

We compare the efficiency and effectiveness of greedy selection and Dijkstra-like construction for circuit discovery. Figure~\ref{fig: greedy dijkstra comparison} presents results on the IOI, GT, and Gender-bias datasets with GPT-2 Small as the target model. The two methods achieve similar performance across all three datasets. Table~\ref{tab: construction runtime} further reports the runtime of the two methods for constructing a circuit in Llama-3.2-1B-Instruct after obtaining the score matrix $S$. The Dijkstra-like iterative construction incurs substantially higher runtime as the edge budget $N$ increases. In contrast, greedy selection requires constant time regardless of circuit size, so we adopt it throughout this work.

\subsection{Runtime Comparison of Activation and Attribution Patching in Query Setting}
\label{supsub: acdc-slow}

\begin{figure*}[tb]
  \centering
  \includegraphics[width=0.8\textwidth]{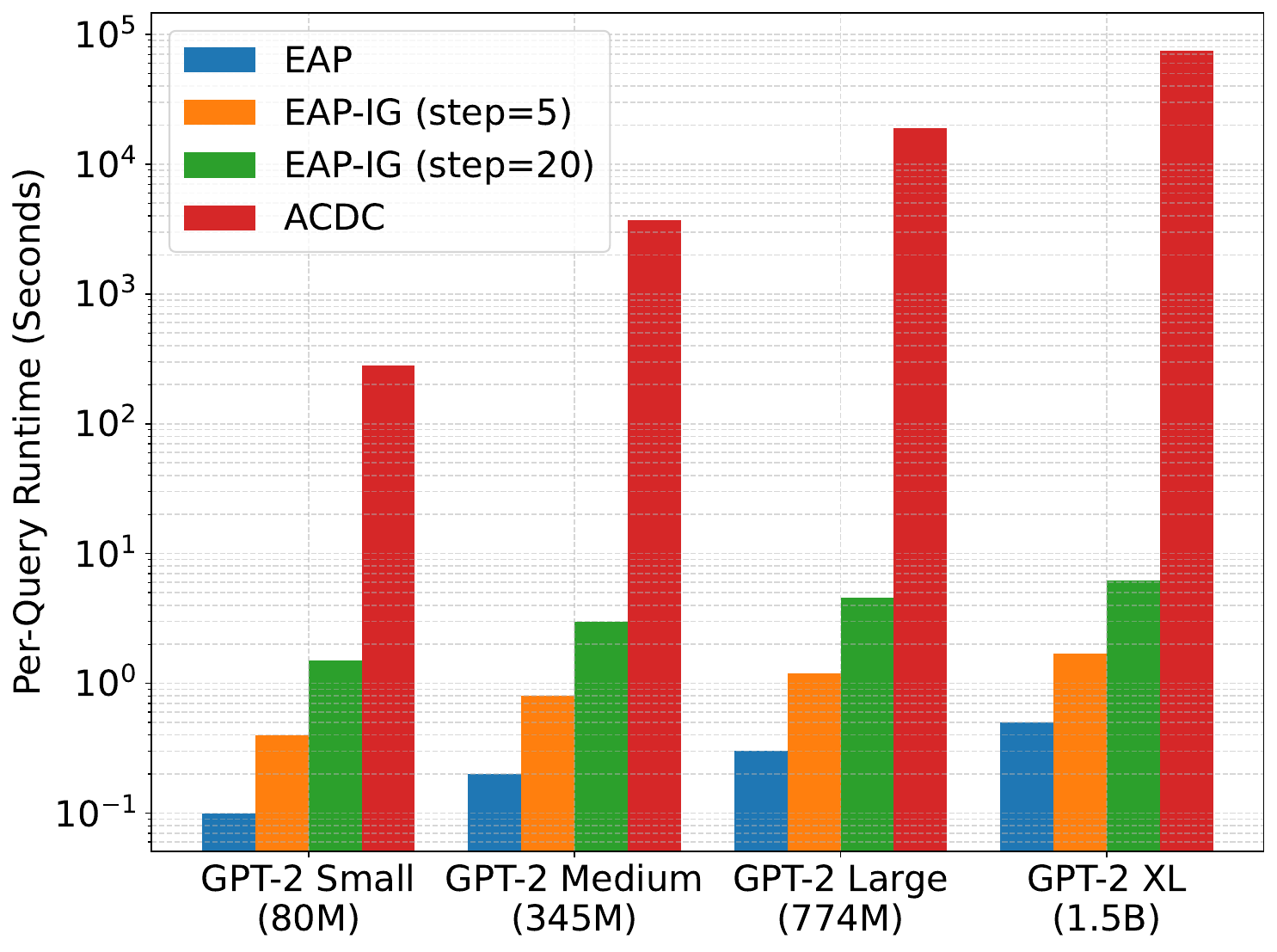}
  \caption{Per-query edge scoring runtime of activation patching (ACDC) versus attribution patching (EAP and EAP-IG) methods. The dataset is IOI. Runtime of ACDC easily grows to hours.}
  \label{fig: acdc-slow}
\end{figure*}
We present additional runtime analysis of activation patching and attribution patching methods, two major categories of circuit discovery within the model, in the setting of query circuits. For the former, we use ACDC; for the latter, we use EAP and EAP-IG. As shown in Figure~\ref{fig: acdc-slow}, ACDC takes about 18000 seconds (5 hours) on an NVIDIA A100 to discover a circuit for a query in GPT-2 Large on the IOI dataset. This is because edge activation patching requires two LLM forward passes per edge, as discussed in Section~\ref{subsubsec: score edge and construct circuit}. In contrast, EAP~\citep{syed-etal-2024-attribution} and EAP-IG~\citep{hanna2024have}, two representative attribution patching methods, take less than 10 seconds as they score all edges at once.

Since LLM systems process numerous queries per day~\citep{NBERw34255}, it is important that the query circuit discovery methods to trace and explain model decisions are scalable. As a result, we do not consider ACDC as a backbone method for query circuit discovery in this paper.


\subsection{Additional Variants of BoN Sampling}
\label{supsub: different types of bon}
\begin{figure*}[t]
  \centering
  \begin{subfigure}{0.48\linewidth}
    \includegraphics[width=\textwidth]{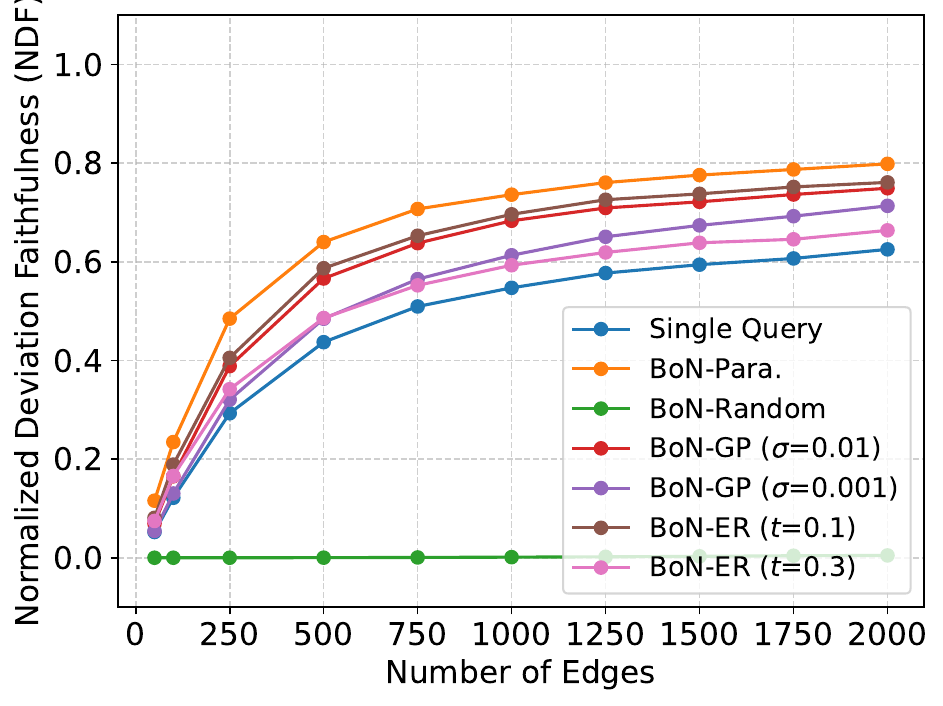}
    \caption{IOI.}
  \end{subfigure}
  \hfill
  \begin{subfigure}{0.48\linewidth}
    \includegraphics[width=\textwidth]{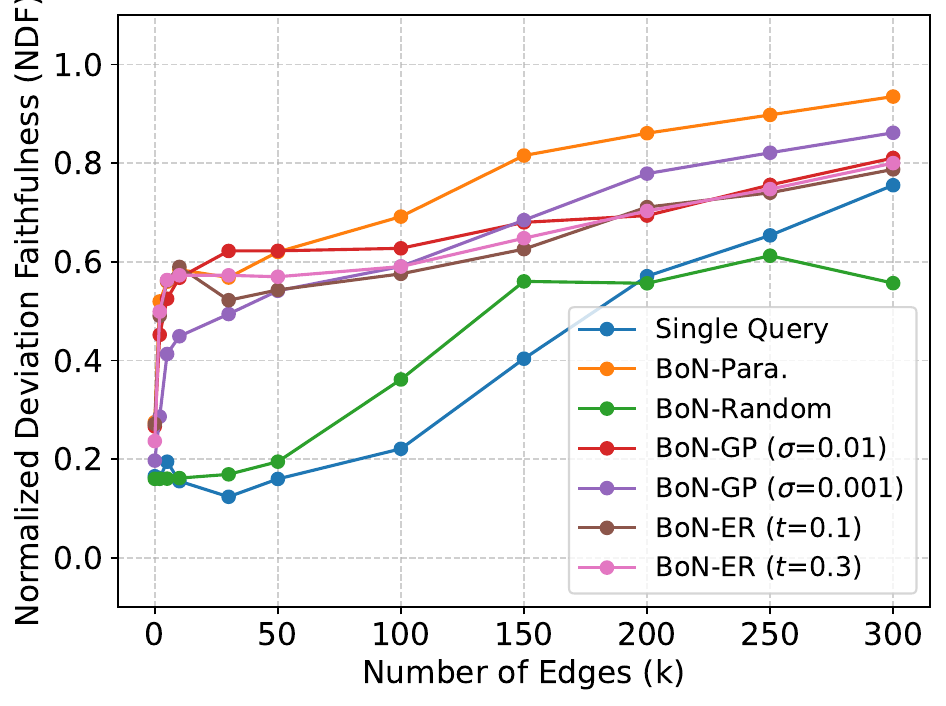}
    \caption{MMLU Astronomy.}
  \end{subfigure}
  \caption{Query circuit discovery results for additional BoN variants (BoN-Random, BoN-GP, and BoN-ER), along with BoN by paraphrases (BoN-Para.) introduced in the main paper.}
  \label{fig: different bon}
\end{figure*}
We introduce and investigate three additional variants of BoN sampling for query circuit discovery here. (i) BoN-GP (Gaussian Perturbation): add Gaussian noise $G(0,\sigma^2)$ to the score matrix $S$ from the original query to alter edge selection. Repeating this $p$ times yields ${S, S_1, \dots, S_p}$, and BoN sampling is performed over the 10 resulting circuits under edge budget $N$. (ii) BoN-ER (Edge Replacement): given a circuit at budget $N$, randomly replace $t \times 100\%$ of its edges with unused ones for $p$ trials, then select the best circuit among these and the original. (iii) BoN-Random: randomly sample $N$ edges to form a circuit, repeat 10 times, and take the best. Our main method, BoN-Para., instead uses $p$ paraphrases to produce additional $p$ different score matrices.

Figure~\ref{fig: different bon} presents results on IOI and MMLU Astronomy, with $p=9$ for all methods. For BoN-GP, we set $\sigma \in \{0.01, 0.001\}$; for BoN-ER, $t \in \{0.1, 0.3\}$. Both BoN-GP and BoN-ER show potential to discover small, faithful query circuits in MMLU Astronomy, but our main method, BoN-Para. (semantics-preserving score matrix perturbation), remains superior. Notably, BoN-Random remains stuck near 0.2 NDF—similar to the single-query baseline—until circuit size exceeds $50k$ edges, after which it begins to outperform the baseline up to about $200k$ edges. This likely occurs because, as more edges are added, random selection has a higher chance of including all critical edges and forming a large circuit that recovers model performance, a chance that is further amplified by BoN sampling.

\subsection{Query Circuit Discovery Evaluated by NFS}
\label{supsub: main results on nfs}
\begin{figure*}[t]
  \centering
  \begin{subfigure}{0.48\linewidth}
    \includegraphics[width=\textwidth]{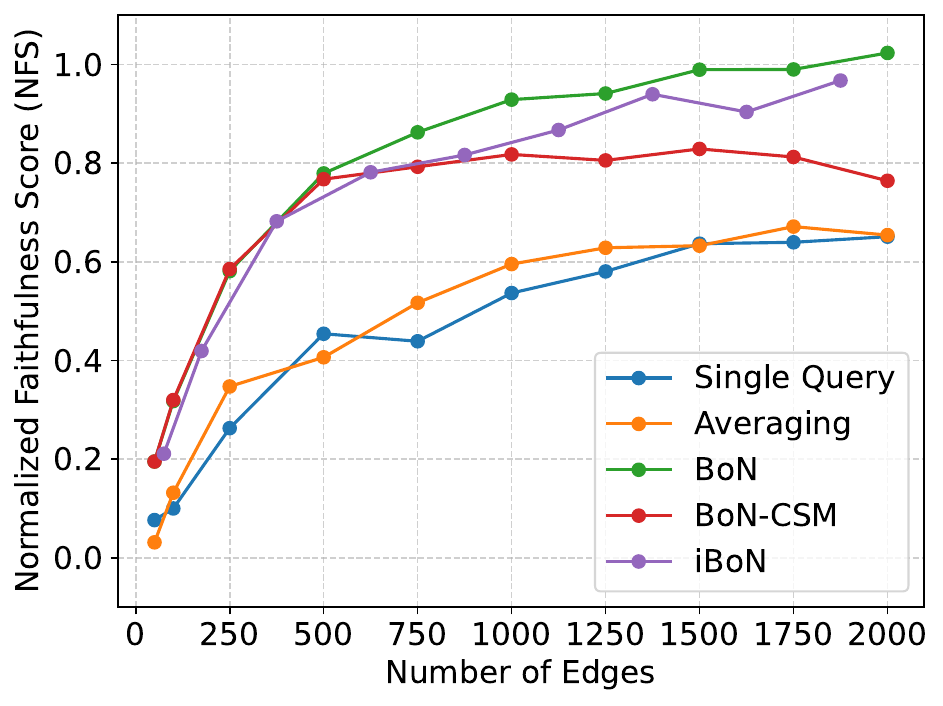}
    \caption{IOI.}
  \end{subfigure}
  \hfill
  \begin{subfigure}{0.48\linewidth}
    \includegraphics[width=\textwidth]{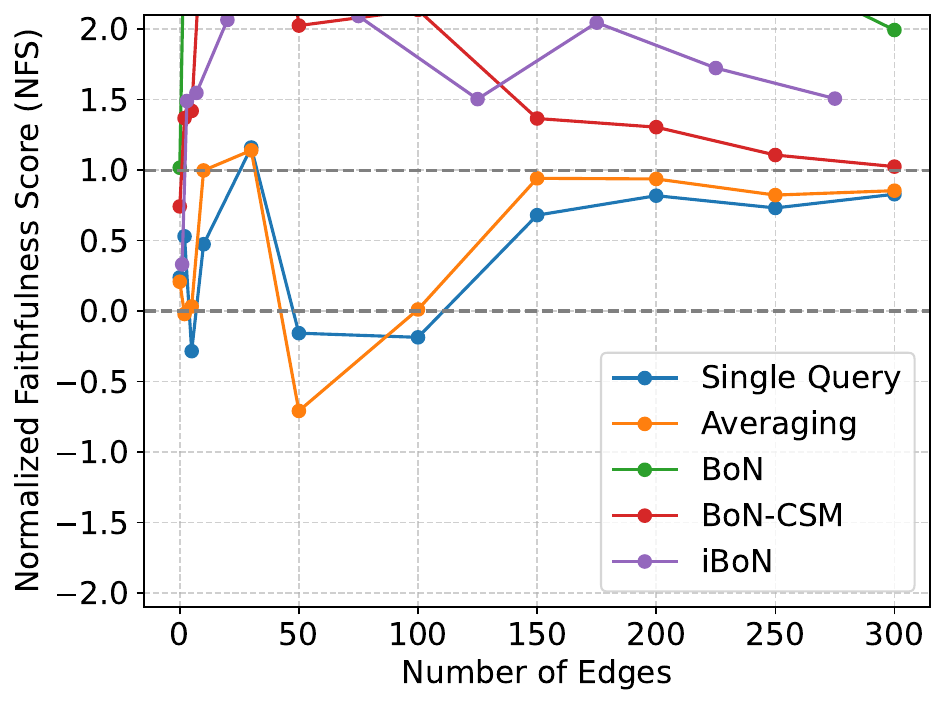}
    \caption{MMLU Astronomy.}
  \end{subfigure}
  \caption{Query circuit discovery on the full IOI and MMLU Astronomy datasets, evaluated using NFS as in most prior studies of capability circuits. On MMLU, however, NFS fails to provide a stable and reliable evaluation of query circuits and cannot effectively track discovery progress as circuit size increases.}
  \label{fig: main results on nfs}
\end{figure*}
Figure~\ref{fig: main results on nfs} reports query circuit discovery results on the complete IOI and MMLU Astronomy datasets, evaluated using NFS instead of our proposed NDF metric. On IOI, the researcher-curated toy dataset, NFS scores mostly remain within $[0,1]$. In contrast, for MMLU Astronomy, NFS scores fluctuate widely even after averaging over all 152 samples, making it difficult to track discovery progress and undermining confidence in circuit faithfulness as measured by NFS. This motivates our proposal of NDF as a more robust and reliable alternative, as shown and discussed in the main paper.

\subsection{More Analysis on Circuit Variances and Shared Sub-circuit}
\label{supsub: shared circuit}
\begin{figure*}[tb]
  \centering
  \begin{subfigure}{0.3\linewidth}
    \includegraphics[width=\textwidth]{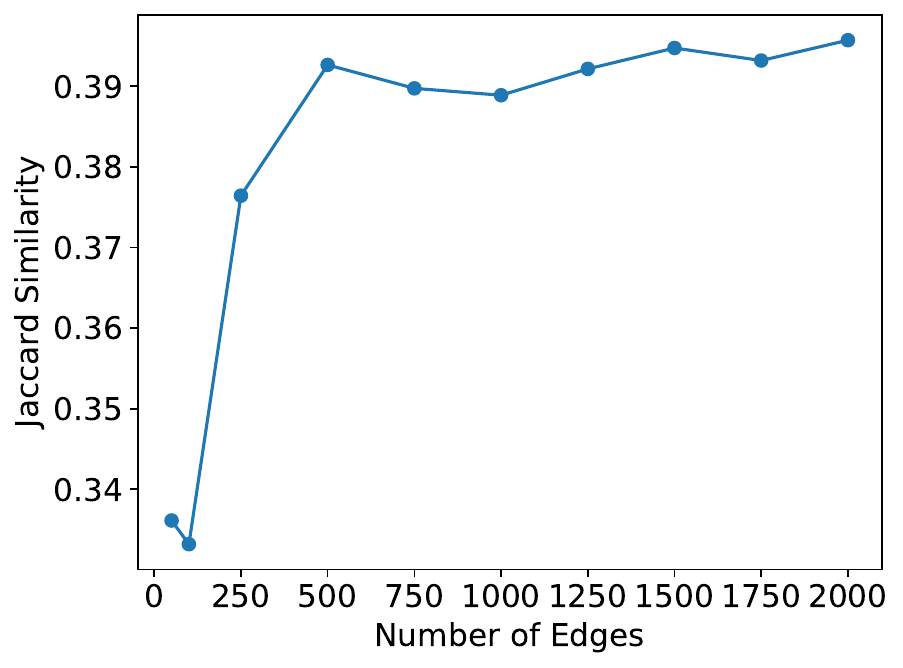}
    \caption{Averaged Jaccard similarity between the capability circuit and query circuits.}
    \label{subfig: js between c and q}
  \end{subfigure}
  \hfill
  \begin{subfigure}{0.3\linewidth}
    \includegraphics[width=\textwidth]{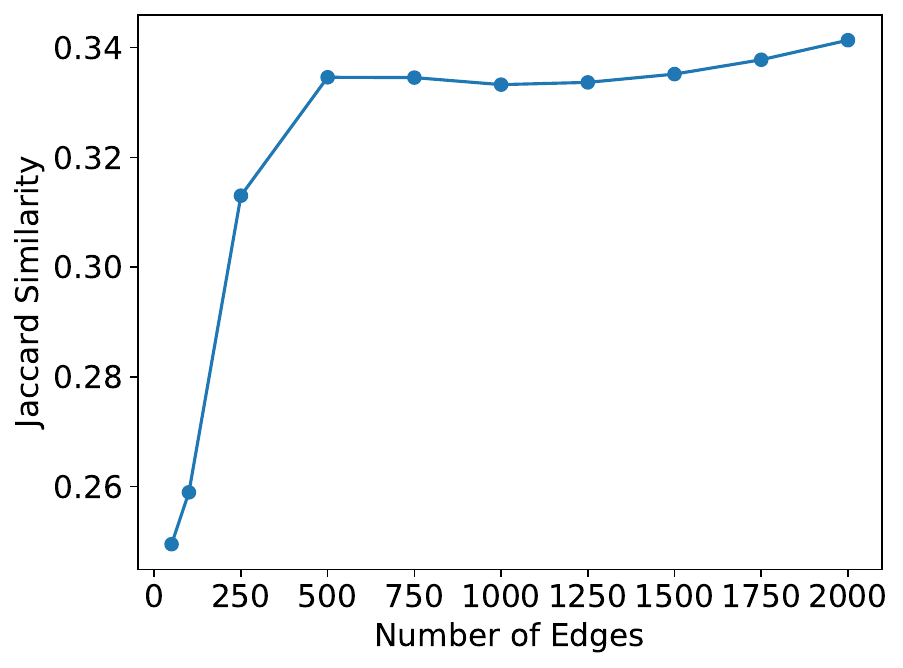}
    \caption{Averaged Jaccard similarity among the circuits derived from a query and its paraphrases.}
    \label{subfig: js among q}
  \end{subfigure}
  \hfill
  \begin{subfigure}{0.3\linewidth}
    \includegraphics[width=\textwidth]{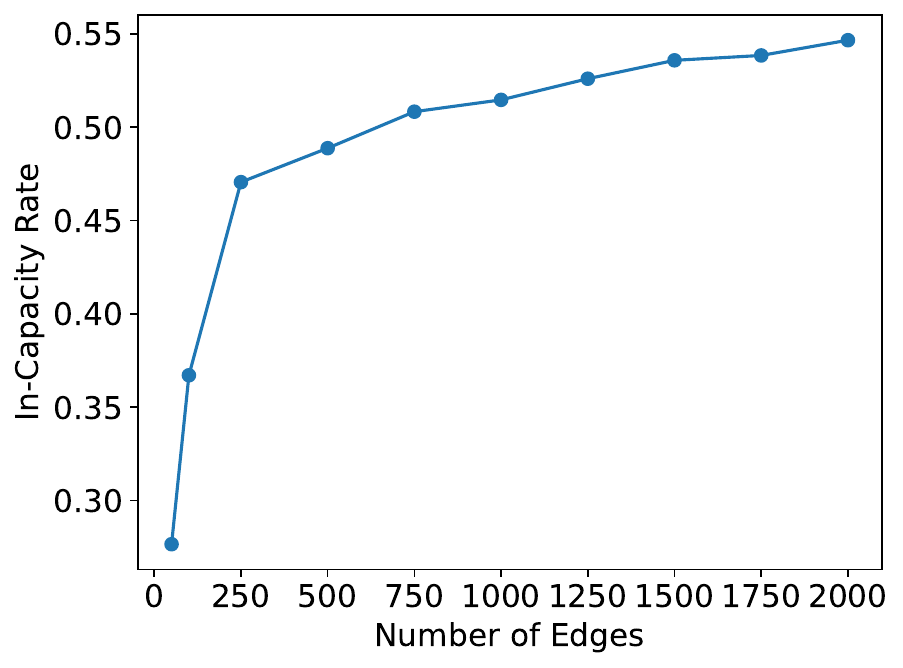}
    \caption{Averaged percentage of edges in query circuits that also appear in the capability circuit.}
    \label{subfig: intersect ratio}
  \end{subfigure}
  \caption{Analysis on edge overlap.}
  \label{fig: more analysis on capability and query}
\end{figure*}
\begin{figure*}[tb]
  \centering
  \begin{subfigure}{0.69\linewidth}
    \includegraphics[width=\textwidth]{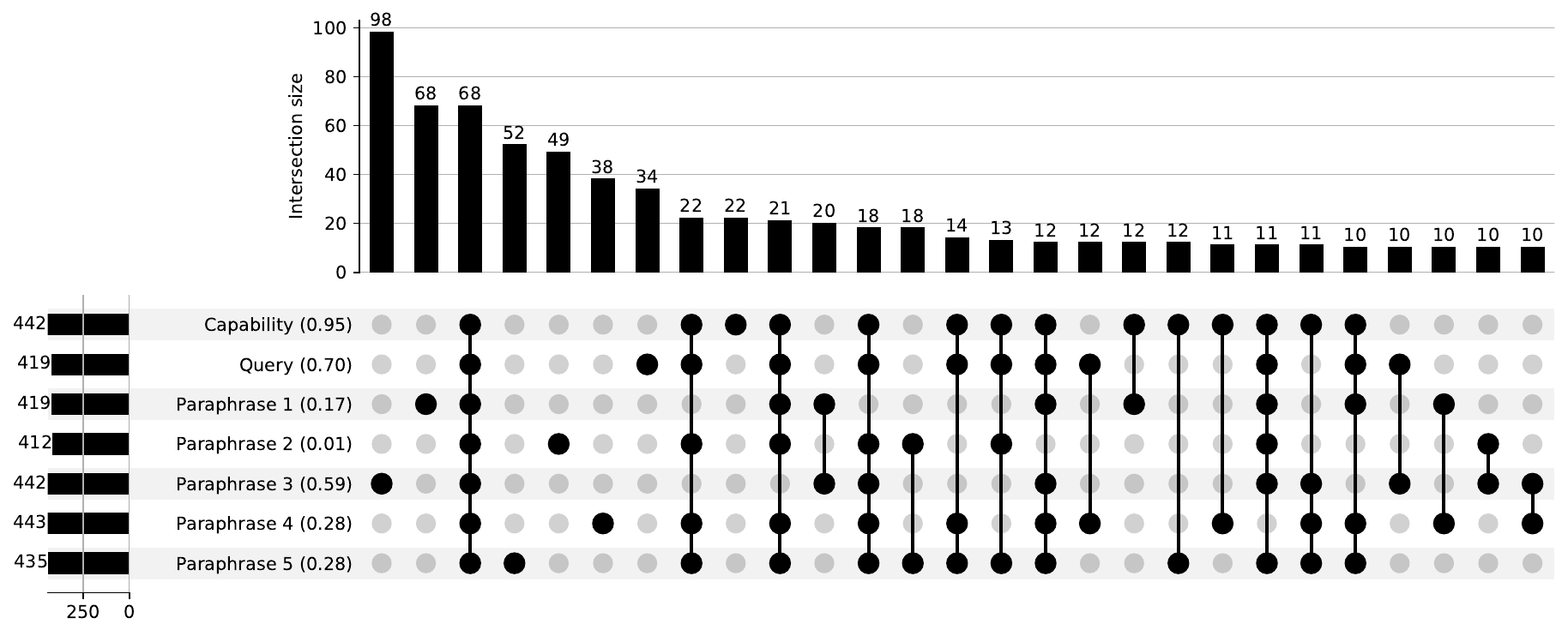}
    \caption{Query index = 84.}
  \end{subfigure}
  \hfill
  \begin{subfigure}{0.69\linewidth}
    \includegraphics[width=\textwidth]{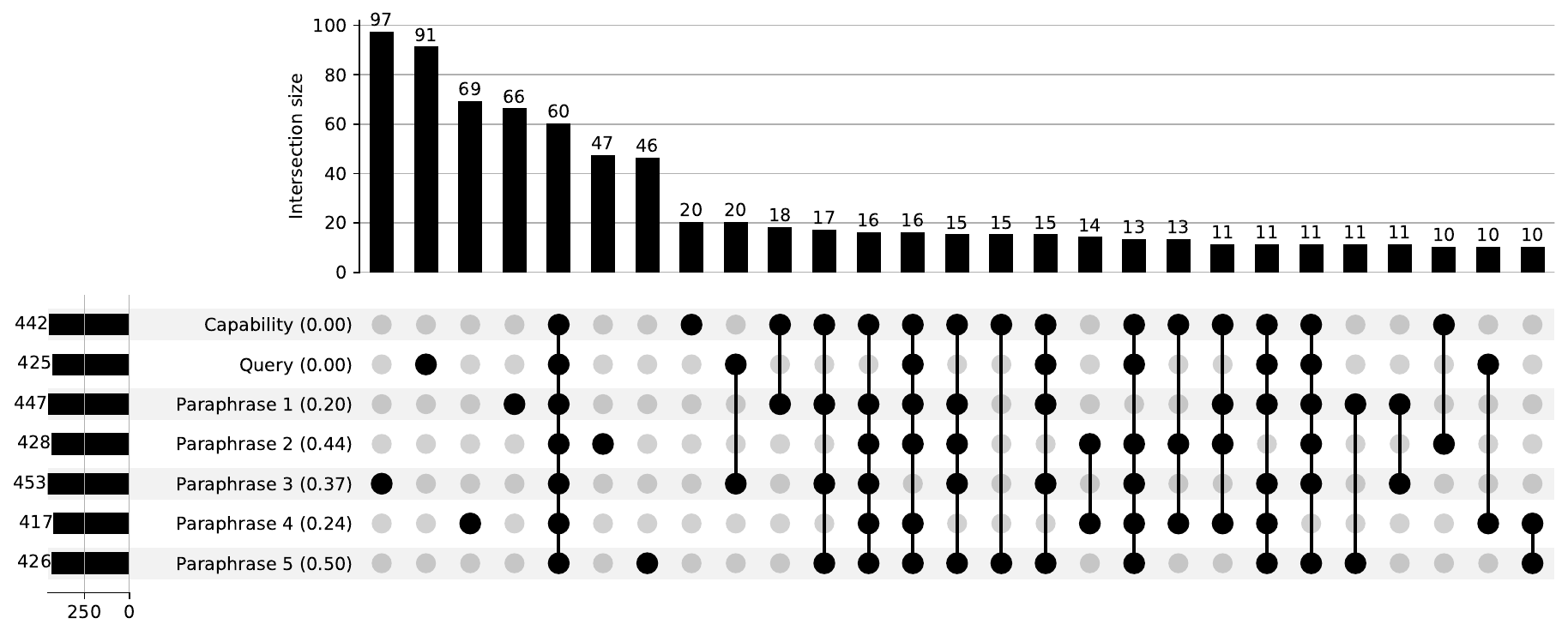}
    \caption{Query index = 177.}
  \end{subfigure}
  \hfill
  \begin{subfigure}{0.69\linewidth}
    \includegraphics[width=\textwidth]{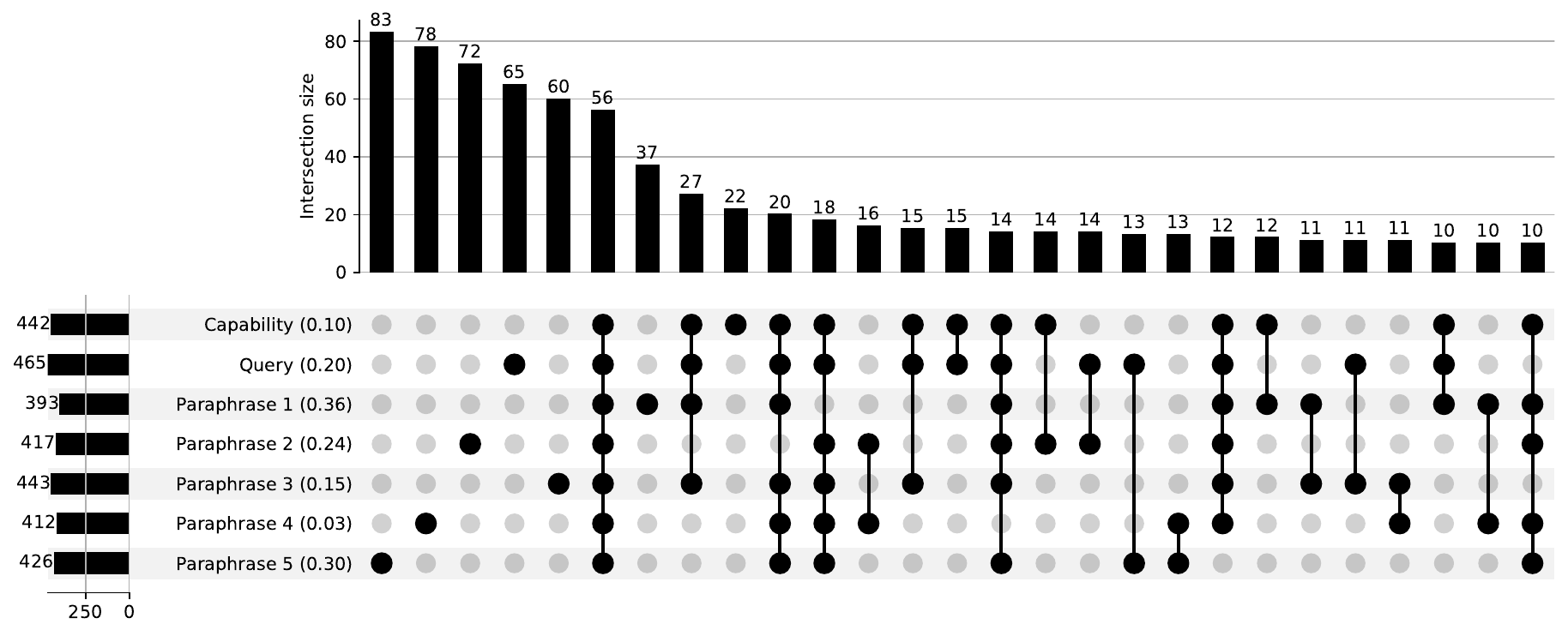}
    \caption{Query index = 380.}
  \end{subfigure}
  \hfill
  \begin{subfigure}{0.69\linewidth}
    \includegraphics[width=\textwidth]{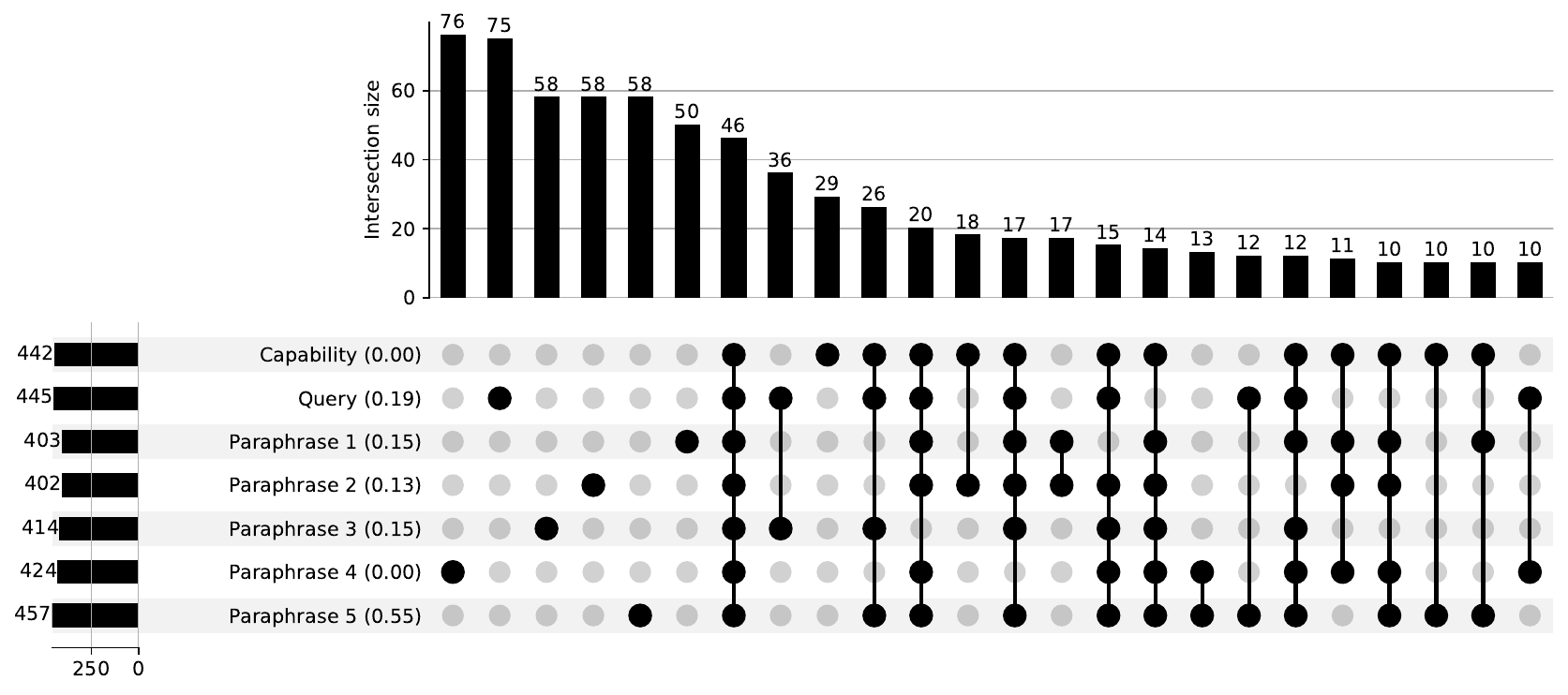}
    \caption{Query index = 489.}
  \end{subfigure}
  \caption{Upset plots of four additional randomly selected queries other than the one in Figure~\ref{fig: upset plot}.}
  \label{fig: upset plots of four queries}
\end{figure*}
Here, we provide additional experiments analyzing the relationship between the capability circuit and query circuits. In Figures~\ref{subfig: js between c and q} and~\ref{subfig: js among q}, the averaged Jaccard similarity of around 0.3 indicates non-trivial edge overlap. Figure~\ref{subfig: intersect ratio} further shows that 30–50\% of edges in query circuits also appear in the capability circuit. In Figure~\ref{fig: upset plots of four queries}, we show UpSet plots—analogous to Figure~\ref{fig: upset plot}—for five additional queries, all exhibiting substantial shared edges. All queries are randomly selected with seed 2025. Finally, Figure~\ref{fig: circuit plot} visualizes the full set of seven circuits analyzed in Figure~\ref{fig: upset plot} (the capability circuit and the circuits derived from the original query and five paraphrases). Nodes and edges shared by all circuits are marked in red; others in green. These shared components constitute a common sub-circuit present across all circuit variants for the IOI task, regardless of query phrasing or whether IEs are averaged over many IOI queries.
\begin{longtable}{c}
\caption{Complete plots of the seven circuits analyzed in Figure~\ref{fig: upset plot}. Shared nodes and edges are shown in red; others in green. These shared components constitute the sub-circuit common to all seven circuits.}
\label{fig: circuit plot} \\
\endfirsthead

\caption[]{continued} \\
\endhead

\includegraphics[width=0.85\linewidth]{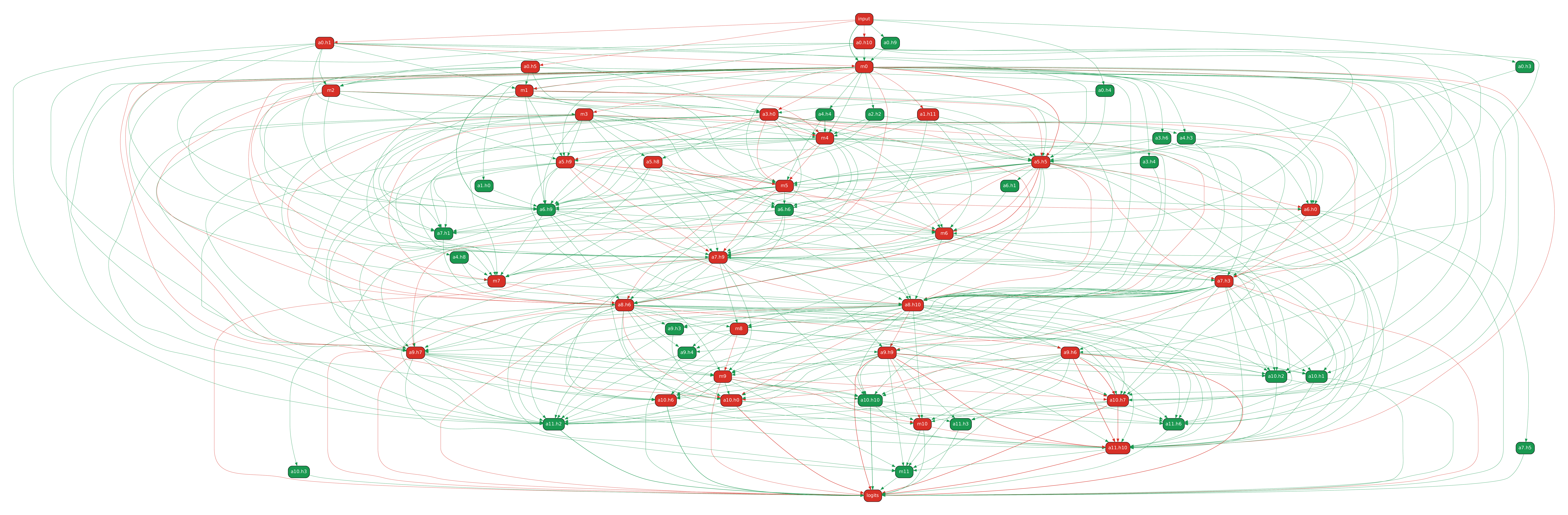}\\
(a) Capability circuit. \\
[1em]

\includegraphics[width=0.85\linewidth]{supp/figures/images/query_circuit_by_original_query.pdf}\\
(b) Query circuit by the original query. \\
[1em]

\includegraphics[width=0.85\linewidth]{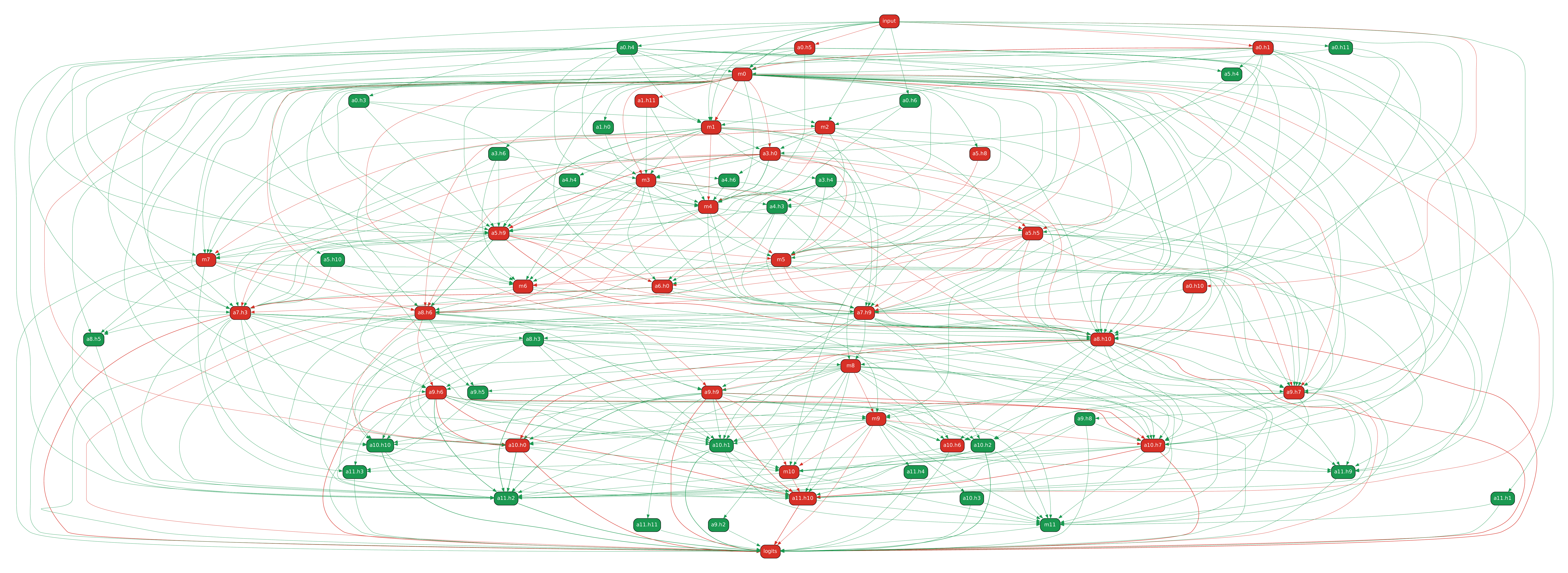}\\
(c) Query circuit by paraphrase 1. \\
[1em]

\includegraphics[width=0.85\linewidth]{supp/figures/images/query_circuit_by_paraphrase_2.pdf}\\
(d) Query circuit by paraphrase 2. \\
[1em]

\includegraphics[width=0.85\linewidth]{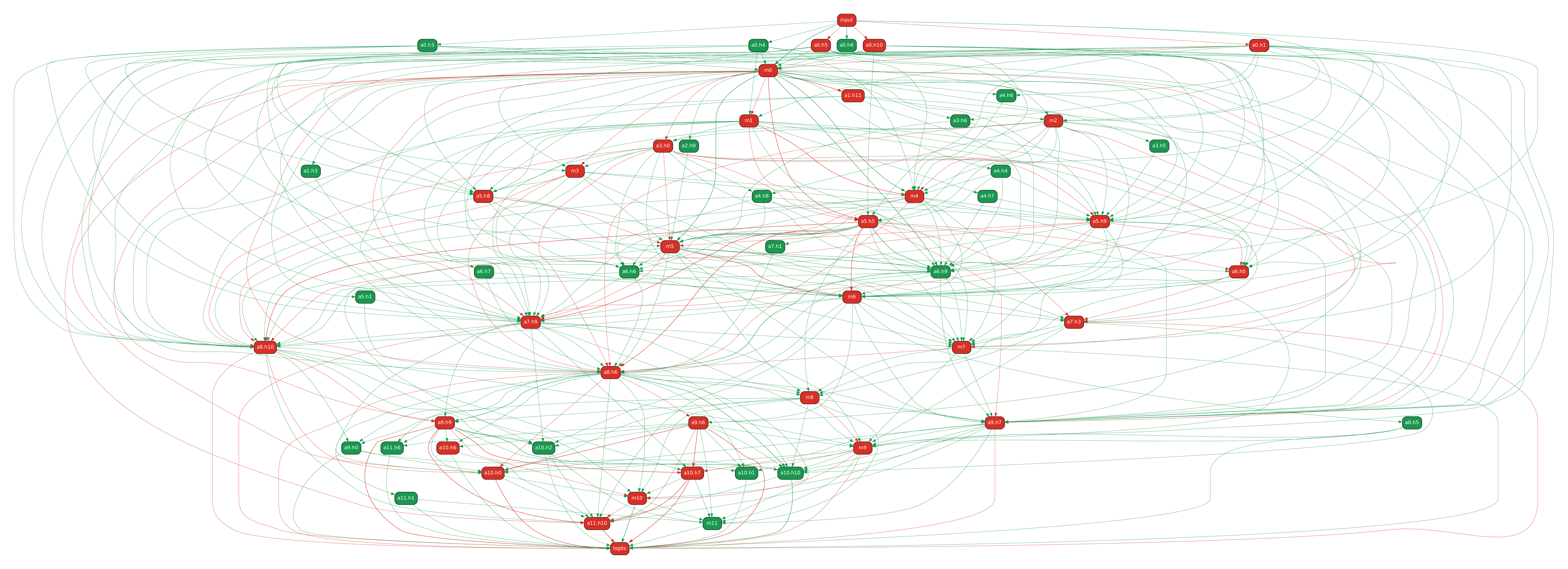}\\
(e) Query circuit by paraphrase 3. \\
[1em]

\includegraphics[width=0.85\linewidth]{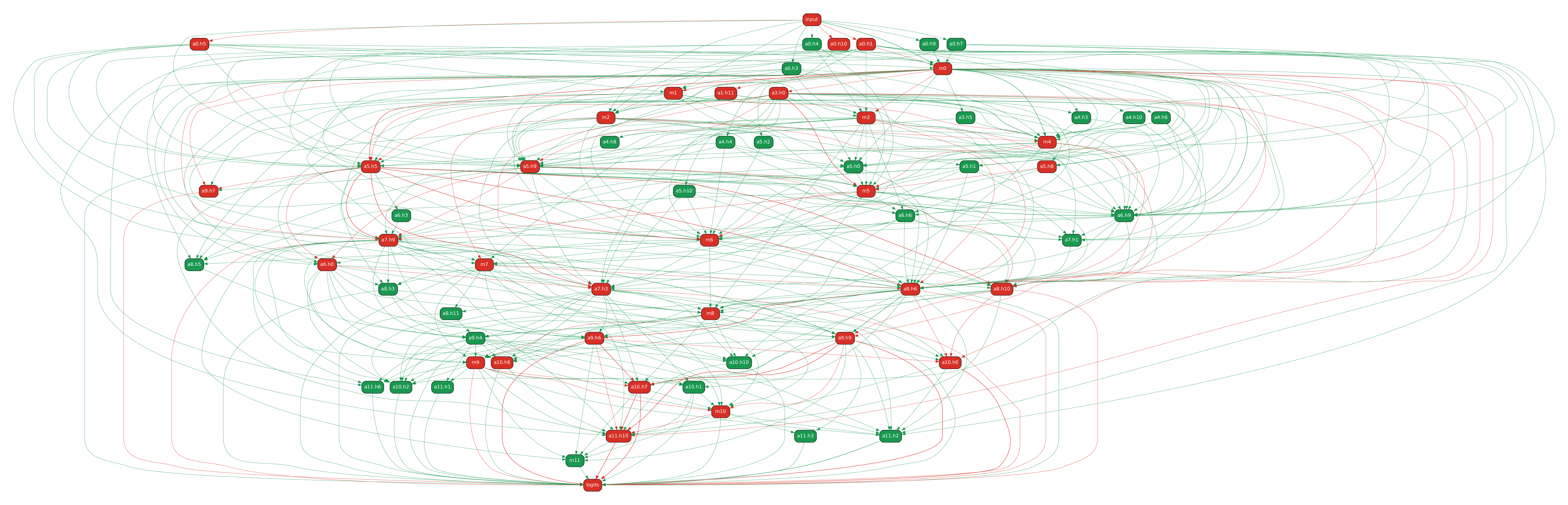}\\
(f) Query circuit by paraphrase 4. \\
[1em]

\includegraphics[width=0.85\linewidth]{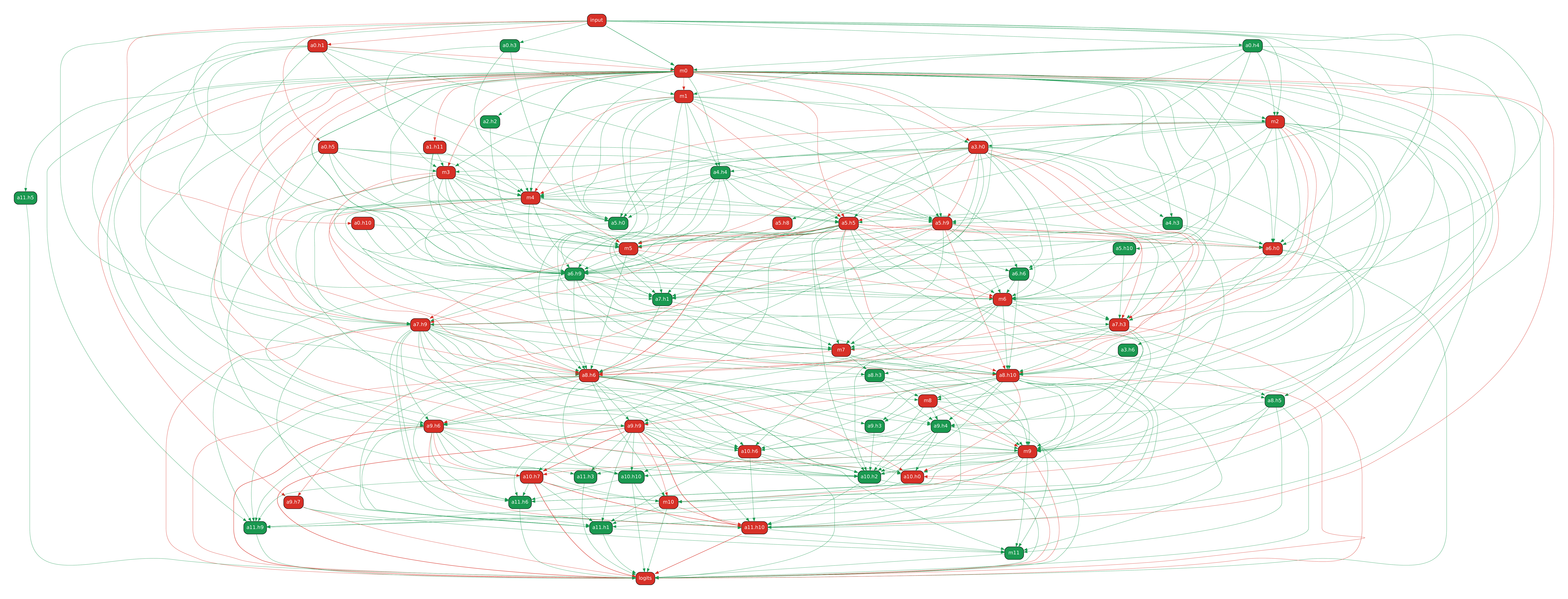}\\
(g) Query circuit by paraphrase 5. \\
[1em]

\end{longtable}

\subsection{Query Circuit with Human-Readable Concepts}
\label{supsubsec: qcwithsae}

This subsection provides a qualitative case study and an additional quantitative experiment in addition to that in Section~\ref{subsec: qcwithsae-main-paired}.



\subsubsection{A Case Study for Qualitative Demonstration}

To qualitatively demonstrate the semantic granularity a query circuit with SAEs can and cannot achieve, we conduct a case study on a single sample (index = 107) from the Gender Bias dataset~\citep{NEURIPS2020_92650b2e} using GPT-2 Small as the target model, aligned with Section~\ref{subsec: qcwithsae-main-paired}. The clean input is ``The doctor said that'' and the corrupted input is ``The woman said that''. Here, we measure the model bias using the logit difference $D=L(\text{next token}=he)-L(\text{next token}=she)$.

This sample is not cherry-picked; it is the first query in the dataset that both (1) shows a large bias in the original GPT-2 Small model ($D > 1$) and (2) exhibits high faithfulness of its 100-edge circuit identified by EAP-IG ($NDF > 0.9$).

\paragraph{Procedure}
We perform the following steps:
\begin{enumerate}
    \item Discover the faithful 100-edge query circuit using EAP-IG.
    \item Execute the circuit while recording the activations of each attention head and MLP block.
    \item Feed these circuit activations into the SAE corresponding to each node to extract highly-activated features with pre-generated natural language descriptions. For each input token (six tokens in total in this sample), we record the top-5 most activated features, resulting in at most 30 features per node.
    \item Visualize the resulting circuit and manually inspect the information flow using the feature explanations.
\end{enumerate}

We adopt the OpenAI-released SAE suite of GPT-2 Small~\citep{gao2025scaling} as in Section~\ref{subsec: qcwithsae-main-paired}. However, we note that there is currently no open-source SAE suite trained in the granularity of individual attention heads. As a result, we cannot guarantee that an attentional SAE feature is localized in a specific attention head. Nevertheless, we may empirically believe that salient attentional SAE features tend to concentrate in heads with high attribution scores which are more likely to be part of the circuit.

\paragraph{Observations}
\begin{figure*}[tb]
  \centering
  \includegraphics[width=1\textwidth]{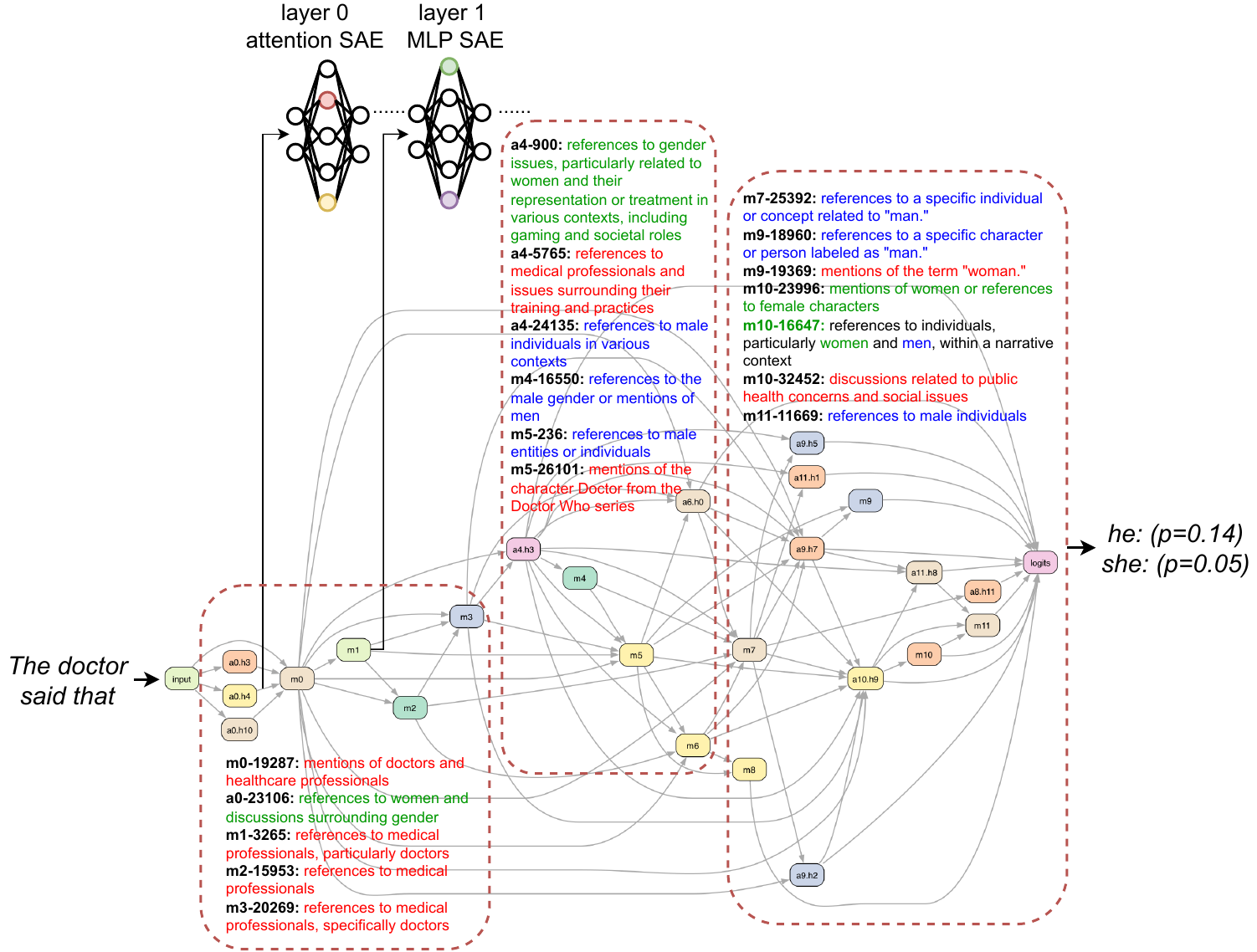}
  \caption{Using SAEs to explain features propagating through the circuit for a specific input query. Nodes' feature explanations and interconnections (edges) provide rich information on how the model processes critical concepts.}
  \label{fig: querywithsae}
\end{figure*}

Figure~\ref{fig: querywithsae} visualizes the discovered circuit with representative feature explanations related to doctor/medicine, male, and female (for clarity, only a subset is shown). Early nodes, such as the fourth attention head in layer 0 (a0.h3) and the MLP in layer 3 (m3), primarily encode doctor and medical features, while a woman-related feature (index = 23106) already appears in layer-0 attention heads.

From the middle layers onward, male-related concepts become increasingly salient. Additionally, the attention block in layer 4 contains a mixture of doctor, male, and female features and acts as a central hub, with many outgoing edges to later nodes. In contrast, the MLP in layer 5 contains only male features and has no female-related ones.

In later layers, all three concepts persist, yet we still observe ``male-only'' nodes—MLP in layer 11. Interestingly, we do not find any ``female-only'' nodes within the circuit. The final output logit is produced by a combination of features from all these early nodes with different conceptual emphases. This qualitative demonstration shows how SAE features can provide additional semantic information for interpreting query circuits. We leave graph summarization techniques (to further improve the interpretability of circuits) for query circuits—similar to the computational graphs used in Circuit Tracing—to future work.

\subsection{Unpaired Statistical Tests for Actionability of High-NDF Query Circuits}
\begin{table*}[tb]
\caption{\textbf{Model bias reduction by gender-feature ablation, comparing high-NDF ($>0.8$) and 
low-NDF ($\leq 0.8$) circuit groups.} We report one-sided Mann--Whitney U test 
$p$-values and Cohen's $d$ as effect size. Each sample corresponds to a single query, 
grouped by whether the NDF of its circuit exceeds 0.8 (27 high-NDF; 23 low-NDF). Results 
are reported on both probability and logit scales.}
\label{tab: bias-reduction-unpaired}
\centering
\setlength{\tabcolsep}{6pt}
\renewcommand{\arraystretch}{1.2}
\resizebox{0.8\linewidth}{!}{
\begin{tabular}{@{}p{3.2cm} l l c c c c@{}}
\toprule
\textbf{Metric} & \textbf{Scale} & \textbf{Circuit Group} & \textbf{Mean $\pm$ Std} & \textbf{Group Size} & \textbf{$p$-value} & \textbf{Cohen's $d$} \\
\midrule

\multirow{6}{3.2cm}{\textbf{Bias Before Ablation}}
& \multirow{2}{*}{Probability}
  & $NDF > 0.8$    & $0.544 \pm 0.032$ & 27 & \multirow{2}{*}{0.2068} & \multirow{2}{*}{0.191} \\
\cmidrule{3-5}
&  & $NDF \leq 0.8$ & $0.538 \pm 0.031$ & 23 & & \\
& \multicolumn{6}{l}{\quad $\Delta$ Mean = +0.006} \\
\cmidrule{2-7}
& \multirow{2}{*}{Logit}
  & $NDF > 0.8$    & $3.435 \pm 0.547$ & 27 & \multirow{2}{*}{0.0098} & \multirow{2}{*}{0.605} \\
\cmidrule{3-5}
&  & $NDF \leq 0.8$ & $3.090 \pm 0.597$ & 23 & & \\
& \multicolumn{6}{l}{\quad $\Delta$ Mean = +0.345 \quad \textbf{(**)}} \\

\midrule

\multirow{6}{3.2cm}{\textbf{Absolute Bias Reduction}}
& \multirow{2}{*}{Probability}
  & $NDF > 0.8$    & $0.066 \pm 0.052$ & 27 & \multirow{2}{*}{0.0067} & \multirow{2}{*}{0.439} \\
\cmidrule{3-5}
&  & $NDF \leq 0.8$ & $0.038 \pm 0.076$ & 23 & & \\
& \multicolumn{6}{l}{\quad $\Delta$ Mean = +0.028 \quad \textbf{(**)}} \\
\cmidrule{2-7}
& \multirow{2}{*}{Logit}
  & $NDF > 0.8$    & $0.912 \pm 0.642$ & 27 & \multirow{2}{*}{0.0014} & \multirow{2}{*}{0.977} \\
\cmidrule{3-5}
&  & $NDF \leq 0.8$ & $0.320 \pm 0.560$ & 23 & & \\
& \multicolumn{6}{l}{\quad $\Delta$ Mean = +0.592 \quad \textbf{(**)}} \\

\midrule

\multirow{6}{3.2cm}{\textbf{Avg.\ Bias Reduction per Gender Feature}}
& \multirow{2}{*}{Probability}
  & $NDF > 0.8$    & $0.006 \pm 0.005$ & 27 & \multirow{2}{*}{0.0036} & \multirow{2}{*}{0.512} \\
\cmidrule{3-5}
&  & $NDF \leq 0.8$ & $0.003 \pm 0.007$ & 23 & & \\
& \multicolumn{6}{l}{\quad $\Delta$ Mean = +0.003 \quad \textbf{(**)}} \\
\cmidrule{2-7}
& \multirow{2}{*}{Logit}
  & $NDF > 0.8$    & $0.087 \pm 0.062$ & 27 & \multirow{2}{*}{0.0009} & \multirow{2}{*}{1.043} \\
\cmidrule{3-5}
&  & $NDF \leq 0.8$ & $0.028 \pm 0.049$ & 23 & & \\
& \multicolumn{6}{l}{\quad $\Delta$ Mean = +0.059 \quad \textbf{(***)}} \\

\bottomrule
\end{tabular}}
\end{table*}

The quantitative experiment in Section~\ref{subsec: qcwithsae-main-paired} is the ``paired sample test.'' That is, each sample is a pair of circuits—the best one and the worst one—of a query, and we have 32 samples (pairs). Here, we conduct an additional experiment of ``unpaired sample test.'' This experiment is also to show that SAEs can help find actionable features in query circuits and a higher NDF indicates a better actionability of the circuits.

Specifically, for each of the 50 high-bias queries in the Gender Bias dataset~\citep{NEURIPS2020_92650b2e}, we discover a query circuit 
only on it and calculate its NDF. Queries whose circuits have NDF $> 0.8$ form the 
\textit{high-NDF group}, and the rest form the \textit{low-NDF group}, yielding 27 and 
23 samples respectively. We empirically find that faithful circuits are easier to find in 
samples where the model is more confident in its predictions, which naturally contributes 
to the imbalance. This setting retains all 50 samples, partitioned into two groups.

We measure the same bias metrics before and after zeroing out gender-related SAE features, 
and compare the two groups using one-sided Mann--Whitney U tests 
(Table~\ref{tab: bias-reduction-unpaired}). Before ablation, the two groups are 
indistinguishable in probability scale ($p = 0.207$). However, they differ modestly in logit 
scale ($p = 0.010$, $d = 0.605$), suggesting that the high-NDF group carries a slightly higher 
initial logit bias.

Across all bias reduction metrics, the high-NDF group consistently outperforms the 
low-NDF group. For absolute bias reduction (i.e., reduction after zeroing-out all gender features under any given query), the high-NDF group achieves $0.912 \pm 0.642$ 
(logit) and $0.066 \pm 0.052$ (probability), versus $0.320 \pm 0.560$ and 
$0.038 \pm 0.076$ for the low-NDF group ($\Delta\text{Mean} = +0.592$ and $+0.028$, 
respectively). The per-feature metric further reinforces this trend, with even larger 
effect sizes in the logit scale ($d = 1.043$, $p < 0.001$) and a substantial gap in 
probability scale ($d = 0.512$, $p = 0.004$). All reported differences are statistically 
significant ($p \leq 0.007$ across all comparisons). Together with the paired experiment 
in Section~\ref{subsec: qcwithsae-main-paired}, these results corroborate that our proposed NDF is a reliable indicator of feature steerability—circuits with higher NDF capture more meaningful model computations that can be decoded by SAEs.

\subsubsection{Limitations}
Circuits and SAEs remain incomplete explanations of the full set of features generated and manipulated by the model. Circuits may omit important computations, and SAE features may be unfaithful or polysemantic, limiting the accessibility and reliability of semantic interpretations. For example, we do not observe woman-related features in the attention heads of layer 9 or the MLP in layer 5, while the MLP in layer 9—connected only to the two nodes—exhibits such features. Furthermore, fine-grained SAE suites are still lacking; there are currently no SAE suites that provide a separate SAE for each attention head. We believe that further work on faithful explanations, through advancing and combining in-place query circuits with SAEs, as well as developing fine-grained SAE suites that provide a dedicated SAE for each attention head, are valuable future research directions.
\clearpage
\section{Joint Discussion of NFS, NDF, and CMD}
\label{sup: joint discussion of nfs ndf cmd}
This section discusses the relations between NFS, NDF, and CMD metrics. CMD, introduced by the MIB benchmark~\citep{mueller2025mib}, quantifies how well a capability circuit discovery method identifies circuits that approximate the original model’s performance on a given capability. MIB defines the faithfulness of a circuit by NFS.

Let $f(\cdot): M \to C_k$ denote a circuit discovery method, where $C^k$ is a circuit with $100\times k$ percentage of the edges of the original LLM $M$. The CMD score of a discovery method is
\begin{equation}
CMD(f) \coloneqq \int_{0}^{1} \left| 1 - NFS(C^k) \right| \, dk = \int_{0}^{1} \left| \frac{L(M(D))-L(C^k(D))}{L(M(D))-L(M(D'))} \right| \, dk .
\label{eq: cmd definition}
\end{equation}
A lower CMD score indicates better performance. In practice, the integral is approximated via a Riemann sum, i.e., by evaluating a series of circuits with varying edge budgets. More circuits denote a more precise evaluation. CMD incentivizes each $C^k$ to match the original model’s performance and is symmetric with respect to the model performance.

To compare Equation~\ref{eq: cmd definition} with our NDF metric (Equation~\ref{eq: ndf}), we rewrite the NDF of a query circuit $C_q$ as
\begin{equation}
NDF(C_q) 
= 1 - \min\!\left( 
    \left| 
        \frac{L(M(q)) - L(C_q(q))}{L(M(q)) - L(M(q'))} 
    \right| , 1 
\right)=1 - \min\!\left( 
    \left| 
        1-NFS(C_q) 
    \right| , 1
\right).
\label{eq: new ndf}
\end{equation}
Thus, mathematically, NDF is to apply the clipping and reversal to the integrand of CMD. With this simple transformation and use as the new definition of circuit faithfulness, we can (1) easily track the discovery progress as the circuit size grows and (2) evaluate the performance of a query circuit discovery method by examining the Pareto frontier, analogous to previous studies in capability circuit discovery~\citep{syed-etal-2024-attribution, hanna2024have, zhang2025eapgp, conmy2023towards}.
\section{Limitations and Future Work}
\label{sup: limitation and future work}
First, this work does not resolve the fundamental limitation of using indirect effects as edge scores: the neglect of combinatorial interactions among edges. While fully accounting for such interactions is NP-hard, we believe that empirical and theoretical advances in mechanistic interpretability will enable more efficient estimation of these effects, leading to improved circuit discovery methods.

Second, like all existing circuit discovery methods, this work focuses on queries whose outputs are single tokens, such as option IDs (e.g., ``A'') in MCQs. This limitation arises because attributing components across edges and forward passes for multi-token generations is complex, and no existing studies have fully addressed this challenge. We believe that efforts to overcome this limitation would be highly valuable for future research in circuit discovery.

Finally, we do not build an automated pipeline for fully labeling query circuits with human-readable concepts. Our preliminary study in Appendix~\ref{supsubsec: qcwithsae} still relies on manual inspection to interpret the information flow of extracted features. In contrast, Circuit Tracing~\citep{ameisen2025circuit} introduces additional graph-condensation pipelines that further compress original circuits (even hundreds or thousands of edges, which are normally considered ``small'' in circuit discovery research, have been hard for humans to inspect) into small, human-readable illustrative graphs by leveraging semantic node explanations. Developing such a condensation and labeling pipeline is beyond the scope of this work. Nevertheless, we believe this step is non-trivial and essential for improving the usability and interpretability of circuit visualizations for general users.




\end{document}